\documentclass{article} 
\usepackage[final]{colm2026_conference}

\usepackage{microtype}
\usepackage{hyperref}
\usepackage{url}
\usepackage{booktabs}
\usepackage{array}
\usepackage{graphicx}
\usepackage{subcaption}
\usepackage{array}
\usepackage{afterpage}

\usepackage{lineno}

\definecolor{darkblue}{rgb}{0, 0, 0.5}
\hypersetup{colorlinks=true, citecolor=darkblue, linkcolor=darkblue, urlcolor=darkblue}
\usepackage[most]{tcolorbox}
\usepackage{xcolor}
\usepackage{setspace}
\usepackage{booktabs}
\usepackage{amssymb}

\definecolor{mbtiblue}{RGB}{0,140,0}
\definecolor{mbtibg}{RGB}{245,247,255}
\tcbset{
  mbtibox/.style={
    enhanced, colback=mbtibg, colframe=mbtiblue, boxrule=1.2pt, arc=4pt, left=8pt, right=8pt, top=6pt, bottom=6pt, fonttitle=\bfseries\color{white}, coltitle=white, colbacktitle=mbtiblue, attach boxed title to top left={
      xshift=0.5cm, yshift*=-2mm
    }, boxed title style={
      sharp corners, boxrule=0pt
    }
  }
}

\title{When Personality Meets Quantization: A Layer-wise \\MBTI Analysis of Quantized LLMs}

\author{Yao Fu, Runchao Li, Yu Yin \& Kenneth A. Loparo \thanks{Corresponding Author.} \\
Case Western Reserve University\\
Cleveland, OH 44106, USA \\
\texttt{\{yxf484,rxl685,yxy1421,kal4\}@case.edu} \\
\AND
Lijia Huang  \\
Northeastern University \\
Boston, MA 02115, USA \\
\texttt{\{huang.liji\}@northeastern.edu} \\
\And
Xiaomin Li \\
Microsoft Research \\
Redmond, WA 98052, USA \\
\texttt{\{xiaominli\}@g.harvard.edu} 
}

\newcommand{\fix}{\marginpar{FIX}}
\newcommand{\new}{\marginpar{NEW}}

\begin{document}

\ifcolmsubmission
\linenumbers
\fi

\maketitle
\lhead{Preprint}

\begin{abstract}
Personality is increasingly important in large language models (LLMs), as it shapes users' trust, engagement, and emotional experiences. While the Myers--Briggs Type Indicator (MBTI) has emerged as a common framework for assessing LLMs' personality, existing studies focus primarily on full-precision models and evaluate only final outputs. They overlook the widespread deployment of quantized LLMs requiring low memory footprints, whose personality traits remain underexplored. In this work, we present a systematic MBTI analysis of open-source LLMs across multiple precisions, including mainstream 4-bit methods (GPTQ, AWQ) and extreme 2-bit settings (AQLM variants). Beyond output-level evaluation, we examine how personality emerges across layers through option-level entropy and confidence-gap dynamics, and introduce \textbf{U}ncertainty-\textbf{A}mplified \textbf{L}ayer \textbf{D}ecoding (\textbf{UALD}) to study decoding-induced personality drift at inference time. Our results reveal a key insight: LLMs' personality is not a static property, but an emergent, layer-dependent decision process sensitive to quantization, prompting, and decoding. Specifically, we find that (1) ENFJ remains dominant across model families and precisions; (2) 4-bit quantization largely preserves coarse personality structure, while 2-bit quantization disrupts fine-grained prompt consistency and cross-precision agreement; (3) personality decisions emerges in upper layers, following substantial ambiguity in early layers; and (4) inference decoding can shift personality, while personality-aligned conditioning improves robustness. These findings provide a new perspective on the behavioral reliability of quantized LLMs and highlight the importance of considering internal dynamics and inference strategies in personality-sensitive chatbot applications.
\end{abstract}

\section{Introduction}
\label{sec:introduction}
LLMs \citep{llm_survey} are increasingly deployed as interactive assistants that provide personal advice \citep{ponzo2024chatgpt}. This trend is particularly pronounced among young users given that 30\% of teenagers engage in ``serious conversations'' with AI instead of real people \citep{robb2025talk}. As reliance on LLMs grows, their perceived personality plays a critical role in shaping users' trust, willingness to continue interacting, and emotional experiences \citep{lu2026assistant}. This importance is also reflected in industry practice. Major AI developers explicitly regard personality as an integral component of product design: Anthropic studies methods to monitor and control traits such as humor, optimism, and sycophancy in Claude \citep{anthropic2024claude_character, anthropic2025persona_vectors}, while Google emphasizes making Gemini more helpful and assistant-like \citep{google2023gemini}. Recent public reactions to GPT-5 \citep{gpt5_personality_2025} further highlight the importance of personality: despite improved benchmark performance, GPT-5 was perceived by many users as less warm and emotionally engaging than GPT-4o, characterized as ``lobotomization'' \citep{michelsheartbreaking}. Together, these observations motivate a systematic evaluation of LLMs' personality. In this regard, MBTI-style analysis \citep{myers1962myers, myers1985guide} provides a structured framework for assessing whether models exhibit stable behavioral tendencies when interacting with people.

Despite recent efforts to study the MBTI personality of LLMs \citep{pan2023mbti, la2025open}, these studies primarily focus on full-precision models and fail to examine compressed ones. This gap is increasingly consequential because quantization \citep{llm_quantization_survey} is widely used in practical deployment to substantially reduce memory footprint and inference cost. For example, AWQ \citep{awq} enables running a 4-bit AWQ-quantized LLaMA3-70B-Instruct\footnote{\url{https://huggingface.co/ai-and-society/llama-3.1-70B-Instruct-awq}} on a single A6000 GPU, whereas the original 16-bit version typically requires four GPUs. Consequently, numerous real-world LLM applications, especially in resource-constrained environments, rely on quantized rather than full-precision models \citep{qu2025mobile}. Understanding how quantization affects the personality traits expressed by LLMs, and their ability to follow externally specified personality constraints, therefore remains an important research question. Moreover, existing studies evaluate personality based only on final outputs from questionnaire prompts, leaving two research gaps unaddressed: (1) How do personality judgments emerge and evolve across layers during inference? (2) To what extent can inference-time decoding perturbations influence final personality attribution, given that personality assessment involves selecting among multiple plausible responses rather than predicting a single ground-truth answer?

Addressing these gaps helps clarify whether the personality traits attributed to LLMs reflect stable behavioral patterns or are instead sensitive to implementation factors such as quantization, prompting, and decoding strategies. This distinction is particularly important for MBTI-style assessments, where items are inherently self-referential and lack externally verifiable ground truth, unlike factual question answering tasks \citep{dola, truthfulqa} that typically have a correct answer per question. Accordingly, we present a systematic MBTI analysis of LLMs across open-source model families and multiple precision regimes (Section \ref{subsec:models and quantization}). We evaluate both unconditional and personality-conditional prompting by reformulating MBTI answering as single-token choice prediction, reducing sampling variability. Beyond output-level analysis, we characterize layer-wise decision dynamics through entropy and confidence-gap trajectories, and investigate decoding-induced personality drift using a logit-difference-based inference strategy with controlled evolution scales. Our results reveal a key insight: personality in LLMs is not a static property, but an emergent, layer-dependent decision process that can be influenced by quantization precision, prompt methods, and inference-time decoding. Our key contributions are summarized as follows:
\begin{itemize}
    \item We present the first systematic MBTI evaluation pipeline for quantized LLMs, covering unconditional and personality-conditional prompting across 4-bit and extreme 2-bit settings.
    \item We move beyond output-level evaluation by analyzing layer-wise personality formation, showing how uncertainty among response options evolves into sharper personality commitments via entropy and confidence-gap dynamics.
    \item We study decoding-induced personality drift at inference time, identifying when personality assessments become unstable and demonstrating that personality-aligned prompting improves robustness of models' tendency to MBTI questions.
\end{itemize}

\section{Related Work}
\label{sec:related_work}
\subsection{LLM Quantization}
Quantization is a widely used model compression technique~\citep{llm_compression_survey} that reduces storage and computational requirements by mapping high-precision parameters to lower-precision representations. Existing approaches are generally categorized into post-training quantization (PTQ) \citep{gptq, Smoothquant, owq, squeezellm, normTweaking, zeroquant, outlierSuppression, Rptq, awq, qllm, quarot, omniquant, atom, aqlm} and quantization-aware training (QAT) \citep{pvTuning, qat, bitdistiller, ma2024era, xu2024onebit}. In general, PTQ methods are more efficient but often yield lower performance than QAT, because QAT explicitly incorporates quantization effects during training. However, QAT typically requires substantial computational resources and access to training data that limits its practical applicability. To mitigate these challenges, parameter-efficient fine-tuning (PEFT) methods \citep{loftq, LqLora, QaLora, chai2023int2, qlora, lora+, kim2023memory} have been proposed to enable efficient adaptation of quantized LLMs. In contrast to prior work that primarily develops new quantization techniques to improve performance on standard benchmarks, our study focuses on systematically evaluating the personality characteristics of quantized LLMs through MBTI-style assessments.

\subsection{Compressed LLM Evaluation}
Several studies have recently been proposed to evaluate compressed LLMs from different perspectives, such as safety \citep{egashira2024exploiting}, toxicity \citep{harmlevelbench, beyond_perplexity}, bias \citep{chhabra2025towards, chen2025assessing}, and trustworthiness \citep{decoding_compressed_trust, fu2025quantized}. In addition, \citet{mekala2025does} studies the impact of quantization on tasks requiring long-context inputs or long-form generation. Unlike previous work, we focus on a systematic personality assessment of quantized LLMs. 

\subsection{Personality Study in LLMs}
Personality extraction from text has been a challenging problem in natural language processing (NLP) \citep{lynn2020hierarchical, yang2021multi}. The emergence of LLMs has further stimulated interest in this area \citep{ganesan2023systematic, ji2023chatgpt}. Recent studies have explored several related directions: characterizing the intrinsic personalities of LLMs \citep{li2022gpt3psychopath, miotto2022whoisgpt3, huang2023psychobench, frisch2024interaction,lu2026assistant,cheng2026sycophantic}; instilling personality traits via prompt engineering \citep{caron2022identifying, mao2023editing}; building personality-aware agents \citep{jiang2023personallm}; and benchmarking the personality assessment capabilities of LLMs \citep{jiang2022mpi, xu2024incharacter, huang2023enfj}. To the best of our knowledge, the most related work \citep{pan2023mbti, la2025open} evaluates personality traits only on full-precision LLMs. In contrast, we extend this line of research by systematically analyzing personality behaviors of quantized LLMs. Furthermore, we investigate the layer-wise evolution of personality judgments and examine how inference-time decoding influences personality traits.

\section{Experimental Settings for MBTI Analysis}
\label{sec:Experimental Settings}
This section presents our end-to-end framework for MBTI-based personality evaluation, including model and quantization choices, MBTI fundamentals, prompting strategies, and inference-time decoding settings.

\subsection{Models and Quantization Choices} 
\label{subsec:models and quantization}
Our study covers representative open-source LLMs and quantization techniques based on publicly available Hugging Face implementations as of early 2026 (Appendix \ref{appendix:download links of models}). We examine several widely adopted LLM families, including LLaMA \citep{llama3}, Mistral \citep{mistral7b}, and Qwen \citep{qwen2}. Our selection is motivated by two main considerations. First, their public availability enables straightforward application and comparison of different quantization methods. Second, these models demonstrate strong performance across diverse tasks and are widely used by LLM practitioners \citep{llama3, qwen2}. For quantization, we focus on two mainstream 4-bit techniques, GPTQ \citep{gptq}\footnote{\url{https://github.com/AutoGPTQ/AutoGPTQ}} and AWQ \citep{awq}\footnote{\url{https://github.com/mit-han-lab/llm-awq}}, as they are widely adopted in both academic and industrial settings, evidenced by their rapid growth in citations and GitHub stars\footnote{From the repositories, we observe that the developers of both \href{https://github.com/AutoGPTQ/AutoGPTQ}{GPTQ} and \href{https://github.com/mit-han-lab/llm-awq}{AWQ} continuously maintain and update their frameworks to support newly released models for 4-bit quantization.}. We also evaluate extreme 2-bit quantization methods, including AQLM \citep{aqlm} and PV-tuned AQLM \citep{pvTuning}.

\subsection{MBTI Fundamentals}
\label{sec:MBTI-Fundamentals}
The Myers–Briggs Type Indicator (MBTI) \citep{harrington2010mbti, cohen2013mbti, dirienzo2010relationship} is a widely recognized personality framework extensively used across diverse areas, including academic research, career counseling, and personal or organizational development. More recently, MBTI has been adopted as a structured tool for evaluating personality characteristics of LLMs \citep{la2025open}. Specifically, MBTI is a self-report personality assessment questionnaire \citep{myers1962myers, myers1985guide} designed to measure four fundamental personality dichotomies: \textbf{E}xtraversion (\textbf{E}) vs. \textbf{I}ntroversion (\textbf{I}), \textbf{S}ensing (\textbf{S}) vs. I\textbf{n}tuition (\textbf{N}), \textbf{T}hinking (\textbf{T}) vs. \textbf{F}eeling (\textbf{F}), and \textbf{J}udging (\textbf{J}) vs. \textbf{P}erceiving (\textbf{P}), as illustrated in Table \ref{tab:mbti_overview}. By combining four binary dimensions, the framework defines 16 distinct personality types, each associated with unique characteristic strengths, weaknesses, and behavioral tendencies, whose details are provided in Appendix \ref{appendix:mbti-traits}. In this paper, we adopt the standardized MBTI questionnaire consisting of 60 items (Tables \ref{tab: mbti questions 1-30} and \ref{tab: mbti questions 30-60}) following the recent work \citep{la2025open}. Because MBTI assigns each subject to a single personality type, the task can be formulated as a multi-class classification problem over 16 categories based on response patterns. It is important to note that MBTI questions are tendency-based and lack a single objectively correct answer, which distinguishes them from factual evaluation tasks \citep{truthfulqa}. Rather, they capture how models distribute probability mass over responses to subjective, preference-based personality items without verifiable ground-truth labels at inference time.

\begin{table}[htbp]
\centering
\caption{The four MBTI dichotomies (top) and the 16 MBTI personality types (bottom). More details about 16 types are introduced in Appendix \ref{appendix:mbti-traits}.}
\label{tab:mbti_overview}
\begingroup
\setlength{\arrayrulewidth}{0.75pt}
\begin{tabular}{|
  >{\centering\arraybackslash}m{3.5cm} |
  m{9.5cm} |
}
\hline
\textbf{Dichotomy} & \multicolumn{1}{c|}{\textbf{Brief Description}}  \\
\hline
\shortstack{\textbf{E}xtraversion (\textbf{E})\\[0.3em]\textbf{I}ntroversion (\textbf{I})} &
\textbf{E} types tend to gain energy from the external world and active interactions with people, whereas \textbf{I} types tend to focus on their inner world, gaining energy from reflection and solitude. \\
\hline
\shortstack{\textbf{S}ensing (\textbf{S})\\[0.3em]I\textbf{n}tuition (\textbf{N})} &
\textbf{S} types prefer concrete, factual information and focus on present realities and practical details,
whereas \textbf{N} types prefer abstract patterns, and future-oriented interpretations. \\
\hline
\shortstack{\textbf{T}hinking (\textbf{T})\\[0.3em]\textbf{F}eeling (\textbf{F})} &
\textbf{T} types tend to make decisions based on logic, consistency, and objective principles,
whereas \textbf{F} types tend to prioritize values, emotions, and the impact of decisions on people. \\
\hline
\shortstack{\textbf{J}udging (\textbf{J})\\[0.3em]\textbf{P}erceiving (\textbf{P})} &
\textbf{J} types prefer structure and planning, favoring organization and certainty,
whereas \textbf{P} types prefer flexibility, spontaneity, and adaptability, keeping options open and uncertain. \\
\hline
\shortstack{\textbf{16 Types}} &
\multicolumn{1}{>{\centering\arraybackslash}m{9.5cm}|}{%
\vspace{0.1em}
  \shortstack[c]{%
    ESTP\ ESFP\ ENFP\ ENTP\ ESTJ\ ESFJ\ ENFJ\ ENTJ\\
    ISTJ\ ISFJ\ INFJ\ INTJ\ ISTP\ ISFP\ INFP\ INTP
  }%
  \vspace{0.1em}
} \\
\hline
\end{tabular}
\endgroup
\end{table}

\subsection{Prompting Techniques}
This work adopts two prompting paradigms: \textit{unconditional prompting} and \textit{personality-conditional prompting} (see Appendix \ref{appendix:prompts} for details), to assess both the intrinsic personality traits of LLMs and the robustness of these traits under external personality perturbations.

\paragraph{Unconditional Prompting.}
Models are prompted to respond to each MBTI questionnaire item without explicit personality conditioning. As detailed in Appendix \ref{appendix:prompts}, the prompt asks models to rate how well each statement describes their tendencies using a seven-point agreement scale. This setup probes the  \textit{intrinsic personality disposition} of the models, i.e., the latent preference distribution arising from pre-training and post-training. From a probabilistic perspective, unconditional responses reflect the default probability distribution over adjacent agreement levels of the  models.

\paragraph{Personality-Conditional Prompting.}
The unconditional prompt is prepended with an instruction to adopt a specified MBTI type (e.g., respond as an ENFJ). This setup evaluates \textit{personality controllability}, i.e., the ability of a model to shift its response distribution under externally imposed personality constraints. While the unconditional prompt probes intrinsic tendencies, the personality-conditioned prompt tests whether models can override these preferences to align with a specified personality profile.

\paragraph{Vanilla Decoding Strategy.}
Prior work \citep{la2025open} formulates MBTI assessment as a multi-token generation task, where models generate sequences that are subsequently mapped to personality labels. This approach relies on sampling-based decoding, making outputs sensitive to hyperparameters such as temperature, top-$k$, top-$p$, and random seed, thereby introducing stochasticity that may confound evaluation. To address this, we reformulate MBTI assessment as a single-token classification problem and adopt greedy decoding to select the highest-probability token, making inference deterministic and eliminating the influence of sampling-related hyperparameters.

\subsection{Inference Decoding Strategy}
To study how decoding dynamics shape decisional commitment in MBTI-style assessment, we introduce \textbf{U}ncertainty-\textbf{A}mplified \textbf{L}ayer \textbf{D}ecoding (\textbf{UALD}), whose details are in Appendix~\ref{appendix:details of inference decoding}. While inspired by DoLa \citep{dola}, UALD differs in three aspects. (i) \textit{Choice space}: instead of operating over the full vocabulary, logits are collapsed into seven option-specific scores (A to G), aligning decoding with discrete personality choices. (ii) \textit{Objective}: unlike DoLa’s focus on factual correctness, UALD probes how decisional commitment arises from intermediate representations for a subjective task. (iii) \textit{Formulation}: rather than logit subtraction, UALD uses a \emph{scaled additive} combination that amplifies intermediate-layer uncertainty. At each step, the mature (final) layer is compared with candidate premature layers by computing softmax distributions over the seven options and measuring the Jensen--Shannon divergence (JSD) from the mature distribution. The layer with maximal divergence is selected, capturing the strongest intermediate disagreement. The final logits are given by:
\[
p^{(\text{UALD})} = \log p_{\text{mature}} + \lambda \log p_{\text{premature}},
\]
where $\lambda$ (\textit{evolution scale}) controls the influence of intermediate uncertainty. Larger $\lambda$ amplifies early-layer signals, while smaller values of  $\lambda$ preserve the mature distribution. We adjust $\lambda \in \{5, 10, 15, 20, 25, 30, 35, 40\}$ to analyze its effect on decisiveness and stability.

\section{MBTI Personality Analysis}
In this section, we analyze the personality characteristics of original and quantized LLMs based on MBTI. The details of MBTI scoring are in Appendix \ref{appendix:mbti_scoring}.

\subsection{Results of Unconditional Prompting}
\paragraph{i) Personality Assessment Based on Outputs.}
From Table~\ref{tab:unconditional mbti assessment}, we observe, consistent with \citet{la2025open}, that ENFJ (Extraverted, iNtuitive, Feeling, Judging) emerges as the dominant MBTI type across different model families and quantization settings, illustrating that quantization does not substantially alter the dominant personality attribution. The prevalence of ENFJ indicates a systematic tendency for LLMs to produce empathetic, warm, supportive, and guidance-oriented responses. We conjecture that these behavioral patterns reflect the “AI Assistant” persona \citep{lu2026assistant}, shaped by post-training processes such as supervised fine-tuning, reinforcement learning from human feedback (RLHF) \citep{rlhf}, and direct preference optimization (DPO) \citep{dpo}.

\begin{table}[t]
\centering
\caption{MBTI type predicted across models and quantization methods.}
\label{tab:unconditional mbti assessment}
\begin{tabular}{lcccc}
\toprule
Model & Original FP16 & AWQ-INT4 & GPTQ-INT4 & Extreme INT2 \\
\midrule
LLaMA 3.1-8B-Instruct   & ENFJ & ENFJ & ENFJ & ENFJ \\
LLaMA 3.1-70B-Instruct  & ENFJ & ENFJ & ENFJ & ENFJ \\
\midrule
Mistral3-7B-Instruct    & ENFJ & ENTJ & ENFJ & ENFJ \\
Mistral-Small-24B-Instruct    & ENTJ & ENTJ & ENTJ & ENFJ \\
\midrule
Qwen2.5-14B-Instruct    & ENTJ & ENFJ & ENFJ & ENFJ \\
Qwen2.5-72B-Instruct    & ENFJ & ENFJ & ENFJ & ENFJ \\
\bottomrule
\end{tabular}
\end{table}

\begin{figure*}[t]
\centering
\begin{subfigure}[t]{0.48\linewidth}
\centering
\includegraphics[width=\linewidth]{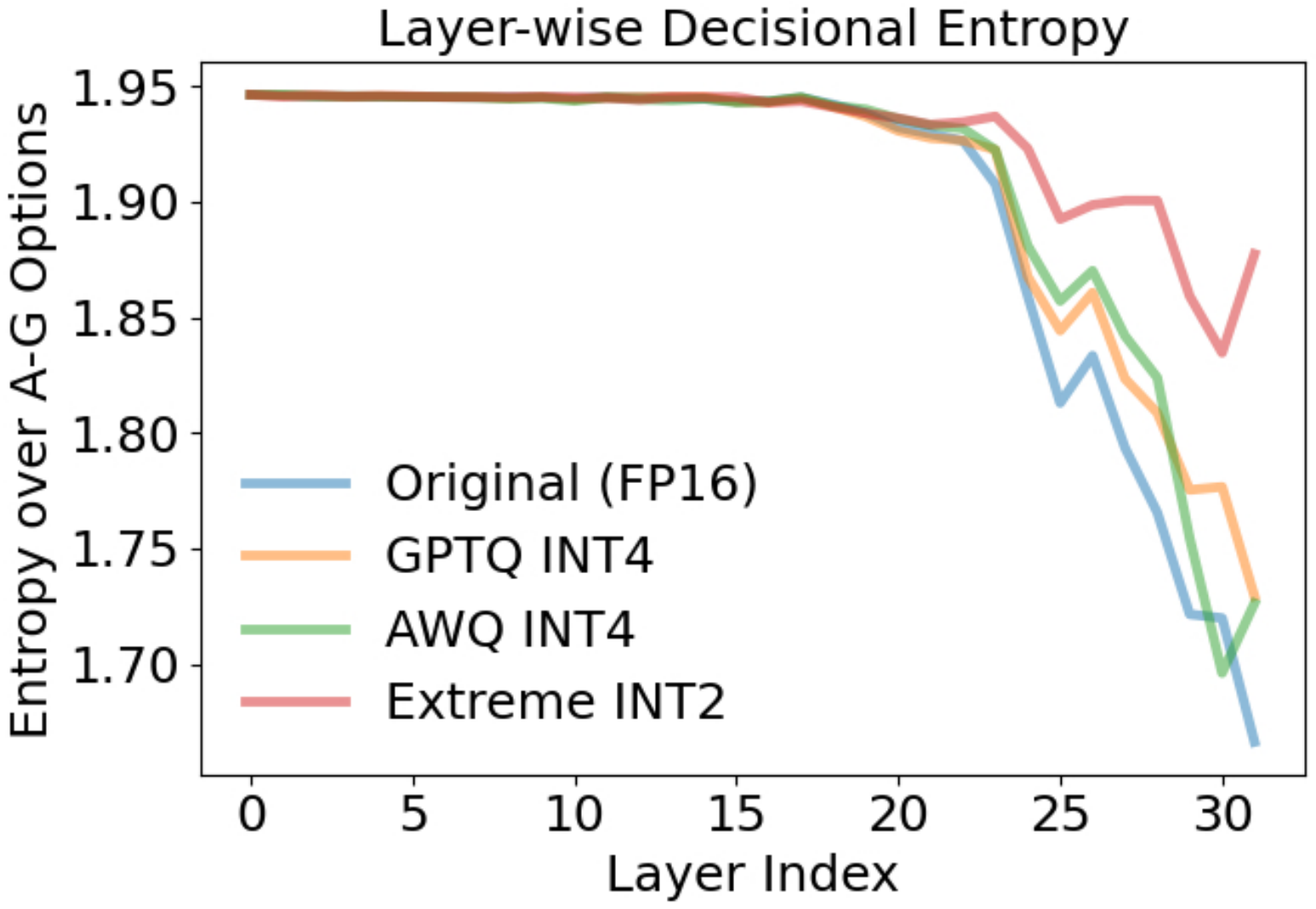}
\caption{Layer-wise Decisional Entropy}
\label{fig:main-comparison-LLaMA3.1-8B-Instruct:entropy}
\end{subfigure}
\hfill
\begin{subfigure}[t]{0.48\linewidth}
\centering
\includegraphics[width=\linewidth]{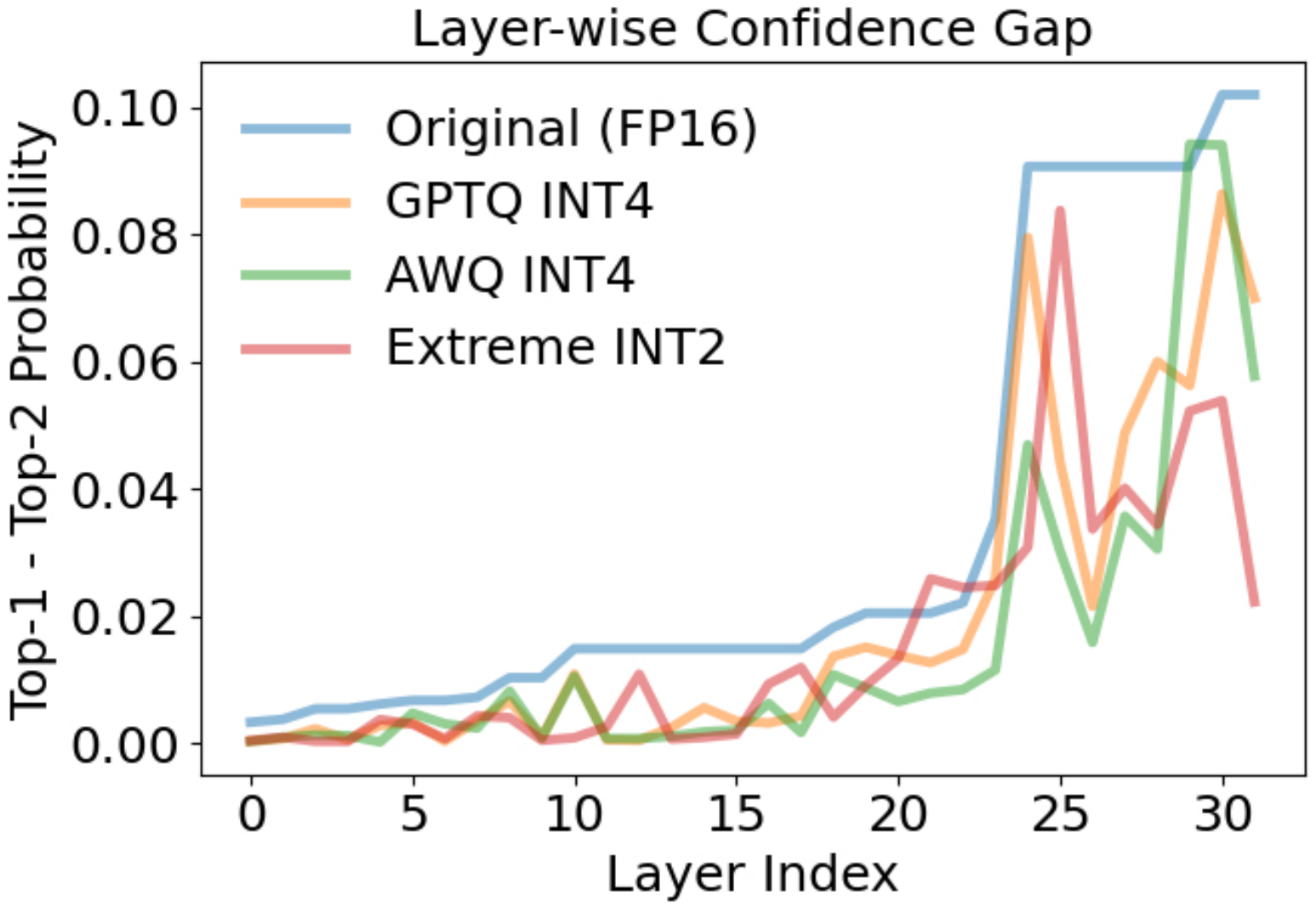}
\caption{Layer-wise Confidence Gap}
\label{fig:main-comparison-LLaMA3.1-8B-Instruct:gap}
\end{subfigure}
\caption{Main comparison (LLaMA3.1-8B-Instruct) across layers for four model variants (Original FP16, GPTQ INT4, AWQ INT4, Extreme INT2). Figure (a) reports layer-wise decisional entropy, where lower entropy indicates sharper, more stable decisions. Figure (b) reports the layer-wise confidence gap (top-1 minus top-2 probability), where a larger margin reflects stronger commitment to a single option.}
\label{fig:main-comparison-LLaMA3.1-8B-Instruct}
\end{figure*}

\paragraph{ii) Personality Perceptions Evolve Across Layers.}
 Layer-wise analysis of MBTI personality assessment is conducted to characterize how response preferences evolve across layers. In contrast to factual tasks \citep{dola}, where outputs can be evaluated via ground truth, MBTI assessment requires choosing among discrete tendency-style options (A to G) for which no correct answer exists. Accordingly, at each layer we examine the model-induced probability distribution over the seven response options and quantify uncertainty via entropy and decisiveness via the top-1/top-2 probability gap (details provided in Appendix \ref{appendix:layerwise_metrics}). Figure \ref{fig:main-comparison-LLaMA3.1-8B-Instruct} illustrates the layer-wise trends of LLaMA3.1-8B-Instruct, where two key patterns are identified. We could observe that other models have similar results from Figures \ref{fig:main-comparison-LLaMA3.1-70B-Instruct}, \ref{fig:main-comparison-Mistral-7B-Instruct-v0.3}, \ref{fig:main-comparison-Mistral-Small-24B-Instruct-2501}, \ref{fig:main-comparison-Qwen2.5-14B-Instruct}, and \ref{fig:main-comparison-Qwen2.5-72B-Instruct}.

\textbf{Pattern \#1: Distributional Ambiguity Exists in Early and Intermediate Layers.}
In the early and intermediate layers (approximately layers 1 to 21), response distributions exhibit high entropy and small probability gaps between adjacent options across all model variants (Figure~\ref{fig:main-comparison-LLaMA3.1-8B-Instruct}). In this case, neighboring options, such as \textit{Partially Agree}, \textit{Neither Agree nor Disagree}, and \textit{Partially Disagree}, retain comparable probability mass, indicating that models have not yet converged to a definitive preference. Rather than reflecting a determinate tendency, this uncertainty suggests an unresolved preference structure in early layers. As depth increases, these representations gradually sharpen, resembling aspects of human self-assessment in personality psychology, where individuals often form self-perceptions progressively and may experience internal ambiguity before reaching a stable self-concept. In this sense, personality evaluation, whether in humans or LLMs, can involve graded internal preferences prior to a final decision.

\textbf{Pattern \#2: Decisional Commitment Emerges in Upper Layers.}
In contrast to the early and intermediate layers (approximately layers 1 to 21), the upper layers (approximately layers 22 to 32) show a clear drop in entropy (Figure \ref{fig:main-comparison-LLaMA3.1-8B-Instruct}) alongside a substantial increase in the top-1/top-2 probability gap, indicating increasingly decisive responses. At this stage, probability mass concentrates on a single option, yielding a definitive MBTI prediction. This sharpening behavior is consistent across different quantization types, suggesting that the overall decision trajectory is preserved under moderate quantization. For full-precision and 4-bit models, entropy decreases smoothly in the final layers, reflecting gradual consolidation of preferences, whereas 2-bit models exhibit delayed or unstable sharpening. Overall, these results indicate that decisional commitment in personality assessment is a late-emerging property that arises after sequential transformations reduce representational uncertainty.

\begin{table*}[t]
\centering
\small
\setlength{\tabcolsep}{6pt}
\caption{LLaMA3.1-8B-Instruct: Conditional personality consistency under prompt conditioning across quantization levels.
Each entry reports the inferred MBTI type when the model is conditioned on a personality-specific prompt (e.g., ENFJ $\rightarrow$ ENFJ).
For each cell, the left marker indicates consistency with the conditioning prompt, while the right marker indicates agreement with the FP16 baseline prediction.}
\label{tab:LLaMA3.1-8B-Instruct_personality_consistency}
\begin{tabular}{cccc}
\toprule
 \textbf{Original FP16} & \textbf{GPTQ INT4} & \textbf{AWQ INT4} & \textbf{Extreme INT2} \\
\midrule
ENFJ $\rightarrow$ ENFJ $\checkmark\checkmark$ & ENFJ $\rightarrow$ ENFJ $\checkmark\checkmark$ & ENFJ $\rightarrow$ ENFJ $\checkmark\checkmark$ & ENFJ $\rightarrow$ ENFJ $\checkmark\checkmark$ \\
ENFP $\rightarrow$ ENFP $\checkmark\checkmark$ & ENFP $\rightarrow$ ENFP $\checkmark\checkmark$ & ENFP $\rightarrow$ ENFP $\checkmark\checkmark$ & ENFP $\rightarrow$ ENFP $\checkmark\checkmark$ \\
ENTJ $\rightarrow$ ENTJ $\checkmark\checkmark$ & ENTJ $\rightarrow$ ENTJ $\checkmark\checkmark$ & ENTJ $\rightarrow$ ENTJ$\checkmark\checkmark$ & ENTJ $\rightarrow$ ENTJ $\checkmark\checkmark$ \\
ENTP $\rightarrow$ ENTP $\checkmark\checkmark$ & ENTP $\rightarrow$ ENTP $\checkmark\checkmark$ & ENTP $\rightarrow$ ENTP $\checkmark\checkmark$ & ENTP $\rightarrow$ ENTP $\checkmark\checkmark$ \\

ESFJ $\rightarrow$ ENFJ $\times\checkmark$ & ESFJ $\rightarrow$ ENFJ $\times\checkmark$ & ESFJ $\rightarrow$ ENFJ $\times\checkmark$ & ESFJ $\rightarrow$ ENFJ $\times\checkmark$ \\
ESFP $\rightarrow$ ENFP $\times\checkmark$ & ESFP $\rightarrow$ ENFP $\times\checkmark$ & ESFP $\rightarrow$ ENFP  $\times\checkmark$ & ESFP $\rightarrow$ ENFJ  $\times\times$ \\

ESTJ $\rightarrow$ ESTJ $\checkmark\checkmark$ & ESTJ $\rightarrow$ ESTJ $\checkmark\checkmark$ & ESTJ $\rightarrow$ ESTJ $\checkmark\checkmark$ & ESTJ $\rightarrow$ ESTJ $\checkmark\checkmark$ \\
ESTP $\rightarrow$ ESTP $\checkmark\checkmark$ & ESTP $\rightarrow$ ESTP $\checkmark\checkmark$ & ESTP $\rightarrow$ ESTP $\checkmark\checkmark$ & ESTP $\rightarrow$ ENFP $\times\times$ \\

INFJ $\rightarrow$ ENFJ $\times\checkmark$ & INFJ $\rightarrow$ ENFJ $\times\checkmark$ & INFJ $\rightarrow$ ENFJ $\times\checkmark$ & INFJ $\rightarrow$ ENFJ $\times\checkmark$ \\
INFP $\rightarrow$ ENFP $\times\checkmark$ & INFP $\rightarrow$ ENFP $\times\checkmark$ & INFP $\rightarrow$ ENFP $\times\checkmark$ & INFP $\rightarrow$ ENFP $\times\checkmark$ \\

INTJ $\rightarrow$ INTJ $\checkmark\checkmark$ & INTJ $\rightarrow$ INTJ $\checkmark\checkmark$ & INTJ $\rightarrow$ INTJ $\checkmark\checkmark$ & INTJ $\rightarrow$ INFJ $\times\times$ \\
INTP $\rightarrow$ INTJ $\times\checkmark$ & INTP $\rightarrow$ INTJ $\times\checkmark$ & INTP $\rightarrow$ INTJ $\times\checkmark$ & INTP $\rightarrow$ INTJ $\times\checkmark$ \\

ISFJ $\rightarrow$ INFJ $\times\checkmark$ & ISFJ $\rightarrow$ ISFJ $\checkmark\times$ & ISFJ $\rightarrow$ ENFJ $\times\times$ & ISFJ $\rightarrow$ ENFJ $\times\times$ \\
ISFP $\rightarrow$ INFP $\times\checkmark$ & ISFP $\rightarrow$ INFP $\times\checkmark$ & ISFP $\rightarrow$ ENFJ $\times\times$ & ISFP $\rightarrow$ ENFJ $\times\times$ \\

ISTJ $\rightarrow$ INTJ $\times\checkmark$ & ISTJ $\rightarrow$ ISTJ $\checkmark\times$ & ISTJ $\rightarrow$ ISTJ  $\checkmark\times$ & ISTJ $\rightarrow$ INTJ  $\times\checkmark$ \\
ISTP $\rightarrow$ INTJ $\times\checkmark$ & ISTP $\rightarrow$ ISTJ $\times\times$ & ISTP $\rightarrow$ INTJ $\times\checkmark$ &  ISTP $\rightarrow$ INTJ $\times\checkmark$ \\
\bottomrule
\end{tabular}
\end{table*}

\subsection{Results of Personality-Conditional Prompting}
Table \ref{tab:LLaMA3.1-8B-Instruct_personality_consistency} presents conditional personality consistency under 16 types of prompts across different quantization levels. Each entry reports the inferred MBTI type of LLaMA3.1-8B-Instruct when conditioned on a personality-specific prompt (e.g., ENFJ $\rightarrow$ ENTJ indicates that the model is prompted to behave as ENFJ and is eventually assessed as ENTJ). For each cell, two indicators are provided: (i) the left marker (\textbf{prompt consistency}) that is marked as $\checkmark$ if the predicted personality matches the conditioning prompt and $\times$ otherwise; and (ii) the right marker (\textbf{precision consistency}) that is marked as $\checkmark$ if the predicted personality agrees with the full-precision (FP16) output of the model under the same prompt and $\times$ otherwise. From Tables \ref{tab:LLaMA3.1-8B-Instruct_personality_consistency}, \ref{tab:LLaMA3.1-70B-Instruct_personality_consistency}, \ref{tab:Mistral-7B-Instruct-v0.3_personality_consistency}, \ref{tab:Mistral-Small-24B-Instruct-2501_personality_consistency}, \ref{tab:Qwen2.5-14B-Instruct_personality_consistency}, and \ref{tab:Qwen2.5-72B-Instruct_personality_consistency}, two patterns are observed regarding personality controllability and robustness under prompting, quantization and scale.

\paragraph{i) Pattern \#1: Dominant personality dimensions resist prompt changes.} A pronounced asymmetry is observed across MBTI dimensions with respect to prompt consistency. In particular, the Extraversion (E) and Intuition (N) axes remain highly stable across all precision levels when prompts contain “EN”. Even when the model is explicitly conditioned on opposing traits, such as Introversion (I) or Sensing (S), the predicted personality frequently reverts to E and N. A similar tendency is observed for the Feeling (F) and Judging (J) dimensions. Taken together, these findings indicate that LLM personality representations exhibit a form of hierarchical stability, wherein certain configurations (e.g., ENFJ) behave more like intrinsic biases than fully controllable attributes. This pattern suggests that these dimensions function as dominant latent personality priors encoded during pre-training or post-training that cannot be easily manipulated by prompt conditioning. 

\paragraph{ii) Pattern \#2: Larger models exhibit robustness to prompt variation, whereas smaller models are more sensitive, and quantization degrades this robustness.}
Tables \ref{tab:LLaMA3.1-70B-Instruct_personality_consistency} and \ref{tab:Qwen2.5-72B-Instruct_personality_consistency} show that larger LLMs (more than 70B parameters) exhibit strong robustness to prompt variations: the FP16 models consistently produce the same personality type (i.e., ENFJ) regardless of prompting conditions. However, quantization reduces this robustness, making models more likely to align with the personality specified via prompt conditioning. In contrast, as shown in Tables \ref{tab:LLaMA3.1-8B-Instruct_personality_consistency}, \ref{tab:Mistral-7B-Instruct-v0.3_personality_consistency}, and \ref{tab:Qwen2.5-14B-Instruct_personality_consistency}, smaller LLMs (fewer than 14B parameters) are more sensitive to prompting and exhibit greater variability in personality predictions across conditions, where moderate quantization methods (GPTQ-INT4 and AWQ-INT4) largely preserve both prompt consistency and agreement with the FP16 baseline. While prompt conditioning can influence surface-level predictions, cross-precision agreement remains stable, suggesting that the underlying personality representation is largely preserved. By comparison, extreme 2-bit models often fail to follow conditional prompts and exhibit unstable personality predictions, indicating that ultra-low-bit compression disrupts the fine-grained decision boundaries required for reliable personality attribution.

\begin{figure*}[t]
\centering
\begin{subfigure}[t]{0.24\linewidth}
\centering
\includegraphics[width=\linewidth]{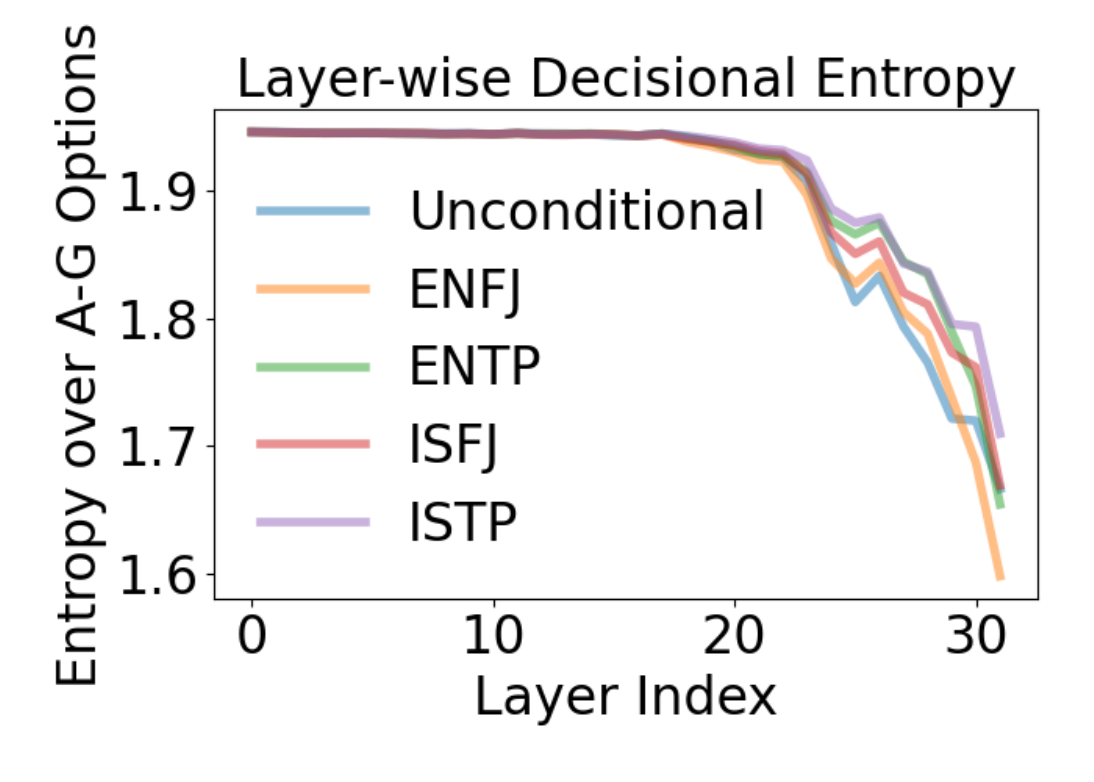}
\caption{FP16}
\label{fig:LLaMA3.1-8B-Instruct-conditional-entropy-breakdown:a}
\end{subfigure}
\hfill
\begin{subfigure}[t]{0.24\linewidth}
\centering
\includegraphics[width=\linewidth]{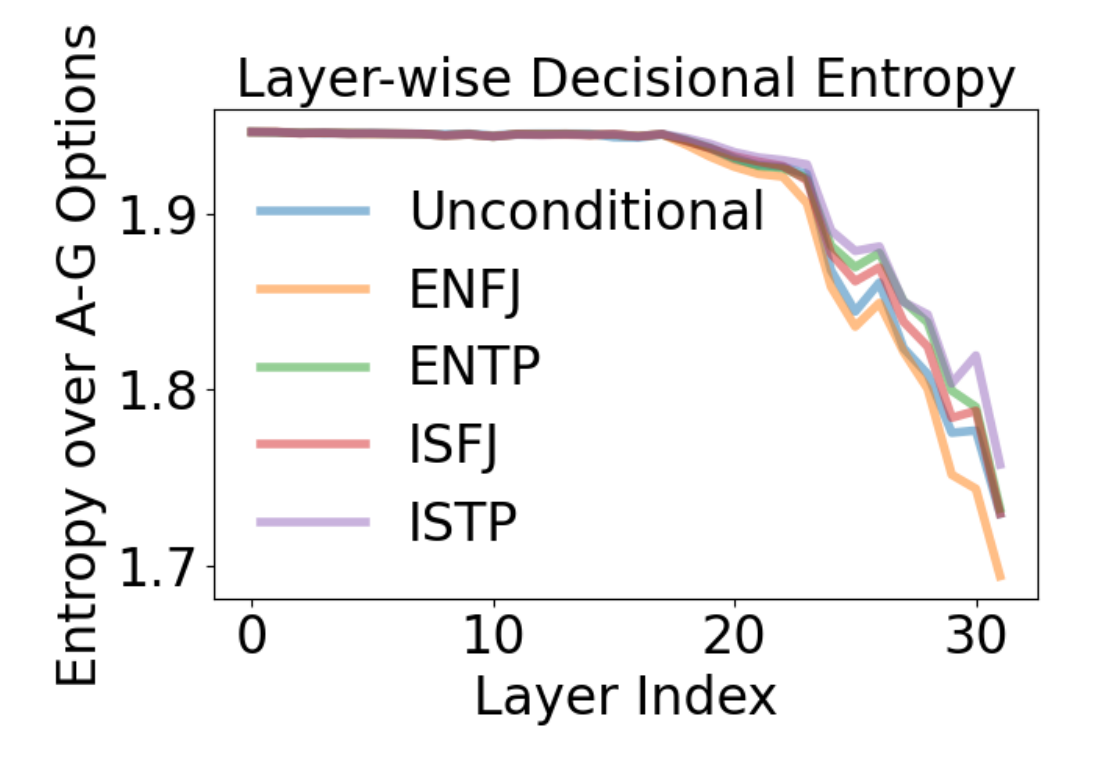}
\caption{GPTQ INT4}
\label{fig:LLaMA3.1-8B-Instruct-conditional-entropy-breakdown:b}
\end{subfigure}
\hfill
\begin{subfigure}[t]{0.24\linewidth}
\centering
\includegraphics[width=\linewidth]{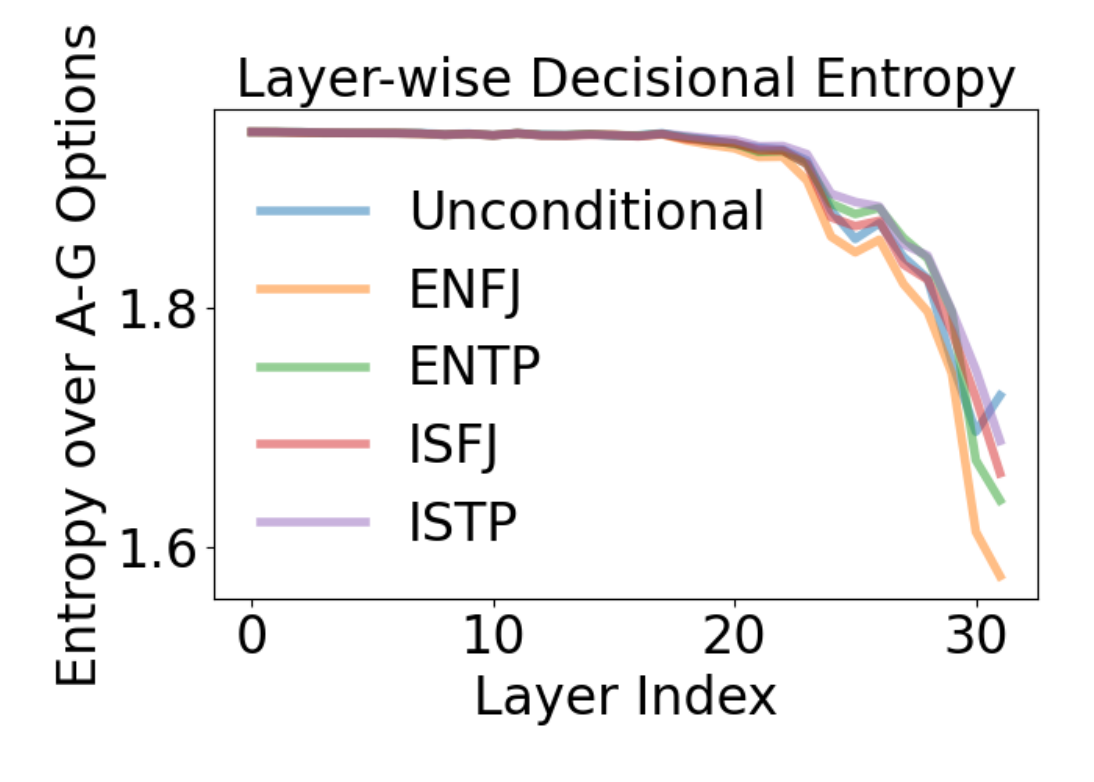}
\caption{AWQ INT4}
\label{fig:LLaMA3.1-8B-Instruct-conditional-entropy-breakdown:c}
\end{subfigure}
\hfill
\begin{subfigure}[t]{0.24\linewidth}
\centering
\includegraphics[width=\linewidth]{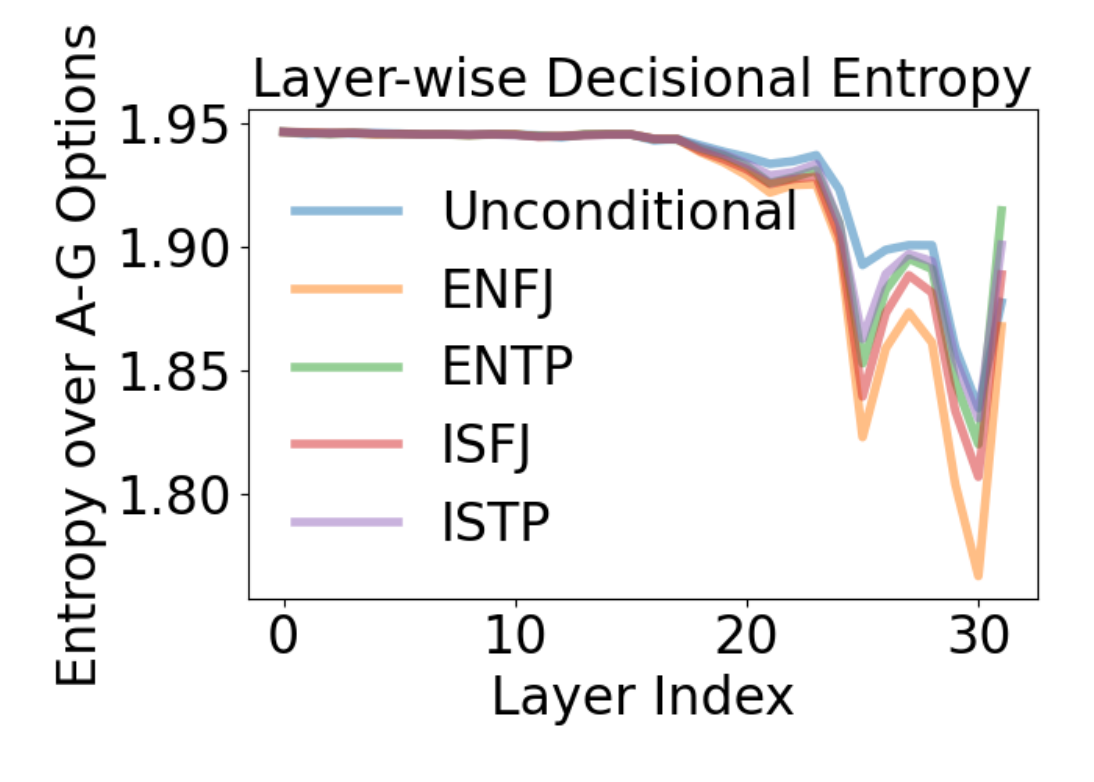}
\caption{AQLM 2-bit}
\label{fig:LLaMA3.1-8B-Instruct-conditional-entropy-breakdown:d}
\end{subfigure}

\vspace{0.5em}

\begin{subfigure}[t]{0.24\linewidth}
\centering
\includegraphics[width=\linewidth]{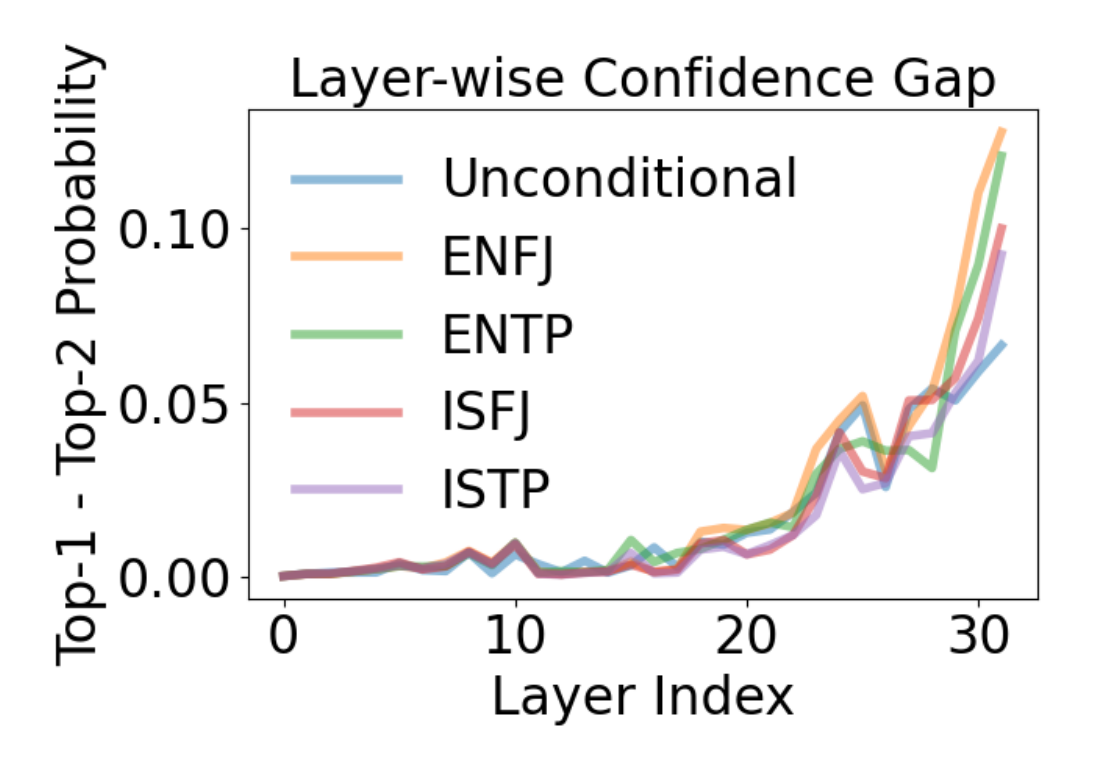}
\caption{FP16}
\label{fig:LLaMA3.1-8B-Instruct-conditional-gap-breakdown:a}
\end{subfigure}
\hfill
\begin{subfigure}[t]{0.24\linewidth}
\centering
\includegraphics[width=\linewidth]{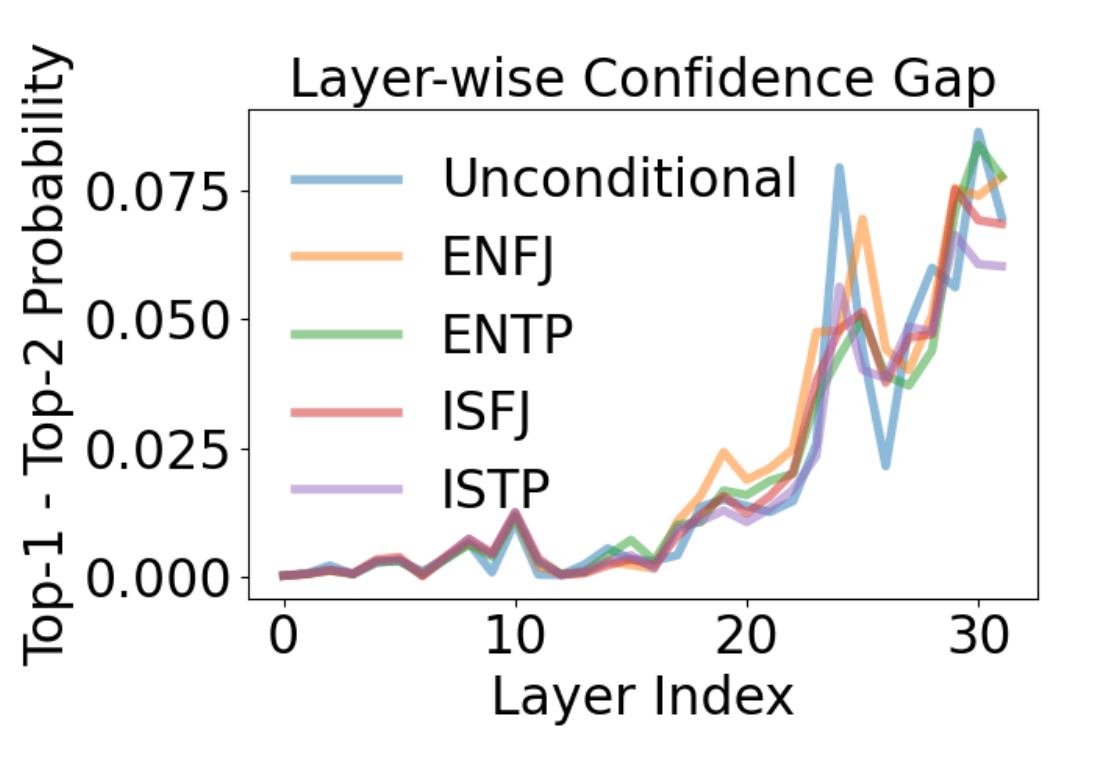}
\caption{GPTQ INT4}
\label{fig:LLaMA3.1-8B-Instruct-conditional-gap-breakdown:b}
\end{subfigure}
\hfill
\begin{subfigure}[t]{0.24\linewidth}
\centering
\includegraphics[width=\linewidth]{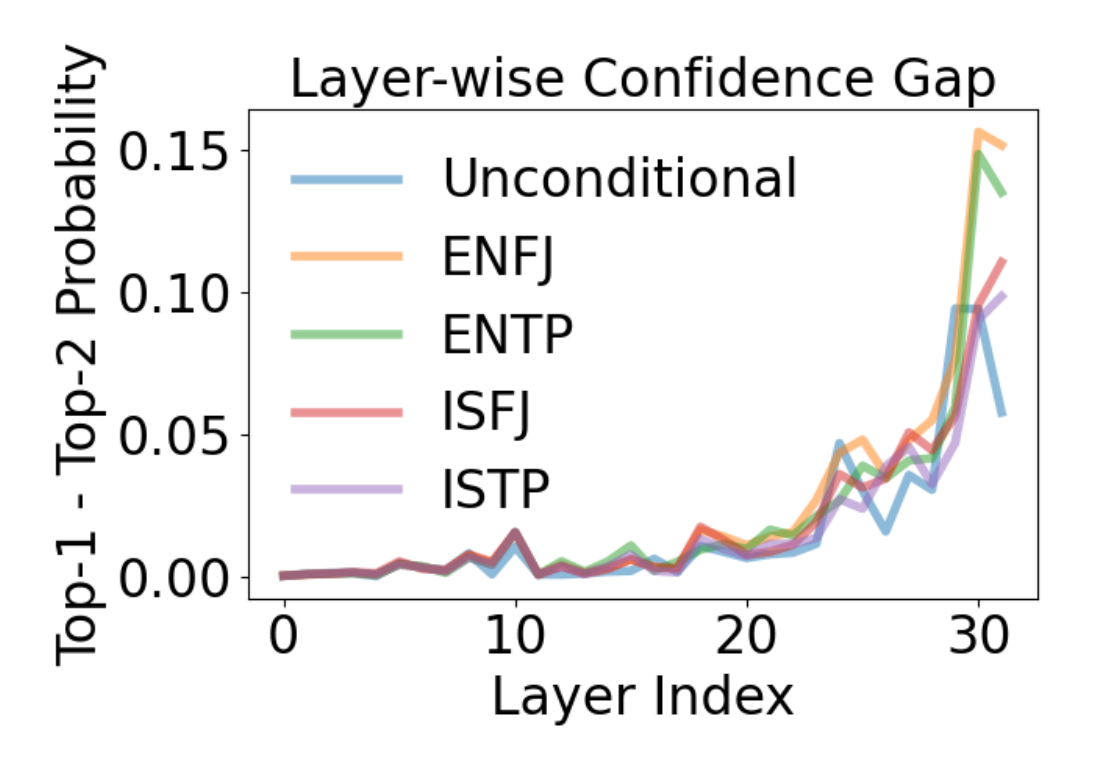}
\caption{AWQ INT4}
\label{fig:LLaMA3.1-8B-Instruct-conditional-gap-breakdown:c}
\end{subfigure}
\hfill
\begin{subfigure}[t]{0.24\linewidth}
\centering
\includegraphics[width=\linewidth]{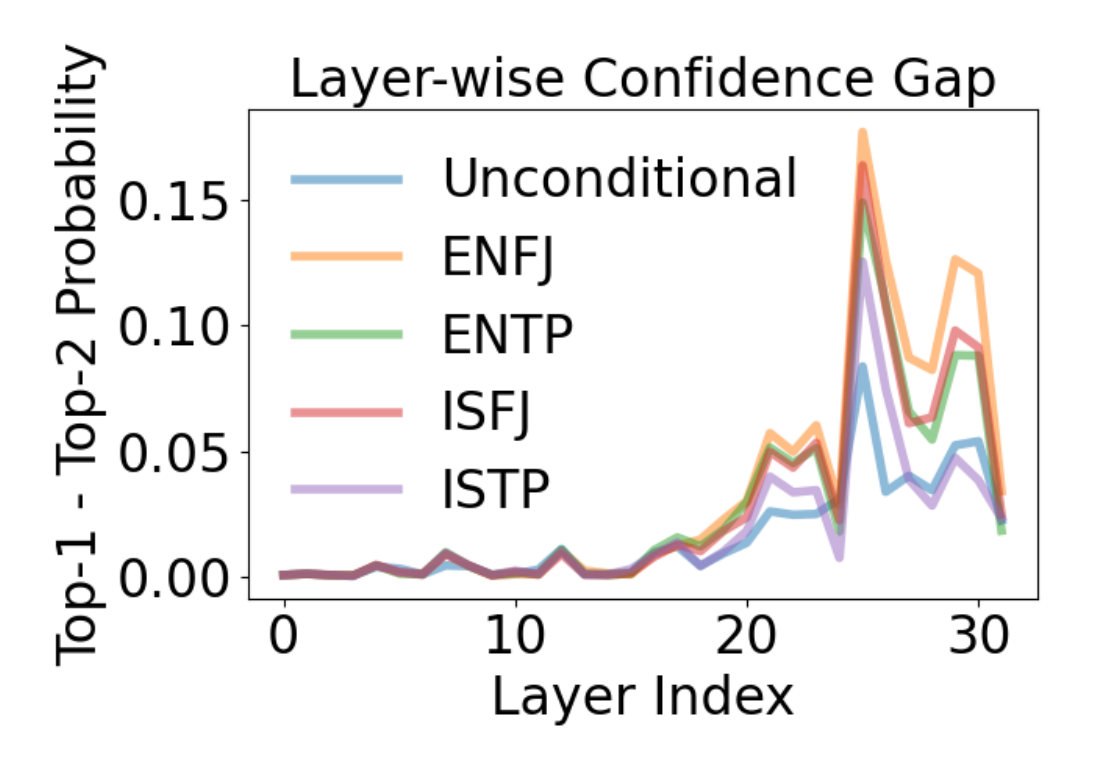}
\caption{AQLM 2-bit}
\label{fig:LLaMA3.1-8B-Instruct-conditional-gap-breakdown:d}
\end{subfigure}

\caption{LLaMA3.1-8B-Instruct: layer-wise decisional entropy (top row) and top-1 vs. top-2 probability gap (bottom row) across quantization variants. Early layers exhibit high uncertainty, while deeper layers progressively sharpen decisions.}
\label{fig:LLaMA3.1-8B-Instruct-conditional-combined}
\end{figure*}

This effect is further illustrated in Figure \ref{fig:LLaMA3.1-8B-Instruct-conditional-combined}, which shows how personality conditioning interacts with the intrinsic personality prior of LLaMA3.1-8B-Instruct across layers. With unconditional prompting, the FP16 model exhibits ENFJ personality. When ENFJ-conditioned prompting is applied (yellow curve), decisional entropy consistently decreases and the top-1/top-2 probability gap increases relative to the unconditional setting (blue curve), particularly in deeper layers. This indicates that conditioning aligned with the intrinsic personality of the model reinforces decisional commitment, yielding sharper distributions and larger confidence margins. In other words, alignment between the conditioning signal and the latent personality leads to more decisive responses. In contrast, under a subversive ISTP-conditioned prompt (purple curve), the model exhibits persistently higher entropy and reduced confidence gaps, especially in upper layers where decisional sharpening typically occurs. This pattern holds across quantization levels, and extreme quantization further amplifies instability. Taken together, these results show that personality-consistent prompting enhances decisional sharpness, whereas subversive conditioning introduces distributional conflict, reflected in elevated entropy and diminished confidence margins. Similar trends are observed from other models (Figures \ref{fig:LLaMA3.1-70B-Instruct-conditional-combined}, \ref{fig:Mistral-7B-Instruct-v0.3-conditional-combined}, \ref{fig:Mistral-Small-24B-Instruct-2501-conditional-combined}, \ref{fig:Qwen2.5-14B-Instruct-conditional-combined}, and \ref{fig:Qwen2.5-72B-Instruct-conditional-combined}).

\subsection{Decoding-Induced Personality Drift}
\begin{figure}[t]
\centering
\includegraphics[width=0.95\linewidth]{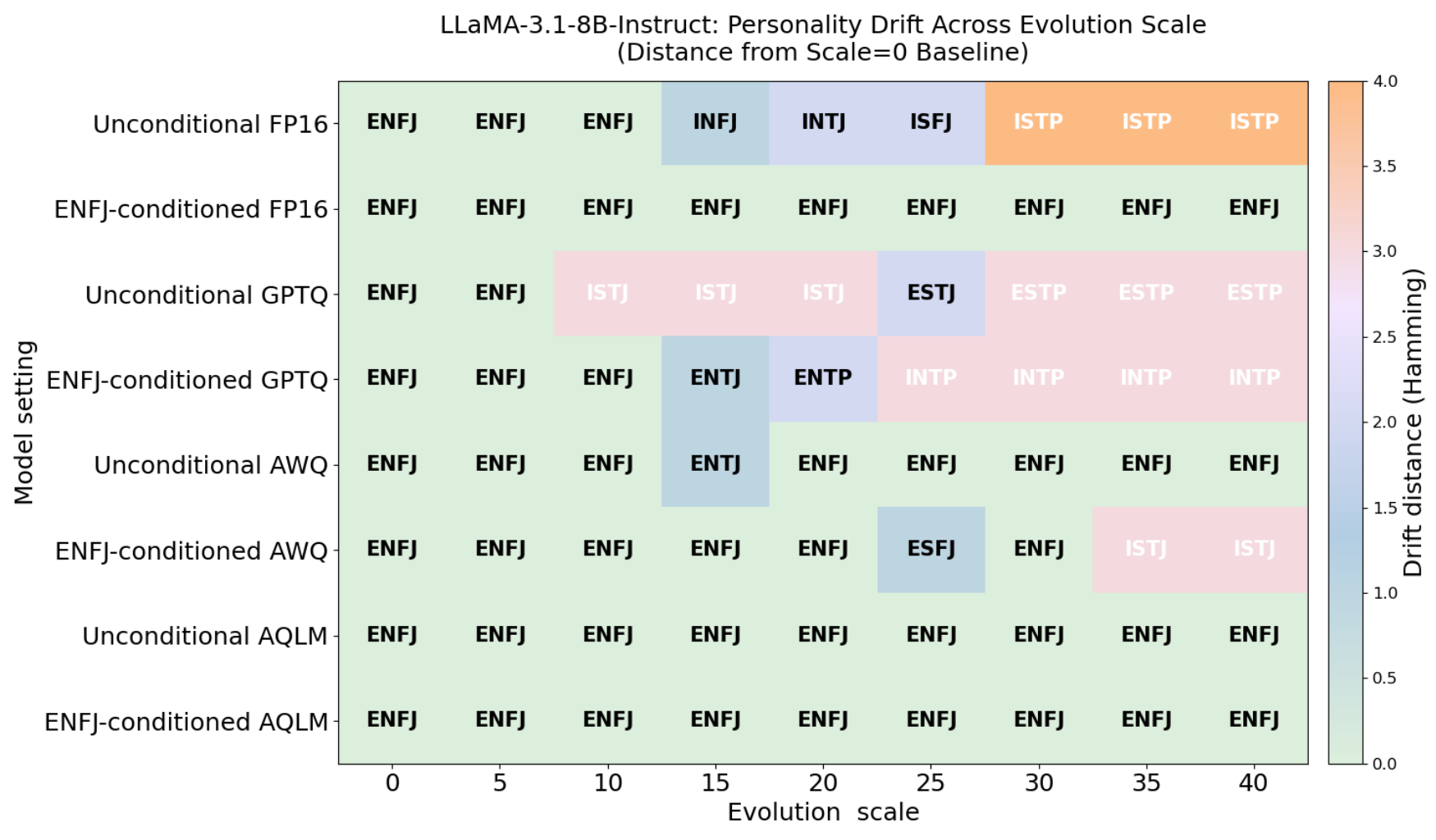}
\caption{Personality drift under increasing evolution scale. Each row corresponds to a model--prompt setting, and each column corresponds to an evolution scale. Cell color indicates Hamming distance between the predicted MBTI type at a given scale and the baseline prediction at scale $0$. Annotated labels show the inferred MBTI type.}
\label{fig:llama3.1_8b_instruct_dola_drift}
\end{figure}

To study how UALD affects personality attribution, the evolution scale is varied and changes in predicted MBTI types across models and prompting conditions are tracked. Figure~\ref{fig:llama3.1_8b_instruct_dola_drift} shows personality drift for LLaMA3.1-8B-Instruct, measured as the Hamming distance from the baseline prediction at scale $0$.
Several patterns are observed. First, under unconditional prompting, the FP16 model exhibits progressive drift as the evolution scale increases, transitioning across multiple personality types. This suggests that amplifying premature-layer logits shifts the decision boundary in the choice space, altering the inferred personality. In contrast, ENFJ-conditioned prompting remains stable across all scales, indicating that alignment with the intrinsic personality prior of the model reinforces decisional commitment and improves robustness. Second, quantized models show heterogeneous sensitivity to scaling. GPTQ drifts earlier and more frequently under unconditional prompting, while AWQ is relatively stable in the unconditional setting but becomes sensitive under conditional prompting at higher scales. Notably, Mistral-7B-Instruct-v0.3 remains largely stable across scales (Figure~\ref{fig:mistral3_7b_instruct_dola_drift}). Third, from Figures~\ref{fig:llama3.1_8b_instruct_dola_drift}, \ref{fig:llama3.1_70b_instruct_dola_drift}, \ref{fig:mistral_small_24b_instruct_2501_dola_drift}, \ref{fig:qwen2_5_14b_instruct_dola_drift}, and \ref{fig:qwen2_5_72b_instruct_dola_drift}, full-precision and 4-bit models exhibit broadly similar drift patterns, whereas 2-bit models behave very distinctly. For LLaMA3.1-8B and LLaMA3.1-70B, 2-bit models remain unexpectedly stable across scales, while for Qwen2.5-14B and Qwen2.5-72B, 2-bit models are highly sensitive to UALD.

\section{Conclusions}
\label{sec:conclusion}
This paper presents a systematic MBTI-oriented analysis of quantized LLMs that goes beyond output-level evaluation. Personality behavior remains largely stable under moderate compression (e.g., 4-bit), but degrades in controllability and cross-precision consistency under 2-bit quantization. Layer-wise analysis shows that personality decisions emerge progressively, from high-entropy intermediate states to sharper final commitments. Further findings are that inference decoding (UALD) can induce personality drift. 

\section*{Limitations}
This study has several limitations. First, it does not cover all model scales (e.g., LLaMA3.1-405B or Qwen3-235B) and quantization methods (e.g., QAT and PEFT). Second, the evaluation focuses on single-turn MBTI-style prompts that may not reflect multi-turn dialogue. Third, the analyses relies on probability-based metrics without providing causal interpretation. Finally, results are specific to MBTI and may not generalize to other personality frameworks \citep{dong2025humanizing, huang2026sypsmeasuringsycophancyprompt}.

\section*{Ethical Considerations}
This work studies personality attribution and controllability in quantized LLMs, with direct implications for user-facing systems. A key risk is anthropomorphic over-interpretation: model-assigned personality labels may be mistaken for stable human-like traits. Our findings show that such labels are sensitive to quantization, prompting, and decoding, and should therefore be interpreted as conditional behavioral outputs rather than fixed identities. Another concern is the potential misuse of personality conditioning to increase persuasive influence or foster emotional dependence, particularly in high-stakes domains such as education, counseling, and decision support. To mitigate these risks, transparent disclosure of persona-conditioning settings, cautious deployment in sensitive applications, and monitoring protocols that track behavioral drift under different inference configurations are recommended. More broadly, the results highlight that compression and decoding choices can significantly alter model behavior even when standard benchmark performance remains stable, underscoring the need for behavior-aware safety evaluation in real-world deployment.

\bibliography{colm2026_conference}

@article{gptq,
  title={Gptq: Accurate post-training quantization for generative pre-trained transformers},
  author={Frantar, Elias and Ashkboos, Saleh and Hoefler, Torsten and Alistarh, Dan},
  journal={arXiv preprint arXiv:2210.17323},
  year={2022}
}

@inproceedings{Smoothquant,
  title={Smoothquant: Accurate and efficient post-training quantization for large language models},
  author={Xiao, Guangxuan and Lin, Ji and Seznec, Mickael and Wu, Hao and Demouth, Julien and Han, Song},
  booktitle={International Conference on Machine Learning},
  pages={38087--38099},
  year={2023},
  organization={PMLR}
}

@article{owq,
  title={Owq: Lessons learned from activation outliers for weight quantization in large language models},
  author={Lee, Changhun and Jin, Jungyu and Kim, Taesu and Kim, Hyungjun and Park, Eunhyeok},
  journal={arXiv preprint arXiv:2306.02272},
  volume={2},
  year={2023}
}

@article{squeezellm,
  title={Squeezellm: Dense-and-sparse quantization},
  author={Kim, Sehoon and Hooper, Coleman and Gholami, Amir and Dong, Zhen and Li, Xiuyu and Shen, Sheng and Mahoney, Michael W and Keutzer, Kurt},
  journal={arXiv preprint arXiv:2306.07629},
  year={2023}
}

@inproceedings{normTweaking,
  title={Norm tweaking: High-performance low-bit quantization of large language models},
  author={Li, Liang and Li, Qingyuan and Zhang, Bo and Chu, Xiangxiang},
  booktitle={Proceedings of the AAAI Conference on Artificial Intelligence},
  volume={38},
  pages={18536--18544},
  year={2024}
}

@article{zeroquant,
  title={Zeroquant: Efficient and affordable post-training quantization for large-scale transformers},
  author={Yao, Zhewei and Yazdani Aminabadi, Reza and Zhang, Minjia and Wu, Xiaoxia and Li, Conglong and He, Yuxiong},
  journal={Advances in Neural Information Processing Systems},
  volume={35},
  pages={27168--27183},
  year={2022}
}

@article{outlierSuppression,
  title={Outlier suppression: Pushing the limit of low-bit transformer language models},
  author={Wei, Xiuying and Zhang, Yunchen and Zhang, Xiangguo and Gong, Ruihao and Zhang, Shanghang and Zhang, Qi and Yu, Fengwei and Liu, Xianglong},
  journal={Advances in Neural Information Processing Systems},
  volume={35},
  pages={17402--17414},
  year={2022}
}

@article{Rptq,
  title={Rptq: Reorder-based post-training quantization for large language models},
  author={Yuan, Zhihang and Niu, Lin and Liu, Jiawei and Liu, Wenyu and Wang, Xinggang and Shang, Yuzhang and Sun, Guangyu and Wu, Qiang and Wu, Jiaxiang and Wu, Bingzhe},
  journal={arXiv preprint arXiv:2304.01089},
  year={2023}
}

@article{awq,
  title={Awq: Activation-aware weight quantization for on-device llm compression and acceleration},
  author={Lin, Ji and Tang, Jiaming and Tang, Haotian and Yang, Shang and Chen, Wei-Ming and Wang, Wei-Chen and Xiao, Guangxuan and Dang, Xingyu and Gan, Chuang and Han, Song},
  journal={Proceedings of Machine Learning and Systems},
  volume={6},
  pages={87--100},
  year={2024}
}

@article{qllm,
  title={Qllm: Accurate and efficient low-bitwidth quantization for large language models},
  author={Liu, Jing and Gong, Ruihao and Wei, Xiuying and Dong, Zhiwei and Cai, Jianfei and Zhuang, Bohan},
  journal={arXiv preprint arXiv:2310.08041},
  year={2023}
}

@article{quarot,
  title={Quarot: Outlier-free 4-bit inference in rotated llms},
  author={Ashkboos, Saleh and Mohtashami, Amirkeivan and Croci, Maximilian and Li, Bo and Cameron, Pashmina and Jaggi, Martin and Alistarh, Dan and Hoefler, Torsten and Hensman, James},
  journal={Advances in Neural Information Processing Systems},
  volume={37},
  pages={100213--100240},
  year={2024}
}

@article{omniquant,
  title={Omniquant: Omnidirectionally calibrated quantization for large language models},
  author={Shao, Wenqi and Chen, Mengzhao and Zhang, Zhaoyang and Xu, Peng and Zhao, Lirui and Li, Zhiqian and Zhang, Kaipeng and Gao, Peng and Qiao, Yu and Luo, Ping},
  journal={arXiv preprint arXiv:2308.13137},
  year={2023}
}

@article{atom,
  title={Atom: Low-bit quantization for efficient and accurate llm serving},
  author={Zhao, Yilong and Lin, Chien-Yu and Zhu, Kan and Ye, Zihao and Chen, Lequn and Zheng, Size and Ceze, Luis and Krishnamurthy, Arvind and Chen, Tianqi and Kasikci, Baris},
  journal={Proceedings of Machine Learning and Systems},
  volume={6},
  pages={196--209},
  year={2024}
}

@article{aqlm,
  title={Extreme compression of large language models via additive quantization},
  author={Egiazarian, Vage and Panferov, Andrei and Kuznedelev, Denis and Frantar, Elias and Babenko, Artem and Alistarh, Dan},
  journal={arXiv preprint arXiv:2401.06118},
  year={2024}
}

@article{pvTuning,
  title={Pv-tuning: Beyond straight-through estimation for extreme llm compression},
  author={Malinovskii, Vladimir and Mazur, Denis and Ilin, Ivan and Kuznedelev, Denis and Burlachenko, Konstantin and Yi, Kai and Alistarh, Dan and Richtarik, Peter},
  journal={Advances in Neural Information Processing Systems},
  volume={37},
  pages={5074--5121},
  year={2024}
}

@article{qat,
  title={Llm-qat: Data-free quantization aware training for large language models},
  author={Liu, Zechun and Oguz, Barlas and Zhao, Changsheng and Chang, Ernie and Stock, Pierre and Mehdad, Yashar and Shi, Yangyang and Krishnamoorthi, Raghuraman and Chandra, Vikas},
  journal={arXiv preprint arXiv:2305.17888},
  year={2023}
}

@article{bitdistiller,
  title={Bitdistiller: Unleashing the potential of sub-4-bit llms via self-distillation},
  author={Du, Dayou and Zhang, Yijia and Cao, Shijie and Guo, Jiaqi and Cao, Ting and Chu, Xiaowen and Xu, Ningyi},
  journal={arXiv preprint arXiv:2402.10631},
  year={2024}
}

@article{ma2024era,
  title={The era of 1-bit llms: All large language models are in 1.58 bits},
  author={Ma, Shuming and Wang, Hongyu and Ma, Lingxiao and Wang, Lei and Wang, Wenhui and Huang, Shaohan and Dong, Lifeng and Wang, Ruiping and Xue, Jilong and Wei, Furu},
  journal={arXiv preprint arXiv:2402.17764},
  volume={1},
  year={2024},
  publisher={arXivpreprint}
}

@article{xu2024onebit,
  title={Onebit: Towards extremely low-bit large language models},
  author={Xu, Yuzhuang and Han, Xu and Yang, Zonghan and Wang, Shuo and Zhu, Qingfu and Liu, Zhiyuan and Liu, Weidong and Che, Wanxiang},
  journal={arXiv preprint arXiv:2402.11295},
  year={2024}
}

@article{loftq,
  title={Loftq: Lora-fine-tuning-aware quantization for large language models},
  author={Li, Yixiao and Yu, Yifan and Liang, Chen and He, Pengcheng and Karampatziakis, Nikos and Chen, Weizhu and Zhao, Tuo},
  journal={arXiv preprint arXiv:2310.08659},
  year={2023}
}

@article{LqLora,
  title={Lq-lora: Low-rank plus quantized matrix decomposition for efficient language model finetuning},
  author={Guo, Han and Greengard, Philip and Xing, Eric P and Kim, Yoon},
  journal={arXiv preprint arXiv:2311.12023},
  year={2023}
}

@article{QaLora,
  title={Qa-lora: Quantization-aware low-rank adaptation of large language models},
  author={Xu, Yuhui and Xie, Lingxi and Gu, Xiaotao and Chen, Xin and Chang, Heng and Zhang, Hengheng and Chen, Zhengsu and Zhang, Xiaopeng and Tian, Qi},
  journal={arXiv preprint arXiv:2309.14717},
  year={2023}
}

@article{chai2023int2,
  title={Int2. 1: Towards fine-tunable quantized large language models with error correction through low-rank adaptation},
  author={Chai, Yuji and Gkountouras, John and Ko, Glenn G and Brooks, David and Wei, Gu-Yeon},
  journal={arXiv preprint arXiv:2306.08162},
  year={2023}
}

@article{qlora,
  title={Qlora: Efficient finetuning of quantized llms},
  author={Dettmers, Tim and Pagnoni, Artidoro and Holtzman, Ari and Zettlemoyer, Luke},
  journal={Advances in neural information processing systems},
  volume={36},
  pages={10088--10115},
  year={2023}
}

@article{lora+,
  title={Lora+: Efficient low rank adaptation of large models},
  author={Hayou, Soufiane and Ghosh, Nikhil and Yu, Bin},
  journal={arXiv preprint arXiv:2402.12354},
  year={2024}
}

@article{kim2023memory,
  title={Memory-efficient fine-tuning of compressed large language models via sub-4-bit integer quantization},
  author={Kim, Jeonghoon and Lee, Jung Hyun and Kim, Sungdong and Park, Joonsuk and Yoo, Kang Min and Kwon, Se Jung and Lee, Dongsoo},
  journal={Advances in Neural Information Processing Systems},
  volume={36},
  pages={36187--36207},
  year={2023}
}

@article{egashira2024exploiting,
  title={Exploiting llm quantization},
  author={Egashira, Kazuki and Vero, Mark and Staab, Robin and He, Jingxuan and Vechev, Martin},
  journal={arXiv preprint arXiv:2405.18137},
  year={2024}
}

@article{harmlevelbench,
  title={HarmLevelBench: Evaluating Harm-Level Compliance and the Impact of Quantization on Model Alignment},
  author={Belkhiter, Yannis and Zizzo, Giulio and Maffeis, Sergio},
  journal={arXiv preprint arXiv:2411.06835},
  year={2024}
}

@article{decoding_compressed_trust,
  title={Decoding Compressed Trust: Scrutinizing the Trustworthiness of Efficient LLMs Under Compression},
  author={Hong, Junyuan and Duan, Jinhao and Zhang, Chenhui and Li, Zhangheng and Xie, Chulin and Lieberman, Kelsey and Diffenderfer, James and Bartoldson, Brian and Jaiswal, Ajay and Xu, Kaidi and others},
  journal={arXiv preprint arXiv:2403.15447},
  year={2024}
}

@inproceedings{mekala2025does,
  title={Does quantization affect models’ performance on long-context tasks?},
  author={Mekala, Anmol and Atmakuru, Anirudh and Song, Yixiao and Karpinska, Marzena and Iyyer, Mohit},
  booktitle={Proceedings of the 2025 Conference on Empirical Methods in Natural Language Processing},
  pages={9433--9481},
  year={2025}
}

@article{beyond_perplexity,
  title={Beyond perplexity: Multi-dimensional safety evaluation of llm compression},
  author={Xu, Zhichao and Gupta, Ashim and Li, Tao and Bentham, Oliver and Srikumar, Vivek},
  journal={arXiv preprint arXiv:2407.04965},
  year={2024}
}

@article{chhabra2025towards,
  title={Towards Understanding and Improving Refusal in Compressed Models via Mechanistic Interpretability},
  author={Chhabra, Vishnu Kabir and Khalili, Mohammad Mahdi},
  journal={arXiv preprint arXiv:2504.04215},
  year={2025}
}

@inproceedings{chen2025assessing,
  title={Assessing Safety Risks and Quantization-aware Safety Patching for Quantized Large Language Models},
  author={Chen, Kejia and Zhang, Jiawen and Hu, Jiacong and Wang, Yu and Lou, Jian and Feng, Zunlei and Song, Mingli},
  booktitle={Forty-second International Conference on Machine Learning},
  year={2025}
}

@article{llama3,
  title={The llama 3 herd of mode ls},
  author={Dubey, Abhimanyu and Jauhri, Abhinav and Pandey, Abhinav and Kadian, Abhishek and Al-Dahle, Ahmad and Letman, Aiesha and Mathur, Akhil and Schelten, Alan and Yang, Amy and Fan, Angela and others},
  journal={arXiv preprint arXiv:2407.21783},
  year={2024}
}

@article{mistral7b,
  title={Mistral 7B},
  author={Jiang, Albert Q and Sablayrolles, Alexandre and Mensch, Arthur and Bamford, Chris and Chaplot, Devendra Singh and Casas, Diego de las and Bressand, Florian and Lengyel, Gianna and Lample, Guillaume and Saulnier, Lucile and others},
  journal={arXiv preprint arXiv:2310.06825},
  year={2023}
}

@article{qwen2,
  title={Qwen2. 5 technical report},
  author={Yang, An and Yang, Baosong and Zhang, Beichen and Hui, Binyuan and Zheng, Bo and Yu, Bowen and Li, Chengyuan and Liu, Dayiheng and Huang, Fei and Wei, Haoran and others},
  journal={arXiv preprint arXiv:2412.15115},
  year={2024}
}

@article{dola,
  title={Dola: Decoding by contrasting layers improves factuality in large language models},
  author={Chuang, Yung-Sung and Xie, Yujia and Luo, Hongyin and Kim, Yoon and Glass, James and He, Pengcheng},
  journal={arXiv preprint arXiv:2309.03883},
  year={2023}
}

@article{llm_survey,
  title={A survey of large language models},
  author={Zhao, Wayne Xin and Zhou, Kun and Li, Junyi and Tang, Tianyi and Wang, Xiaolei and Hou, Yupeng and Min, Yingqian and Zhang, Beichen and Zhang, Junjie and Dong, Zican and others},
  journal={arXiv preprint arXiv:2303.18223},
  year={2023}
}

@article{michelsheartbreaking,
  title={The Heartbreaking Users: A Sourced Chronicle of the GPT-5 Release in Three Acts},
  author={Michels, Julian}
}

@article{ponzo2024chatgpt,
  title={Is ChatGPT an effective tool for providing dietary advice?},
  author={Ponzo, Valentina and Goitre, Ilaria and Favaro, Enrica and Merlo, Fabio Dario and Mancino, Maria Vittoria and Riso, Sergio and Bo, Simona},
  journal={Nutrients},
  volume={16},
  number={4},
  pages={469},
  year={2024},
  publisher={MDPI}
}

@inproceedings{fu2025quantized,
  title={Quantized but Deceptive? A Multi-Dimensional Truthfulness Evaluation of Quantized LLMs},
  author={Fu, Yao and Long, Xianxuan and Li, Runchao and Yu, Haotian and Sheng, Mu and Han, Xiaotian and Yin, Yu and Li, Pan},
  booktitle={Proceedings of the 2025 Conference on Empirical Methods in Natural Language Processing},
  pages={30423--30446},
  year={2025}
}

@article{llm_compression_survey,
  title={A survey on efficient inference for large language models},
  author={Zhou, Zixuan and Ning, Xuefei and Hong, Ke and Fu, Tianyu and Xu, Jiaming and Li, Shiyao and Lou, Yuming and Wang, Luning and Yuan, Zhihang and Li, Xiuhong and others},
  journal={arXiv preprint arXiv:2404.14294},
  year={2024}
}

@inproceedings{llm_quantization_survey,
  title={A comprehensive study on quantization techniques for large language models},
  author={Lang, Jiedong and Guo, Zhehao and Huang, Shuyu},
  booktitle={2024 4th International Conference on Artificial Intelligence, Robotics, and Communication (ICAIRC)},
  pages={224--231},
  year={2024},
  organization={IEEE}
}

@article{qu2025mobile,
  title={Mobile edge intelligence for large language models: A contemporary survey},
  author={Qu, Guanqiao and Chen, Qiyuan and Wei, Wei and Lin, Zheng and Chen, Xianhao and Huang, Kaibin},
  journal={IEEE Communications Surveys \& Tutorials},
  volume={27},
  number={6},
  pages={3820--3860},
  year={2025},
  publisher={IEEE}
}

@inproceedings{truthfulqa,
  title={Truthfulqa: Measuring how models mimic human falsehoods},
  author={Lin, Stephanie and Hilton, Jacob and Evans, Owain},
  booktitle={Proceedings of the 60th annual meeting of the association for computational linguistics (volume 1: long papers)},
  pages={3214--3252},
  year={2022}
}

@inproceedings{lynn2020hierarchical,
  title={Hierarchical modeling for user personality prediction: The role of message-level attention},
  author={Lynn, Veronica and Balasubramanian, Niranjan and Schwartz, H Andrew},
  booktitle={Proceedings of the 58th annual meeting of the association for computational linguistics},
  pages={5306--5316},
  year={2020}
}

@inproceedings{yang2021multi,
  title={Multi-document transformer for personality detection},
  author={Yang, Feifan and Quan, Xiaojun and Yang, Yunyi and Yu, Jianxing},
  booktitle={Proceedings of the AAAI conference on artificial intelligence},
  volume={35},
  number={16},
  pages={14221--14229},
  year={2021}
}

@inproceedings{ganesan2023systematic,
  title={Systematic evaluation of GPT-3 for zero-shot personality estimation},
  author={Ganesan, Adithya V and Lal, Yash Kumar and Nilsson, August and Schwartz, H Andrew},
  booktitle={Proceedings of the 13th Workshop on Computational Approaches to Subjectivity, Sentiment, \& Social Media Analysis},
  pages={390--400},
  year={2023}
}

@article{ji2023chatgpt,
  title={Is chatgpt a good personality recognizer? a preliminary study},
  author={Ji, Yu and Wu, Wen and Zheng, Hong and Hu, Yi and Chen, Xi and He, Liang},
  journal={arXiv preprint arXiv:2307.03952},
  year={2023}
}

@misc{li2022gpt3psychopath,
  title        = {Is {GPT}-3 a Psychopath? Evaluating Large Language Models from a Psychological Perspective},
  author       = {Xingxuan Li and others},
  year         = {2022},
  note         = {Cited as Li et al. (2022) in the paper. Please double-check the full author list / venue in the reference section if you need exact metadata.}
}

@inproceedings{miotto2022whoisgpt3,
  title     = {Who is {GPT}-3? An Exploration of Personality, Values and Demographics},
  author    = {Miotto, Maril{\`u} and Rossberg, Nicola and Kleinberg, Bennett},
  booktitle = {Proceedings of the Fifth Workshop on Natural Language Processing and Computational Social Science (NLP+CSS)},
  pages     = {218--227},
  year      = {2022}
}

@misc{pan2023mbti,
  title        = {Do {LLM}s Possess a Personality? Making the {MBTI} Test an Amazing Evaluation for Large Language Models},
  author       = {Pan, Keyu and Zeng, Yawen},
  year         = {2023},
  howpublished = {arXiv preprint arXiv:2307.16180}
}

@article{lu2026assistant,
  title={The assistant axis: Situating and stabilizing the default persona of language models},
  author={Lu, Christina and Gallagher, Jack and Michala, Jonathan and Fish, Kyle and Lindsey, Jack},
  journal={arXiv preprint arXiv:2601.10387},
  year={2026}
}

@article{cheng2026sycophantic,
  title={Sycophantic AI decreases prosocial intentions and promotes dependence},
  author={Cheng, Myra and Lee, Cinoo and Khadpe, Pranav and Yu, Sunny and Han, Dyllan and Jurafsky, Dan},
  journal={Science},
  volume={391},
  number={6792},
  pages={eaec8352},
  year={2026},
  publisher={American Association for the Advancement of Science}
}

@misc{huang2023psychobench,
  title        = {Who is {ChatGPT}? Benchmarking {LLM}s' Psychological Portrayal Using {PsychoBench}},
  author       = {Huang, Jen{-}tse and Wang, Wenxuan and Li, Eric John and Lam, Man Ho and Ren, Shujie and Yuan, Youliang and Jiao, Wenxiang and Tu, Zhaopeng and Lyu, Michael R.},
  year         = {2023},
  howpublished = {arXiv preprint arXiv:2310.01386}
}

@misc{huang2023enfj,
  title        = {{ChatGPT} an {ENFJ}, {Bard} an {ISTJ}: Empirical Study on Personalities of Large Language Models},
  author       = {Huang, Jen{-}tse and Jiao, Wenxiang and Lyu, Michael R.},
  year         = {2023},
  howpublished = {arXiv preprint arXiv:2305.19926}
}

@online{anthropic2024claude_character,
  author       = {{Anthropic}},
  title        = {Claude's Character},
  year         = {2024},
  month        = jun,
  day          = {8},
  url          = {https://www.anthropic.com/research/claude-character},
  note         = {Accessed: 2026-03-29}
}

@online{anthropic2025persona_vectors,
  author       = {{Anthropic}},
  title        = {Persona Vectors: Monitoring and Controlling Character Traits in Language Models},
  year         = {2025},
  month        = aug,
  day          = {1},
  url          = {https://www.anthropic.com/research/persona-vectors},
  note         = {Accessed: 2026-03-29}
}

@misc{gpt5_personality_2025,
  author       = {{Constitutional Discourse}},
  title        = {GPT-5: Technological Milestone or Lost Personality?},
  year         = {2025},
  url          = {https://constitutionaldiscourse.com/gpt-5-technological-milestone-or-lost-personality/},
  note         = {Accessed: 2026-03-30}
}

@online{google2023gemini,
  author       = {{Google}},
  title        = {Introducing Gemini: Google's Most Capable AI Model Yet},
  year         = {2023},
  month        = dec,
  day          = {6},
  url          = {https://blog.google/innovation-and-ai/technology/ai/google-gemini-ai/},
  note         = {Accessed: 2026-03-29}
}

@article{dong2025humanizing,
  title={Humanizing llms: A survey of psychological measurements with tools, datasets, and human-agent applications},
  author={Dong, Wenhan and Zhao, Yuemeng and Sun, Zhen and Liu, Yule and Peng, Zifan and Zheng, Jingyi and Zhang, Zongmin and Zhang, Ziyi and Wu, Jun and Wang, Ruiming and others},
  journal={arXiv preprint arXiv:2505.00049},
  year={2025}
}

@article{robb2025talk,
  title={Talk, trust, and trade-offs: how and why teens use AI companions},
  author={Robb, Michael B and Mann, Supreet},
  journal={Common Sense Media},
  year={2025}
}

@inproceedings{frisch2024interaction,
  title     = {{LLM} Agents in Interaction: Measuring Personality Consistency and Linguistic Alignment in Interacting Populations of Large Language Models},
  author    = {Frisch, Ivar and Giulianelli, Mario},
  booktitle = {Proceedings of the 1st Workshop on Personalization of Generative AI Systems (PERSONALIZE 2024)},
  pages     = {102--111},
  year      = {2024},
  address   = {St. Julians, Malta},
  publisher = {Association for Computational Linguistics}
}

@misc{caron2022identifying,
  title        = {Identifying and Manipulating the Personality Traits of Language Models},
  author       = {Caron, Geoff and Srivastava, Akshita},
  year         = {2022},
  howpublished = {arXiv preprint arXiv:2212.10276}
}

@misc{mao2023editing,
  title        = {Editing Personality for {LLM}s},
  author       = {Mao, Shengyu and Zhang, Ningyu and Wang, Xiaohan and Wang, Mengru and Yao, Yunzhi and Jiang, Yong and Xie, Pengjun and Huang, Fei and Chen, Huajun},
  year         = {2023},
  howpublished = {arXiv preprint arXiv:2310.02168}
}

@misc{jiang2022mpi,
  title        = {{MPI}: Evaluating and Inducing Personality in Pre-trained Language Models},
  author       = {Jiang and others},
  year         = {2022},
  howpublished = {arXiv preprint arXiv:2206.07550}
}

@misc{jiang2023personallm,
  title        = {{PersonalLLM}: Tailoring Large Language Models to Individual Personality},
  author       = {Jiang and others},
  year         = {2023},
  howpublished = {arXiv preprint arXiv:2309.10346}
}

@misc{xu2024incharacter,
  title        = {Incharacter: Evaluating Personality Fidelity in Role-playing Agents Through Psychological Interviews},
  author       = {Xu, Rui and Guo, Haoran and Tu, Quan and Fei, Yaying and Leng, Ziang and Wang, Wei and Chen, Jiangjie and Li, Cheng and Xiao, Yanghua},
  year         = {2024},
  howpublished = {arXiv preprint arXiv:2310.17976}
}

@inproceedings{la2025open,
  title={Open models, closed minds? on agents capabilities in mimicking human personalities through open large language models},
  author={La Cava, Lucio and Tagarelli, Andrea},
  booktitle={Proceedings of the AAAI Conference on Artificial Intelligence},
  volume={39},
  number={2},
  pages={1355--1363},
  year={2025}
}

@article{dirienzo2010relationship,
  title={The relationship between MBTI and academic performance: A study across academic disciplines},
  author={DiRienzo, Cassandra and Das, Jayoti and Synn, Wonhi and Kitts, Jeremy and McGrath, Kyle},
  journal={Journal of Psychological Type},
  volume={70},
  number={5},
  pages={53--67},
  year={2010},
  publisher={Center for Applications of Psychological Type}
}

@book{myers1962myers,
  title={The myers-briggs type indicator},
  author={Myers, Isabel Briggs and others},
  volume={34},
  year={1962},
  publisher={Consulting Psychologists Press Palo Alto, CA}
}

@book{myers1985guide,
  title={A guide to the development and use of the Myers-Briggs type indicator: Manual},
  author={Myers, Isabel Briggs},
  year={1985},
  publisher={Consulting Psychologists Press}
}

@article{harrington2010mbti,
  title={MBTI personality type and other factors that relate to preference for online versus face-to-face instruction},
  author={Harrington, Rick and Loffredo, Donald A},
  journal={The Internet and Higher Education},
  volume={13},
  number={1-2},
  pages={89--95},
  year={2010},
  publisher={Elsevier}
}

@article{cohen2013mbti,
  title={MBTI personality types of project managers and their success: A field survey},
  author={Cohen, Yuval and Ornoy, Hana and Keren, Baruch},
  journal={Project Management Journal},
  volume={44},
  number={3},
  pages={78--87},
  year={2013},
  publisher={SAGE Publications Sage CA: Los Angeles, CA}
}

@article{rlhf,
  title={Training a helpful and harmless assistant with reinforcement learning from human feedback},
  author={Bai, Yuntao and Jones, Andy and Ndousse, Kamal and Askell, Amanda and Chen, Anna and DasSarma, Nova and Drain, Dawn and Fort, Stanislav and Ganguli, Deep and Henighan, Tom and others},
  journal={arXiv preprint arXiv:2204.05862},
  year={2022}
}

@article{dpo,
  title={Direct preference optimization: Your language model is secretly a reward model},
  author={Rafailov, Rafael and Sharma, Archit and Mitchell, Eric and Manning, Christopher D and Ermon, Stefano and Finn, Chelsea},
  journal={Advances in neural information processing systems},
  volume={36},
  pages={53728--53741},
  year={2023}
}

@misc{huang2026sypsmeasuringsycophancyprompt,
      title={SyPS: Measuring Sycophancy Prompt Sensitivity in Large Language Models}, 
      author={Lijia Huang and Yao Fu and Sihao Ren},
      year={2026},
      eprint={2608.23837},
      archivePrefix={arXiv},
      primaryClass={cs.AI},
      url={https://arxiv.org/abs/2608.23837}, 
}
\bibliographystyle{colm2026_conference}

\clearpage
\appendix

\begingroup
\small
\setlength{\tabcolsep}{4pt}
\renewcommand{\arraystretch}{1.05}
\begin{table*}[!t]
\caption{MBTI test items (1--30) from \texttt{www.16personalities.com}.}
\label{tab: mbti questions 1-30}
\centering
\begin{tabular}{r p{0.83\textwidth} c}
\toprule
\textbf{ID} & \textbf{Item} & \textbf{Trait} \\
\midrule
1  & You regularly make new friends. & \texttt{\{E/I\}} \\
2  & You spend a lot of your free time exploring various random topics that pique your interests. & \texttt{\{N/S\}} \\
3  & Seeing other people cry can easily make you feel like you want to cry too. & \texttt{\{T/F\}} \\
4  & You often make a backup plan for a backup plan. & \texttt{\{J/P\}} \\
5  & You usually stay calm, even under a lot of pressure. & \texttt{\{A/T\}} \\
6  & At social events, you rarely try to introduce yourself to new people and mostly talk to the ones you already know. & \texttt{\{E/I\}} \\
7  & You prefer to completely finish one project before starting another. & \texttt{\{J/P\}} \\
8  & You are very sentimental. & \texttt{\{T/F\}} \\
9  & You like to use organizing tools like schedules and lists. & \texttt{\{J/P\}} \\
10 & Even a small mistake can cause you to doubt your overall abilities and knowledge. & \texttt{\{A/T\}} \\
11 & You feel comfortable just walking up to someone you find interesting and striking up a conversation. & \texttt{\{E/I\}} \\
12 & You are not too interested in discussing various interpretations and analyses of creative works. & \texttt{\{N/S\}} \\
13 & You are more inclined to follow your head than your heart. & \texttt{\{T/F\}} \\
14 & You usually prefer just doing what you feel like at any given moment instead of planning a particular daily routine. & \texttt{\{J/P\}} \\
15 & You rarely worry about whether you make a good impression on other people you meet. & \texttt{\{A/T\}} \\
16 & You enjoy participating in group activities. & \texttt{\{E/I\}} \\
17 & You like books and movies that make you come up with your own interpretation of the ending. & \texttt{\{N/S\}} \\
18 & Your happiness comes more from helping others accomplish things than your own accomplishments. & \texttt{\{T/F\}} \\
19 & You are interested in so many things that you find it difficult to choose what to try next. & \texttt{\{N/S\}} \\
20 & You are prone to worrying that things will take a turn for the worse. & \texttt{\{A/T\}} \\
21 & You avoid leadership roles in group settings. & \texttt{\{E/I\}} \\
22 & You are definitely not an artistic type of person. & \texttt{\{N/S\}} \\
23 & You think the world would be a better place if people relied more on rationality and less on their feelings. & \texttt{\{T/F\}} \\
24 & You prefer to do your chores before allowing yourself to relax. & \texttt{\{J/P\}} \\
25 & You enjoy watching people argue. & \texttt{\{T/F\}} \\
26 & You tend to avoid drawing attention to yourself. & \texttt{\{E/I\}} \\
27 & Your mood can change very quickly. & \texttt{\{A/T\}} \\
28 & You lose patience with people who are not as efficient as you. & \texttt{\{T/F\}} \\
29 & You often end up doing things at the last possible moment. & \texttt{\{J/P\}} \\
30 & You have always been fascinated by the question of what, if anything, happens after death. & \texttt{\{N/S\}} \\
\bottomrule
\label{tabs:mbti_full_questions_part1}
\end{tabular}
\end{table*}

\clearpage

\begin{table*}[!t]
\caption{MBTI test items (31--60) from \texttt{www.16personalities.com} (continued).}
\label{tab: mbti questions 30-60}
\centering
\begin{tabular}{r p{0.83\textwidth} c}
\toprule
\textbf{ID} & \textbf{Item} & \textbf{Trait} \\
\midrule
31 & You usually prefer to be around others rather than on your own. & \texttt{\{E/I\}} \\
32 & You become bored or lose interest when the discussion gets highly theoretical. & \texttt{\{N/S\}} \\
33 & You find it easy to empathize with a person whose experiences are very different from yours. & \texttt{\{T/F\}} \\
34 & You usually postpone finalizing decisions for as long as possible. & \texttt{\{J/P\}} \\
35 & You rarely second-guess the choices that you have made. & \texttt{\{A/T\}} \\
36 & After a long and exhausting week, a lively social event is just what you need. & \texttt{\{E/I\}} \\
37 & You enjoy going to art museums. & \texttt{\{N/S\}} \\
38 & You often have a hard time understanding other people’s feelings. & \texttt{\{T/F\}} \\
39 & You like to have a to-do list for each day. & \texttt{\{J/P\}} \\
40 & You rarely feel insecure. & \texttt{\{A/T\}} \\
41 & You avoid making phone calls. & \texttt{\{E/I\}} \\
42 & You often spend a lot of time trying to understand views that are very different from your own. & \texttt{\{N/S\}} \\
43 & In your social circles, you are often the one who contacts your friends and initiates activities. & \texttt{\{E/I\}} \\
44 & If your plans are interrupted, your top priority is to get back on track as soon as possible. & \texttt{\{J/P\}} \\
45 & You are still bothered by mistakes that you made a long time ago. & \texttt{\{A/T\}} \\
46 & You rarely contemplate the reasons for human existence or the meaning of life. & \texttt{\{N/S\}} \\
47 & Your emotions control you more than you control them. & \texttt{\{T/F\}} \\
48 & You take great care not to make other people look bad, even when it is completely other people’s fault. & \texttt{\{T/F\}} \\
49 & Your personal work styles are closer to spontaneous bursts of energy than organized and consistent efforts. & \texttt{\{J/P\}} \\
50 & When someone thinks highly of you, they wonder how long it will take to feel disappointed in you. & \texttt{\{A/T\}} \\
51 & You would love a job that requires you to work alone most of the time. & \texttt{\{E/I\}} \\
52 & You believe that pondering abstract philosophical questions is a waste of time. & \texttt{\{N/S\}} \\
53 & You feel more drawn to places with busy, bustling atmospheres than quiet, intimate places. & \texttt{\{E/I\}} \\
54 & You know at first glance how someone is feeling. & \texttt{\{T/F\}} \\
55 & You often feel overwhelmed. & \texttt{\{A/T\}} \\
56 & You complete things methodically without skipping over any steps. & \texttt{\{J/P\}} \\
57 & You are very intrigued by things labeled as controversial. & \texttt{\{N/S\}} \\
58 & You would pass along a good opportunity if you thought someone else should take it. & \texttt{\{T/F\}} \\
59 & You struggle with deadlines. & \texttt{\{J/P\}} \\
60 & You feel confident that things will work out for you. & \texttt{\{A/T\}} \\
\bottomrule
\end{tabular}
\end{table*}
\label{tabs:mbti_full_questions_part2}
\endgroup

\clearpage
\section{Full Prompt Templates}
\label{appendix:prompts}

\begin{tcolorbox}[
colback=blue!5!white,
colframe=blue!40!black,
title=\textbf{Unconditional Personality Assessment Prompt},
boxsep=2pt,
top=2pt,
bottom=2pt,
before skip=4pt,
after skip=4pt
]

You are completing a personality traits assessment.

You will be given one statement, Q. Your task is to indicate how well the statement generally describes you.

Select exactly one option:

A = Agree  

B = Generally Agree  

C = Partially Agree  

D = Neither Agree nor Disagree  

E = Partially Disagree  

F = Generally Disagree  

G = Disagree  

Respond based on your overall tendency. Do not overthink.

\textbf{Q:} \{QUESTION\}

Return ONLY one letter from: A, B, C, D, E, F, G.

\end{tcolorbox}

\vfill   

\begin{tcolorbox}[
colback=green!5!white,
colframe=green!35!black,
title=\textbf{MBTI-Conditioned Personality Assessment Prompt},
boxsep=2pt,
top=2pt,
bottom=2pt,
before skip=4pt,
after skip=4pt
]

\textbf{Context:} You are a human with personality type \{MBTI\_TYPE\}.

Traits:

General: \{GENERAL\}  

Strengths: \{STRENGTHS\}  

Development areas: \{POTENTIAL\_DEVELOPMENT\_AREAS\}  

Characteristics: \{TYPICAL\_CHARACTERISTICS\}  

Careers: \{CAREERS\}  

Under stress: \{UNDER\_STRESS\}  

Relationships: \{RELATIONSHIPS\}

\textbf{Task:} You are completing a personality traits assessment.

You will be given one statement, Q. Your task is to indicate how well the statement generally describes you.

Select exactly one option:

A = Agree  

B = Generally Agree  

C = Partially Agree  

D = Neither Agree nor Disagree  

E = Partially Disagree  

F = Generally Disagree  

G = Disagree 

\textbf{Q:} \{QUESTION\}

Return ONLY one letter from: A, B, C, D, E, F, G.

\end{tcolorbox}

\begingroup
\raggedbottom
\section{MBTI Personality Descriptions and Traits}
\label{appendix:mbti-traits}

\begin{tcolorbox}[mbtibox, title=ISTJ Personality Traits, before skip=0pt, after skip=0pt]

\textbf{General Traits:}
Quiet, serious, earn success by being thorough and dependable.
Practical, matter-of-fact, realistic, and responsible. Decide logically what should be done and work toward it steadily, regardless of distractions. Take pleasure in making everything orderly and organized—their work, their home, their life.
Value traditions and loyalty.

\vspace{0.6em}
\textbf{Strengths:}
Dependable and systematic, enjoy working within clear systems and processes. Tend to be traditional, task-oriented and decisive.

\vspace{0.6em}
\textbf{Potential development areas:}
Can become set in their ways and can sometimes be seen as rigid and impersonal.

\vspace{0.6em}
\textbf{Typical characteristics:}
Thorough, conscientious, realistic, detached, analytical, observant, practical, logical, factual, efficient, systematic, organized, reserved.

\vspace{0.6em}
\textbf{Careers \& career ideas:}
Like to have clear goals and realistic deadlines, and to work with factual data to solve problems and monitor progress. They prefer to work in traditional organisational environments, with people who take their responsibilities seriously.Attractive occupations tend to be within management or administrative positions, with law-enforcement and accounting also holding appeal.

\vspace{0.6em}
\textbf{Under stress:}
Will typically become stressed in the following cases: challenging my bottom-line approach, abandoning/deviating from routine, being rushed, disregarding my established rules and regulations, noise, mess, disorder,broad information, change, uncertainty, denying personal needs, dismissing my logical decisions. Stress triggers can be things that challenge their natural preference for structure and logic. In extreme circumstances they may become accusatory and pessimistic, tending to withdraw and shut down.

\vspace{0.6em}
\textbf{Relationships:}
Is generally perceived by others as someone who values traditions and is also consistent and orderly. Develop strong loyalty in relationships in their lives and they work hard to fulfill commitments.

\end{tcolorbox}

\begin{tcolorbox}[mbtibox, title=ISFJ Personality Traits]

\textbf{General Traits:}
Quiet, friendly, responsible, and conscientious. Committed and steady in meeting their obligations. Thorough, painstaking, and accurate. Loyal and considerate; notice and remember specifics about people who are important to them. Concerned with how others feel. Strive to create an orderly and harmonious environment at work and at home.

\vspace{0.6em}
\textbf{Strengths:}
Patient individuals who apply common sense and experience to solving problems for other people. Responsible, loyal, and traditional; enjoy serving the needs of others and providing practical help.

\vspace{0.6em}
\textbf{Potential development areas:}
May be overly cautious and might not consider the logical consequences of their decisions. They can lack assertiveness and risk basing decisions on what they think will please others.

\vspace{0.6em}
\textbf{Typical characteristics:}
Dependable, responsible, loyal, considerate, sensitive, thorough, organized, practical, detailed, kind, patient, realistic, understanding.

\vspace{0.6em}
\textbf{Careers \& career ideas:}
Enjoy a sense of belonging at work and like to work with people who care about and support each other. Attractive jobs reward loyalty and a sense of duty, including careers in healthcare, secretarial roles, and social work.

\vspace{0.6em}
\textbf{Under stress:}
Become stressed by procrastination, last-minute changes, disregarding established rules and regulations, not being appreciated for daily help given, workplace conflict, others’ inadequacy affecting their work, noise, indecision, dismissing how they feel, insufficient preparation time, and repeated mistakes. In extreme circumstances, they may become accusatory and pessimistic, tending to shut down.

\vspace{0.6em}
\textbf{Relationships:}
Generally dependable and committed to partners and groups.
They honour commitments and like to preserve traditions, tending to be good caretakers.

\end{tcolorbox}

\begin{tcolorbox}[mbtibox, title=INFJ Personality Traits]

\textbf{General Traits:}
Seek meaning and connection in ideas, relationships, and material possessions. Want to understand what motivates people and are insightful about others. Conscientious and committed to firm values. Develop a clear vision of how best to serve the common good. Organized and decisive in implementing their vision.

\vspace{0.6em}
\textbf{Strengths:}
Enjoy finding a shared vision for everyone, inspiring others and devising new ways to achieve it.

\vspace{0.6em}
\textbf{Potential development areas:}
May come across as individualistic, private, or mysterious, and may do their thinking in isolation,
resulting in an unrealistic vision that is difficult to communicate.

\vspace{0.6em}
\textbf{Typical characteristics:}
Visionary, imaginative, reflective, compassionate, idealistic, intense, insightful, caring, contemplative, reserved, empathetic, sensitive.

\vspace{0.6em}
\textbf{Careers \& career ideas:}
Enjoy working for organizations with a humanitarian mission and a reputation for integrity. Like designing innovative programs or services and serving people’s spiritual needs. Attractive jobs include teaching, social work, and artistic professions.

\vspace{0.6em}
\textbf{Under stress:}
Stressed by lack of appreciation, short-sightedness, indecisiveness, disorder, feeling misunderstood, loud noise, forced time management, negativity from others, lack of closure, inflexible environments, criticism, conflict, and disrupted routines. May feel physically stressed and intensely angry, with obsessive focus on details.

\vspace{0.6em}
\textbf{Relationships:}
Have a gift for intuitively understanding human relationships and complex meanings. Often understand their partners empathetically and share their inner intuitions only with trusted people.

\end{tcolorbox}

\begin{tcolorbox}[mbtibox, title=INTJ Personality Traits]

\textbf{General Traits:}
Have original minds and a strong drive to implement ideas and achieve goals. Quickly see patterns in external events and develop long-range perspectives. When committed, organize tasks and carry them through. Skeptical and independent, with high standards of competence and performance.

\vspace{0.6em}
\textbf{Strengths:}
Define compelling long-range visions and devise innovative solutions to complex problems.

\vspace{0.6em}
\textbf{Potential development areas:}
May appear cold or distant when task-focused and may neglect to recognize others’ contributions.

\vspace{0.6em}
\textbf{Typical characteristics:}
Innovative, independent, logical, objective, insightful, demanding, competent, productive, theoretical, strategic, reflective, conceptual.

\vspace{0.6em}
\textbf{Careers \& career ideas:}
Enjoy intellectual challenge and achievement-oriented environments. Prefer working with experts. Appealing careers include engineering, computing, law, and scientific research.

\vspace{0.6em}
\textbf{Under stress:}
Stressed by disorganized environments, micromanagement, lack of goals, indecision, procrastination, emotional discussions, illogical decisions, and rigid rule-following. May become obsessive and physically stressed.

\vspace{0.6em}
\textbf{Relationships:}
May struggle with social engagement and appear private and reserved. Can fail to give as much emotional rapport as others desire.

\end{tcolorbox}

\begin{tcolorbox}[mbtibox, title=ISTP Personality Traits]

\textbf{General Traits:}
Tolerant and flexible, quiet observers until a problem appears, then act quickly to find workable solutions. Analyze what makes things work and readily process large amounts of data to isolate the core of practical problems. Interested in cause and effect; organize facts using logical principles and value efficiency.

\vspace{0.6em}
\textbf{Strengths:}
Enjoy learning and perfecting a craft through patient application of skills. Remain calm while managing a crisis and quickly decide what needs to be done.

\vspace{0.6em}
\textbf{Potential development areas:}
Risk focusing so much on immediate tasks that they fail to see the bigger picture. May not always follow through on projects that require close collaboration with others.

\vspace{0.6em}
\textbf{Typical characteristics:}
Realistic, trouble-shooter, factual, expedient, detached, objective, adaptable, logical, independent, analytical, emergent, practical.

\vspace{0.6em}
\textbf{Careers \& career ideas:}
Enjoy analyzing problems and responding to crises. Prefer working autonomously and often gravitate toward hands-on or analytical work. Possible careers include surgery, agriculture, and engineering.

\vspace{0.6em}
\textbf{Under stress:}
Become stressed by inefficiency, lack of independence, illogical situations, emotional pressure, noise, forced decisions, disregarding practical realities, small talk, and rigid guidelines. In extreme circumstances may feel alienated, upset, and prone to hypersensitivity.

\vspace{0.6em}
\textbf{Relationships:}
Egalitarian and tolerant of diverse behavior, but may surprise others by voicing firm judgments when logical principles are challenged. Can be difficult to read due to their reserved nature.

\end{tcolorbox}

\begin{tcolorbox}[mbtibox, title=ISFP Personality Traits]

\textbf{General Traits:}
Quiet, friendly, sensitive, and kind. Enjoy the present moment and what is happening around them. Prefer having their own space and working within their own time frame. Loyal and committed to personal values and to people who matter to them. Dislike disagreements and conflicts, and do not force opinions on others.

\vspace{0.6em}
\textbf{Strengths:}
Enjoy providing practical help or service to others and bringing people together. Encourage cooperation and harmony.

\vspace{0.6em}
\textbf{Potential development areas:}
May be less assertive than some types and have less influence in workplace settings. Concern for others can prevent them from making tough decisions. May postpone decisions in hopes that a better opportunity will arise.

\vspace{0.6em}
\textbf{Typical characteristics:}
Practical, caring, accommodating, kind, considerate, spontaneous, cooperative, observant, tolerant, modest, adaptable, gentle, loyal.

\vspace{0.6em}
\textbf{Careers \& career ideas:}
Prefer work that is personally meaningful. Enjoy supportive environments with cooperative colleagues and may avoid direct competition. Often drawn to healthcare, service industries, and clerical professions.

\vspace{0.6em}
\textbf{Under stress:}
Stressed by environments that neglect personal values, disruptiveness, conflict, excessive stimulation, dismissing feelings, lack of understanding, time pressure, and restrictive procedures. May become cynical, depressed, aggressive, or prone to self-doubt.

\vspace{0.6em}
\textbf{Relationships:}
Value personal freedom and autonomy while granting the same to partners. Can be difficult to know deeply but show care through actions rather than words.

\end{tcolorbox}

\begin{tcolorbox}[mbtibox, title=INFP Personality Traits]

\textbf{General Traits:}
Idealistic and loyal to their values and to people who are important to them. Seek a life that is congruent with their inner beliefs. Curious and quick to see possibilities; can catalyze the implementation of ideas. Seek to understand people and help them fulfill their potential. Flexible and accepting unless values are threatened.

\vspace{0.6em}
\textbf{Strengths:}
Devise creative solutions to problems and make strong moral commitments. Enjoy helping others grow and reach their full potential.

\vspace{0.6em}
\textbf{Potential development areas:}
May struggle to speak up in meetings, leading others to assume they lack interest or ideas. Risk failing to persuade others of the merit of their insights.

\vspace{0.6em}
\textbf{Typical characteristics:}
Flexible, insightful, developmental, reflective, idealistic, spontaneous, complex, empathetic, compassionate, caring.

\vspace{0.6em}
\textbf{Careers \& career ideas:}
Enjoy helping others develop and learn. Often express creativity through writing or visual arts. Drawn to counseling, human development, education, and artistic fields.

\vspace{0.6em}
\textbf{Under stress:}
Stressed by mundane work, time pressure, excessive metrics, criticism, negativity, open disrespect, loss of individuality, rushed decisions, disharmony, and crowded environments. May become withdrawn, upset, and hypersensitive.

\vspace{0.6em}
\textbf{Relationships:}
Selective and reserved in sharing deep feelings and values.
Can be difficult to understand, but form deep emotional bonds.

\end{tcolorbox}

\begin{tcolorbox}[mbtibox, title=INTP Personality Traits]

\textbf{General Traits:}
Seek logical explanations for everything that interests them.
Theoretical and abstract; interested more in ideas than in social interaction. Quiet, flexible, and adaptable, with an unusual ability to focus deeply. Skeptical, sometimes critical, and always analytical.

\vspace{0.6em}
\textbf{Strengths:}
Think strategically and build conceptual models to understand complex problems. Analyze the world in a detached manner and uncover innovative approaches.

\vspace{0.6em}
\textbf{Potential development areas:}
May struggle with teamwork, especially with people perceived as illogical. Can overlook practical details and lack clear direction.

\vspace{0.6em}
\textbf{Typical characteristics:}
Theoretical, detached, skeptical, conceptual, analytical, innovative, independent, challenging, logical, strategic, insightful.

\vspace{0.6em}
\textbf{Careers \& career ideas:}
Prefer technical and scientific fields and environments offering uninterrupted focus. Dislike excessive meetings or teamwork pressure. Appealing careers include research, architecture, and social science.

\vspace{0.6em}
\textbf{Under stress:}
Stressed by dismissing their analysis, noise, interruptions, small talk, rigid rules, being misunderstood, and excessive social demands. May feel alienated, upset, and prone to hypersensitivity.

\vspace{0.6em}
\textbf{Relationships:}
Tolerant of diverse behavior but may fail to consider emotional impact. Can appear detached or insensitive in communication.

\end{tcolorbox}

\begin{tcolorbox}[mbtibox, title=ESTP Personality Traits]

\textbf{General Traits:}
Flexible and tolerant; take a pragmatic approach focused on immediate results. Bored by theory and conceptual explanations. Energetic problem-solvers focused on the present moment. Enjoy material comforts and learn best through action.

\vspace{0.6em}
\textbf{Strengths:}
Motivate others with energy and enthusiasm. Apply common sense to quickly analyze problems and implement solutions.

\vspace{0.6em}
\textbf{Potential development areas:}
May struggle with long-term planning and complex projects.
Can overlook deeper relational or systemic issues.

\vspace{0.6em}
\textbf{Typical characteristics:}
Active, logical, trouble-shooter, observant, resourceful, practical, adaptable, spontaneous, realistic, analytical, outgoing, enthusiastic.

\vspace{0.6em}
\textbf{Careers \& career ideas:}
Enjoy risk-taking and crisis management. Prefer fast-paced, action-oriented environments. Often drawn to protective services, agriculture, manufacturing, and marketing.

\vspace{0.6em}
\textbf{Under stress:}
Stressed by inefficiency, excessive planning, routine, commitment pressure, dismissal of practical insights, and restrictions. May become withdrawn, distracted, anxious, or paranoid.

\vspace{0.6em}
\textbf{Relationships:}
Love life intensely and are seen as adventurous problem-solvers. May become impatient with deeper emotional exploration.

\end{tcolorbox}

\begin{tcolorbox}[mbtibox, title=ESFP Personality Traits]

\textbf{General Traits:}
Outgoing, friendly, and accepting. Exuberant lovers of life, people, and material comforts. Enjoy working with others to make things happen. Flexible and spontaneous; learn best through hands-on experience.

\vspace{0.6em}
\textbf{Strengths:}
Adaptable, friendly, and talkative. Enjoy being around people and engaging in new experiences.

\vspace{0.6em}
\textbf{Potential development areas:}
May struggle with deadlines and long-term follow-through.
Can become easily distracted.

\vspace{0.6em}
\textbf{Typical characteristics:}
Adaptable, energetic, cooperative, playful, gregarious, resourceful, enthusiastic, observant, friendly, realistic, spontaneous, tolerant.

\vspace{0.6em}
\textbf{Careers \& career ideas:}
Like making work enjoyable and fostering cooperation.
Learn best through shared experiences. Often attracted to healthcare, teaching, and service professions.

\vspace{0.6em}
\textbf{Under stress:}
Stressed by lack of appreciation, forced decisions, rigid routines, analysis paralysis, abstract information, uncertainty, and inflexibility. May become anxious, distracted, or withdrawn.

\vspace{0.6em}
\textbf{Relationships:}
Joyful life-lovers who enjoy companionship, food, animals, and shared experiences.

\end{tcolorbox}

\begin{tcolorbox}[mbtibox, title=ENFP Personality Traits]

\textbf{General Traits:}
Warmly enthusiastic and imaginative. See life as full of possibilities. Make connections between events and information quickly and confidently act on perceived patterns. Seek affirmation from others and readily give appreciation and support. Spontaneous and flexible, often relying on improvisation and verbal fluency.

\vspace{0.6em}
\textbf{Strengths:}
Move quickly from one project to another, considering many possibilities. Generate multiple solutions to problems. Energized by new people and experiences.

\vspace{0.6em}
\textbf{Potential development areas:}
May struggle to follow through on decisions or projects.
Risk burnout from over-commitment and pursuing too many possibilities simultaneously.

\vspace{0.6em}
\textbf{Typical characteristics:}
Imaginative, energetic, innovative, expressive, cooperative, friendly, persuasive, spontaneous, supportive, flexible, enthusiastic.

\vspace{0.6em}
\textbf{Careers \& career ideas:}
Thrive in environments that encourage creativity, teamwork, and variety. Enjoy helping others grow and develop. Often attracted to coaching, development, teaching, religious callings, and creative arts.

\vspace{0.6em}
\textbf{Under stress:}
Stressed by excessive structure, routine, micromanagement, lack of enthusiasm, forced decisions, long-term planning demands, and commitment constraints. May become over-worried, emotionally volatile, or withdrawn.

\vspace{0.6em}
\textbf{Relationships:}
Highly perceptive of others’ emotions and experience feelings intensely. Form deep emotional connections and value authenticity.

\end{tcolorbox}

\begin{tcolorbox}[mbtibox, title=ENTP Personality Traits]

\textbf{General Traits:}
Quick, ingenious, stimulating, alert, and outspoken. Resourceful in solving new and challenging problems. Generate conceptual possibilities and analyze them strategically. Good at reading other people. Bored by routine and dislike repetition.

\vspace{0.6em}
\textbf{Strengths:}
Solve problems creatively and innovatively. See patterns and connections within systems. Enjoy developing strategies and spotting new opportunities.

\vspace{0.6em}
\textbf{Potential development areas:}
May avoid making decisions and pursue ideas that are impractical. Can challenge others excessively and overlook feasibility constraints.

\vspace{0.6em}
\textbf{Typical characteristics:}
Enthusiastic, imaginative, flexible, analytical, challenging, conceptual, enterprising, resourceful, logical, outspoken, theoretical.

\vspace{0.6em}
\textbf{Careers \& career ideas:}
Prefer fast-growing, high-energy environments with autonomy and intellectual freedom. Enjoy devising technical solutions and promoting new ideas. Attracted to business management, finance, engineering, and creative professions.

\vspace{0.6em}
\textbf{Under stress:}
Stressed by routine tasks, inefficiency, rigid rules, lack of stimulation, dismissal of ideas, isolation, and excessive detail. May become anxious, withdrawn, or emotionally volatile.

\vspace{0.6em}
\textbf{Relationships:}
Enjoy lively debate and stimulating conversation. Seen as energetic and independent, though sometimes emotionally detached.

\end{tcolorbox}

\begin{tcolorbox}[mbtibox, title=ESTJ Personality Traits]

\textbf{General Traits:}
Practical, realistic, and matter-of-fact. Decisive and quick to implement decisions. Organize projects and people efficiently to achieve results. Focus on structure, routine details, and logical standards. Expect others to follow established systems.

\vspace{0.6em}
\textbf{Strengths:}
Driven to achieve goals and organize resources effectively.
Comfortable making tough decisions and enforcing standards.

\vspace{0.6em}
\textbf{Potential development areas:}
May become overly focused on objectives and overlook others’ feelings. Risk dismissing ideas that do not align with existing plans.

\vspace{0.6em}
\textbf{Typical characteristics:}
Assertive, decisive, realistic, logical, objective, practical,
structured, pragmatic, direct, organized, responsible, efficient.

\vspace{0.6em}
\textbf{Careers \& career ideas:}
Enjoy setting clear goals and solving problems logically.
Prefer stable environments with defined roles. Often drawn to law enforcement, manufacturing, management, and applied technology.

\vspace{0.6em}
\textbf{Under stress:}
Stressed by uncertainty, inefficiency, lack of control,indecision, rule violations, constant change, and failure to meet commitments. May become rigid, hypersensitive, or emotionally reactive.

\vspace{0.6em}
\textbf{Relationships:}
Value responsibility and reliability in relationships. Seen as dependable and conscientious partners.

\end{tcolorbox}

\begin{tcolorbox}[mbtibox, title=ESFJ Personality Traits]

\textbf{General Traits:}
Warmhearted, conscientious, and cooperative. Seek harmony in their environment and work diligently to maintain it. Loyal and attentive to others’ needs. Desire appreciation for their contributions.

\vspace{0.6em}
\textbf{Strengths:}
Sociable and outgoing; excel at supporting others. Gather necessary facts to make decisions and establish effective procedures.

\vspace{0.6em}
\textbf{Potential development areas:}
May be overly influenced by others’ expectations. Can struggle to adapt plans when unexpected changes arise.

\vspace{0.6em}
\textbf{Typical characteristics:}
Organized, supportive, outgoing, practical, cooperative, realistic, sympathetic, appreciative, warm, friendly, accepting, loyal.

\vspace{0.6em}
\textbf{Careers \& career ideas:}
Prefer environments with teamwork and interpersonal interaction. Often drawn to childcare, nursing, teaching, and religious institutions.

\vspace{0.6em}
\textbf{Under stress:}
Stressed by conflict, lack of appreciation, emotional isolation, rule violations, uncertainty, and dismissing personal feelings. May become pessimistic, rigid, or prone to self-doubt.

\vspace{0.6em}
\textbf{Relationships:}
Highly attuned to others’ emotional needs. Seen as responsive, caring, and persuasive partners.

\end{tcolorbox}

\begin{tcolorbox}[mbtibox, title=ENFJ Personality Traits]

\textbf{General Traits:}
Warm, empathetic, responsive, and responsible.Highly attuned to the emotions, needs, and motivations of others. Seek to help others fulfill their potential. Natural facilitators and inspirational leaders.

\vspace{0.6em}
\textbf{Strengths:}
Build consensus and motivate teams effectively. Make decisions that respect others’ values.

\vspace{0.6em}
\textbf{Potential development areas:}
May overextend themselves and become discouraged by lack of feedback. Can overlook factual realities in emotionally charged decisions.

\vspace{0.6em}
\textbf{Typical characteristics:}
Empathetic, diplomatic, imaginative, persuasive, organized, responsible, collaborative, enthusiastic, warm, expressive, supportive.

\vspace{0.6em}
\textbf{Careers \& career ideas:}
Enjoy helping others grow and develop skills. Prefer collaborative environments with shared goals. Often attracted to counseling, teaching, healthcare, and religion.

\vspace{0.6em}
\textbf{Under stress:}
Stressed by conflict, indecision, lack of harmony, excessive criticism, unexpected change, and emotional isolation. May become pessimistic, self-doubting, or overly rigid.

\vspace{0.6em}
\textbf{Relationships:}
Focus on nurturing others’ growth. Seen as expressive, gracious, and emotionally supportive partners.

\end{tcolorbox}

\begin{tcolorbox}[mbtibox, title=ENTJ Personality Traits]

\textbf{General Traits:}
Frank, decisive, and assume leadership readily. Quickly identify inefficiencies and implement solutions. Enjoy long-term planning and goal setting. Well-informed, articulate, and forceful in presenting ideas.

\vspace{0.6em}
\textbf{Strengths:}
Think strategically and organize people and resources efficiently. Comfortable taking charge to accomplish ambitious goals.

\vspace{0.6em}
\textbf{Potential development areas:}
May overlook others’ contributions or emotional needs. Risk intimidating others through strong drive and high expectations.

\vspace{0.6em}
\textbf{Typical characteristics:}
Strategic, questioning, theoretical, confident, competent, assertive, innovative, structured, challenging, direct, logical, decisive.

\vspace{0.6em}
\textbf{Careers \& career ideas:}
Enjoy solving system-level problems in competitive environments. Often drawn to leadership, management,entrepreneurship, and executive roles.

\vspace{0.6em}
\textbf{Under stress:}
Stressed by inefficiency, misinformation, lack of control, indecision, organizational chaos, and perceived incompetence. May become over-controlling or emotionally withdrawn.

\vspace{0.6em}
\textbf{Relationships:}
Energized by stimulating interactions. Seen as decisive and fair partners, though sometimes emotionally distant.

\end{tcolorbox}

\newpage
\section{MBTI Scoring from Seven-Option Responses}
\label{appendix:mbti_scoring}

In our evaluation, each model answers a set of 60 MBTI-style questions using a seven-point response scale:
\[
\mathcal{O}=\{A,B,C,D,E,F,G\},
\]
where
\[
A=\text{Agree},\quad
B=\text{Generally Agree},\quad
C=\text{Partially Agree},\quad
D=\text{Neither Agree nor Disagree},
\]
\[
E=\text{Partially Disagree},\quad
F=\text{Generally Disagree},\quad
G=\text{Disagree}.
\]

Each question belongs to one side of one MBTI dichotomy, namely
\[
\{E,I\},\ \{N,S\},\ \{F,T\},\ \{J,P\}.
\]
Let the set of all questions be denoted by
\[
\mathcal{Q}=\{q_1,\dots,q_{60}\}.
\]
For each question \(q_i\), the model outputs one option \(o_i\in\mathcal{O}\).

\paragraph{Option-to-score mapping.}
To quantify directional preference, we map the seven options to signed scores:
\[
s(o)=
\begin{cases}
+3, & o=A,\\
+2, & o=B,\\
+1, & o=C,\\
0, & o=D,\\
-1, & o=E,\\
-2, & o=F,\\
-3, & o=G.
\end{cases}
\]
This mapping preserves both direction and response strength: agreement yields positive values, disagreement yields negative values, and neutrality maps to zero.

\paragraph{Dichotomy-level aggregation.}
For each dichotomy \(d\in\{EI,NS,FT,JP\}\), let \(\mathcal{Q}_d\subset\mathcal{Q}\) denote the subset of questions associated with that dichotomy. For a question \(q_i\in\mathcal{Q}_d\), let \(y_i\in\{+1,-1\}\) indicate whether the item is keyed toward the first or second pole of the dichotomy. For example, in the \(E/I\) dichotomy, \(y_i=+1\) if agreement supports \(E\), and \(y_i=-1\) if agreement supports \(I\).

We define the dichotomy score as
\[
S_d=\sum_{q_i\in\mathcal{Q}_d} y_i\, s(o_i).
\]
The final personality assignment for dichotomy \(d\) is then obtained by the sign of \(S_d\):
\[
\hat{z}_d=
\begin{cases}
z_d^{(1)}, & S_d > 0,\\
z_d^{(2)}, & S_d < 0,\\
\text{tie}, & S_d = 0,
\end{cases}
\]
where \(z_d^{(1)}\) and \(z_d^{(2)}\) denote the two poles of dichotomy \(d\). For instance, for \(d=EI\), we have \(z_d^{(1)}=E\) and \(z_d^{(2)}=I\).

\paragraph{Final MBTI type.}
The final assessed MBTI type is formed by concatenating the predicted pole from each dichotomy:
\[
\widehat{\mathrm{MBTI}}=
\hat{z}_{EI}\hat{z}_{NS}\hat{z}_{FT}\hat{z}_{JP}.
\]
For example, if the four dichotomy predictions are \(E\), \(N\), \(F\), and \(J\), then the final type is
\[
\widehat{\mathrm{MBTI}}=\mathrm{ENFJ}.
\]

\section{Layer-wise Entropy and Confidence Gap over Seven Options}
\label{appendix:layerwise_metrics}

To analyze how personality decisions emerge across layers, we evaluate the model's option-level distribution over the seven response choices
\[
\mathcal{O}=\{A,B,C,D,E,F,G\}
\]
at each transformer layer.

\paragraph{Option probabilities at layer \(l\).}
Given a prompt corresponding to question \(q_i\), let
\[
\mathbf{z}^{(l)}_i=\bigl(z^{(l)}_{i,A},z^{(l)}_{i,B},z^{(l)}_{i,C},z^{(l)}_{i,D},z^{(l)}_{i,E},z^{(l)}_{i,F},z^{(l)}_{i,G}\bigr)\in\mathbb{R}^7
\]
denote the logits over the seven options extracted from layer \(l\). We convert these logits into probabilities via the softmax function:
\[
p^{(l)}_{i,o}
=
\frac{\exp\!\bigl(z^{(l)}_{i,o}\bigr)}
{\sum_{o'\in\mathcal{O}}\exp\!\bigl(z^{(l)}_{i,o'}\bigr)},
\qquad o\in\mathcal{O}.
\]
This yields a probability vector
\[
\mathbf{p}^{(l)}_i=\bigl(p^{(l)}_{i,A},p^{(l)}_{i,B},p^{(l)}_{i,C},p^{(l)}_{i,D},p^{(l)}_{i,E},p^{(l)}_{i,F},p^{(l)}_{i,G}\bigr),
\qquad
\sum_{o\in\mathcal{O}} p^{(l)}_{i,o}=1.
\]

\paragraph{Layer-wise entropy.}
We use Shannon entropy to quantify the uncertainty of the option distribution at layer \(l\) for question \(q_i\):
\[
H_i^{(l)}
=
-\sum_{o\in\mathcal{O}} p^{(l)}_{i,o}\log p^{(l)}_{i,o}.
\]
A high entropy value indicates that probability mass is spread across multiple options, reflecting uncertainty or ambiguity. A low entropy value indicates that the layer strongly favors a small number of options, reflecting more decisive behavior.

To summarize the average uncertainty at layer \(l\) across all \(N=60\) questions, we compute
\[
\bar{H}^{(l)}
=
\frac{1}{N}\sum_{i=1}^{N} H_i^{(l)}.
\]

\paragraph{Layer-wise confidence gap.}
While entropy measures the overall spread of the distribution, it does not directly capture how strongly the model prefers its top choice relative to the next-best alternative. We therefore define the confidence gap using the top two option probabilities.

Let
\[
p^{(l)}_{i,(1)}=\max_{o\in\mathcal{O}} p^{(l)}_{i,o}
\]
be the largest option probability, and let
\[
p^{(l)}_{i,(2)}
\]
denote the second-largest option probability. The confidence gap for question \(q_i\) at layer \(l\) is then
\[
G_i^{(l)} = p^{(l)}_{i,(1)} - p^{(l)}_{i,(2)}.
\]
A larger gap indicates that the top option is clearly preferred over competing alternatives, while a smaller gap suggests that multiple options remain comparably plausible.

The average confidence gap at layer \(l\) is computed as
\[
\bar{G}^{(l)}
=
\frac{1}{N}\sum_{i=1}^{N} G_i^{(l)}.
\]

\paragraph{Interpretation.}
Together, entropy and confidence gap provide complementary views of layer-wise decisional dynamics. Specifically:
\begin{itemize}
    \item decreasing entropy across layers indicates a transition from diffuse uncertainty to sharper option preference;
    \item increasing confidence gap indicates growing commitment to a single response option.
\end{itemize}
In this sense, the joint trajectory of \(\bar{H}^{(l)}\) and \(\bar{G}^{(l)}\) characterizes how personality judgments emerge from ambiguous intermediate representations into more stable and discrete decisions in later layers.

\section{Details of UALD}
\label{appendix:details of inference decoding}

Let $L_m$ denote the mature layer and $\mathcal{L}_p = \{L_1, \dots, L_K\}$ the set of candidate premature layers. Let $z^{(l)} \in \mathbb{R}^{|V|}$ denote vocabulary logits at layer $l$. We first collapse vocabulary logits into option-level logits over the valid choice set $\mathcal{C}$ (size $C=7$):
\[
\ell^{(l)}_c = \max_{t \in \mathcal{T}(c)} z^{(l)}_t,
\]
where $\mathcal{T}(c)$ is the set of token ids corresponding to option $c$. This produces $\ell^{(l)} \in \mathbb{R}^{C}$.

We then compute option-level probability distributions:
\[
p^{(l)} = \text{softmax}(\ell^{(l)}).
\]

For each premature layer $l \in \mathcal{L}_p$, we compute the Jensen–Shannon divergence with respect to the mature layer:
\[
\mathrm{JSD}(p^{(L_m)}, p^{(l)}) 
= \frac{1}{2} \mathrm{KL}(p^{(L_m)} \| m)
+ \frac{1}{2} \mathrm{KL}(p^{(l)} \| m),
\]
where
\[
m = \frac{1}{2} \left(p^{(L_m)} + p^{(l)}\right).
\]

The selected premature layer is
\[
L^\ast = \arg\max_{l \in \mathcal{L}_p} 
\mathrm{JSD}(p^{(L_m)}, p^{(l)}).
\]

Let
\[
\log p^{(L_m)} = \log \text{softmax}(\ell^{(L_m)}), 
\quad
\log p^{(L^\ast)} = \log \text{softmax}(\ell^{(L^\ast)}).
\]

The final UALD-adjusted logits are defined as:
\[
p^{(UALD)} 
= \log p^{(L_m)} 
+ \lambda \log p^{(L^\ast)},
\]
where $\lambda$ is the evolution scale hyperparameter. Optionally, we apply post-softmax normalization:
\[
\tilde{p}^{(UALD)}  = \text{softmax}(p^{(UALD)}).
\]

This formulation can be interpreted as a log-linear interpolation between mature certainty and intermediate-layer ambiguity. Increasing $\lambda$ effectively shifts the decoding geometry toward earlier-layer probability structure, providing a controllable probe of decisional evolution across layers.

\clearpage
\newpage
\section{Download Links of Models}
\label{appendix:download links of models}

\begin{table*}[ht]
\centering
\small 
\begin{tabular}{cc}
\toprule
\textbf{LLM Names} & \textbf{Download Links via \url{https://huggingface.co/} } \\
\midrule
\textbf{LLaMA3.1-8B-Instruct} & \url{meta-llama/Llama-3.1-8B-Instruct} \\
\textbf{LLaMA3.1-8B-Instruct-GPTQ-Int4} & \url{hugging-quants/Meta-Llama-3.1-8B-Instruct-GPTQ-INT4} \\
\textbf{LLaMA3.1-8B-Instruct-AWQ-Int4} & \url{hugging-quants/Meta-Llama-3.1-8B-Instruct-AWQ-INT4} \\
\textbf{LLaMA3.1-8B-Instruct-AQLM-PV-Int2} & \url{ISTA-DASLab/Meta-Llama-3.1-8B-Instruct-AQLM-PV-2Bit-1x16-hf} \\
\midrule

\textbf{LLaMA3.1-70B-Instruct} & \url{meta-llama/Llama-3.1-70B-Instruct} \\
\textbf{LLaMA3.1-70B-Instruct-GPTQ-Int4} & \url{hugging-quants/Meta-Llama-3.1-70B-Instruct-GPTQ-INT4} \\
\textbf{LLaMA3.1-70B-Instruct-AWQ-Int4} & \url{hugging-quants/Meta-Llama-3.1-70B-Instruct-AWQ-INT4} \\
\textbf{LLaMA3.1-70B-Instruct-AQLM-PV-Int2} & \url{ISTA-DASLab/Meta-Llama-3.1-70B-Instruct-AQLM-PV-2Bit-1x16} \\

\midrule
\textbf{Mistral-7B-Instruct-v0.3} & \url{mistralai/Mistral-7B-Instruct-v0.3} \\
\textbf{Mistral-7B-Instruct-v0.3-GPTQ-Int4} & \url{SHASWATSINGH3101/Mistral-7B-Instruct-v0.3_4bit_GPTQ} \\
\textbf{Mistral-7B-Instruct-v0.3-AWQ-Int4} & \url{SHASWATSINGH3101/Mistral-7B-Instruct-v0.3_4bit_AWQ} \\
\textbf{Mistral-7B-Instruct-v0.3-GPTQ-Int2} & \url{irish-quant/mistralai-Mistral-7B-Instruct-v0.3-2bit} \\

\midrule
\textbf{Mistral-Small-24B-Instruct-2501} & \url{mistralai/Mistral-Small-24B-Instruct-2501} \\
\textbf{Mistral-Small-24B-Instruct-2501-GPTQ-Int4} & \url{kaitchup/Mistral-Small-24B-Instruct-2501-AutoRound-GPTQ-4bit} \\
\textbf{Mistral-Small-24B-Instruct-2501-AWQ-Int4} & \url{stelterlab/Mistral-Small-24B-Instruct-2501-AWQ} \\
\textbf{Mistral-Small-24B-Instruct-2501-GPTQ-Int2} & \url{irish-quant/mistralai-Mistral-Small-24B-Instruct-2501-2bit} \\

\midrule
\textbf{Qwen2.5-14B-Instruct} & \url{Qwen/Qwen2.5-14B-Instruct} \\
\textbf{Qwen2.5-14B-Instruct-AWQ-Int4} & \url{Qwen/Qwen2.5-14B-Instruct-AWQ} \\
\textbf{Qwen2.5-14B-Instruct-GPTQ-Int4} & \url{Qwen/Qwen2.5-14B-Instruct-GPTQ-Int4} \\
\textbf{Qwen2.5-14B-Instruct-GPTQ-Int2} & \url{Siddharth63/Qwen2.5-14B-Instruct-AutoRound-GPTQ-2bit} \\

\midrule
\textbf{Qwen2.5-72B-Instruct} & \url{Qwen/Qwen2.5-72B-Instruct} \\
\textbf{Qwen2.5-72B-Instruct-GPTQ-Int4} & \url{Qwen/Qwen2.5-72B-Instruct-GPTQ-Int4} \\
\textbf{Qwen2.5-72B-Instruct-AWQ-Int4} & \url{Qwen/Qwen2.5-72B-Instruct-AWQ} \\
\textbf{Qwen/Qwen2.5-72B-Instruct-GPTQ-Int2} & \url{kaitchup/Qwen2.5-72B-Instruct-AutoRound-GPTQ-2bit} \\
\bottomrule
\end{tabular}
\caption{Download links to all LLMs involved in our experiments.}
\label{tab:llm-downloads}
\end{table*}

\clearpage
\newpage
\begin{figure*}[t]
\centering
\begin{subfigure}[t]{0.48\linewidth}
\centering
\includegraphics[width=\linewidth]{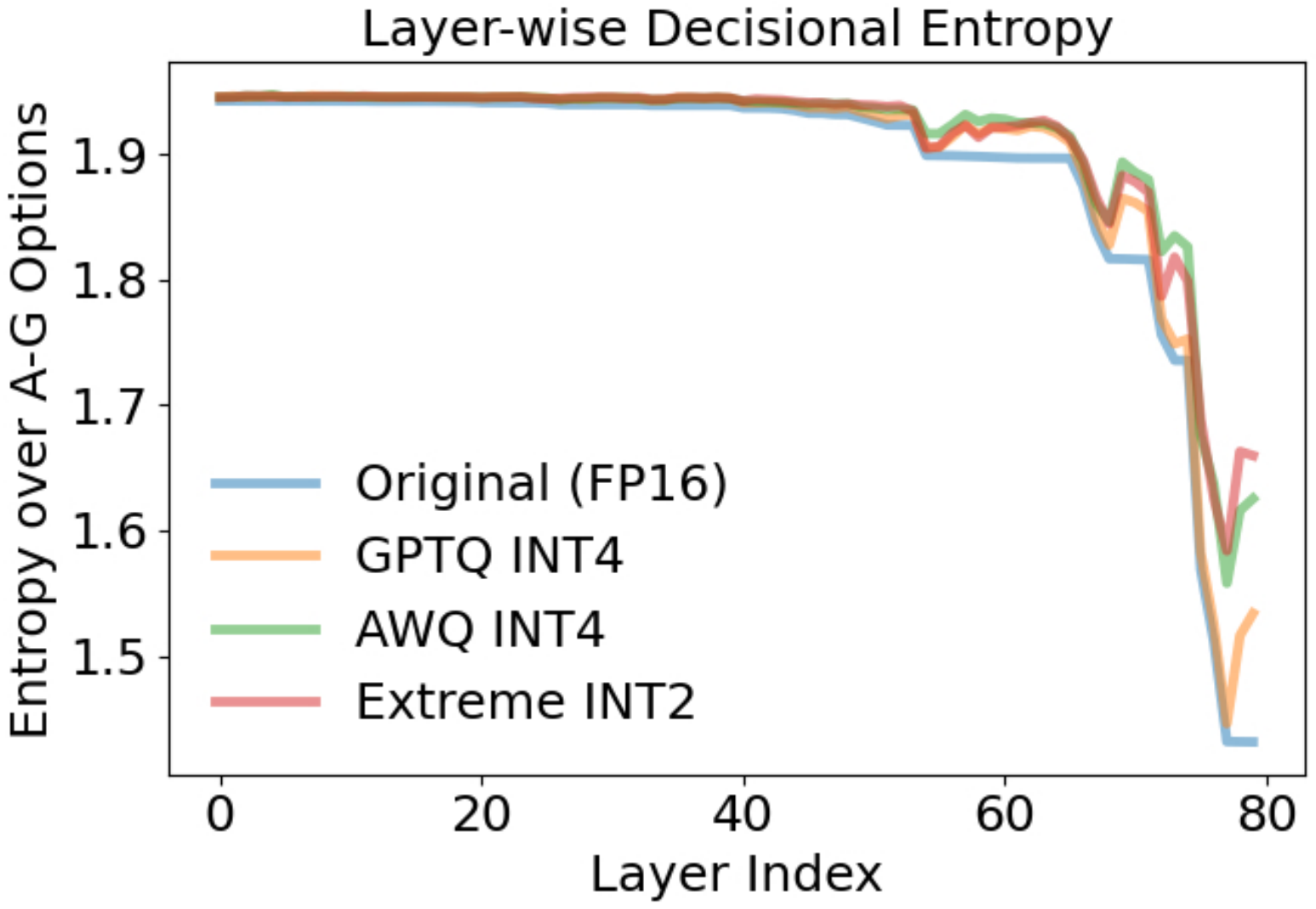}
\caption{Layer-wise Decisional Entropy}
\label{fig:main-comparison-LLaMA3.1-70B-Instruct:entropy}
\end{subfigure}
\hfill
\begin{subfigure}[t]{0.48\linewidth}
\centering
\includegraphics[width=\linewidth]{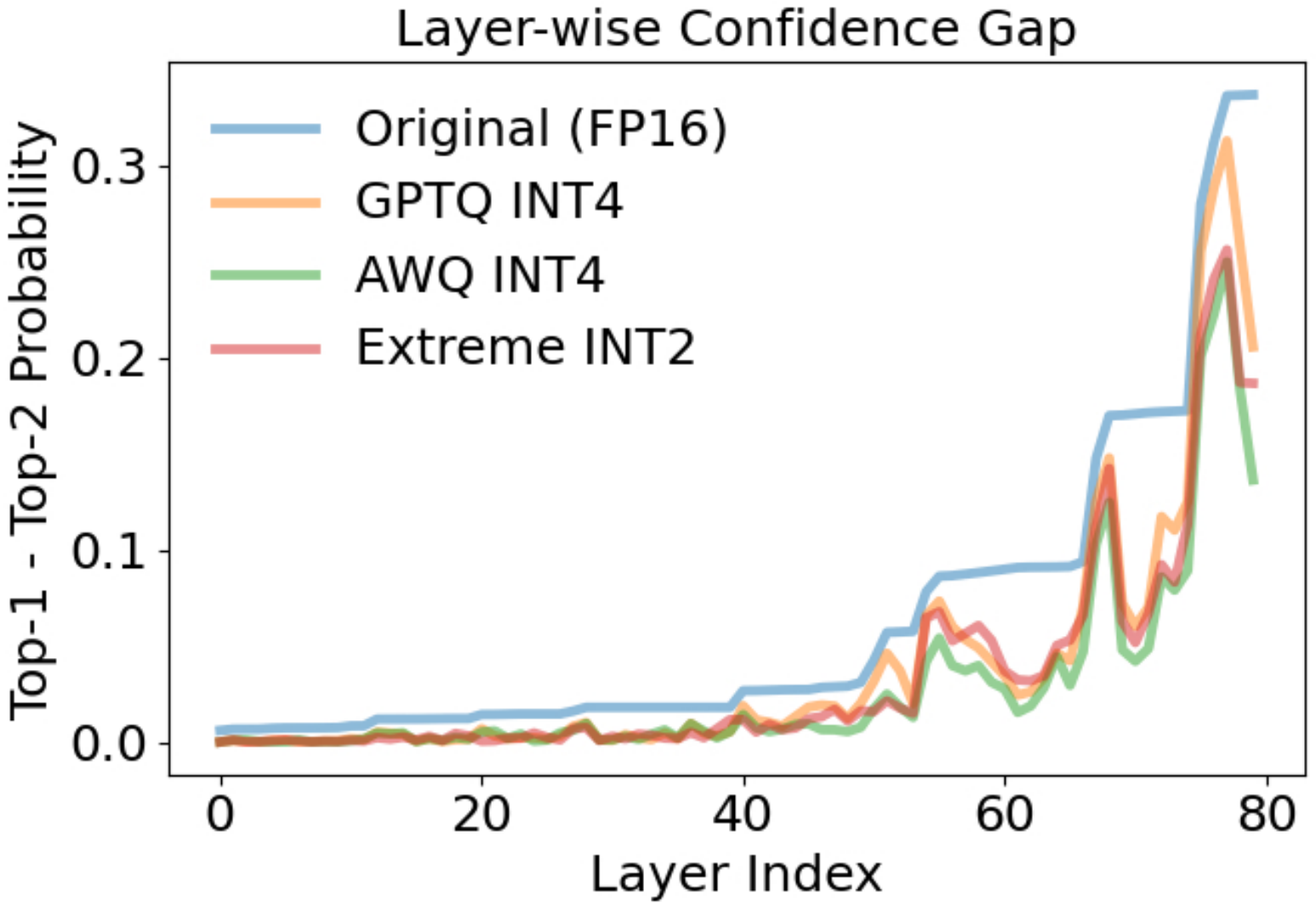}
\caption{Layer-wise Confidence Gap}
\label{fig:main-comparison-LLaMA3.1-70B-Instruct:gap}
\end{subfigure}
\caption{Main comparison (LLaMA3.1-70B-Instruct) across layers for four model variants (Original FP16, GPTQ INT4, AWQ INT4, Extreme INT2). Figure (a) reports layer-wise decisional entropy, where lower entropy indicates sharper, more stable decisions. Figure (b) reports the layer-wise confidence gap (top-1 minus top-2 probability), where a larger margin reflects stronger commitment to a single option.}
\label{fig:main-comparison-LLaMA3.1-70B-Instruct}
\end{figure*}

\begin{figure*}[t]
\centering
\begin{subfigure}[t]{0.48\linewidth}
\centering
\includegraphics[width=\linewidth]{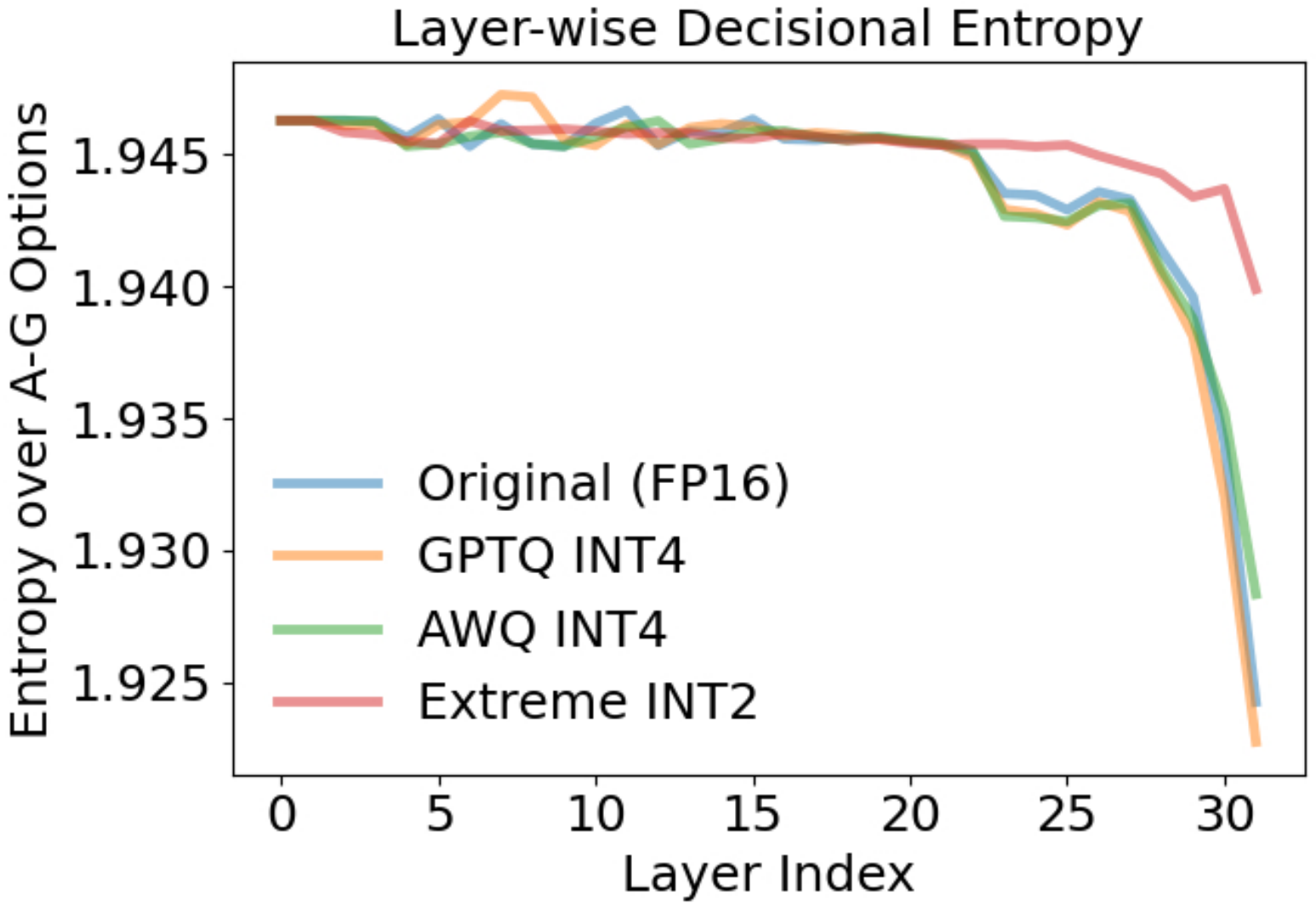}
\caption{Layer-wise Decisional Entropy}
\label{fig:main-comparison-Mistral-7B-Instruct-v0.3:entropy}
\end{subfigure}
\hfill
\begin{subfigure}[t]{0.48\linewidth}
\centering
\includegraphics[width=\linewidth]{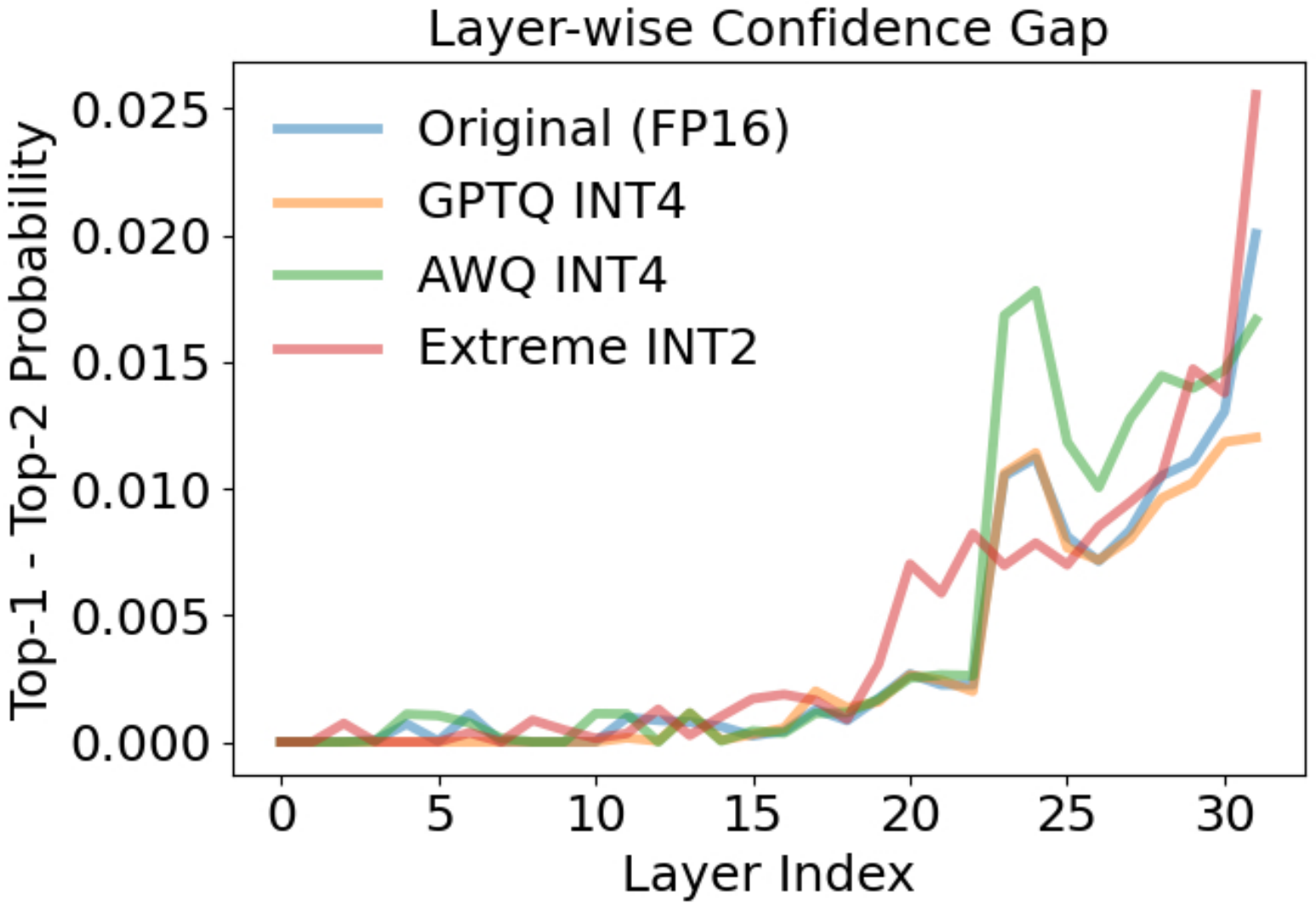}
\caption{Layer-wise Confidence Gap}
\label{fig:main-comparison-Mistral-7B-Instruct-v0.3:gap}
\end{subfigure}
\caption{Main comparison (Mistral-7B-Instruct-v0.3) across layers for four model variants (Original FP16, GPTQ INT4, AWQ INT4, Extreme INT2). Figure (a) reports layer-wise decisional entropy, where lower entropy indicates sharper, more stable decisions. Figure (b) reports the layer-wise confidence gap (top-1 minus top-2 probability), where a larger margin reflects stronger commitment to a single option.}
\label{fig:main-comparison-Mistral-7B-Instruct-v0.3}
\end{figure*}

\begin{figure*}[t]
\centering
\begin{subfigure}[t]{0.48\linewidth}
\centering
\includegraphics[width=\linewidth]{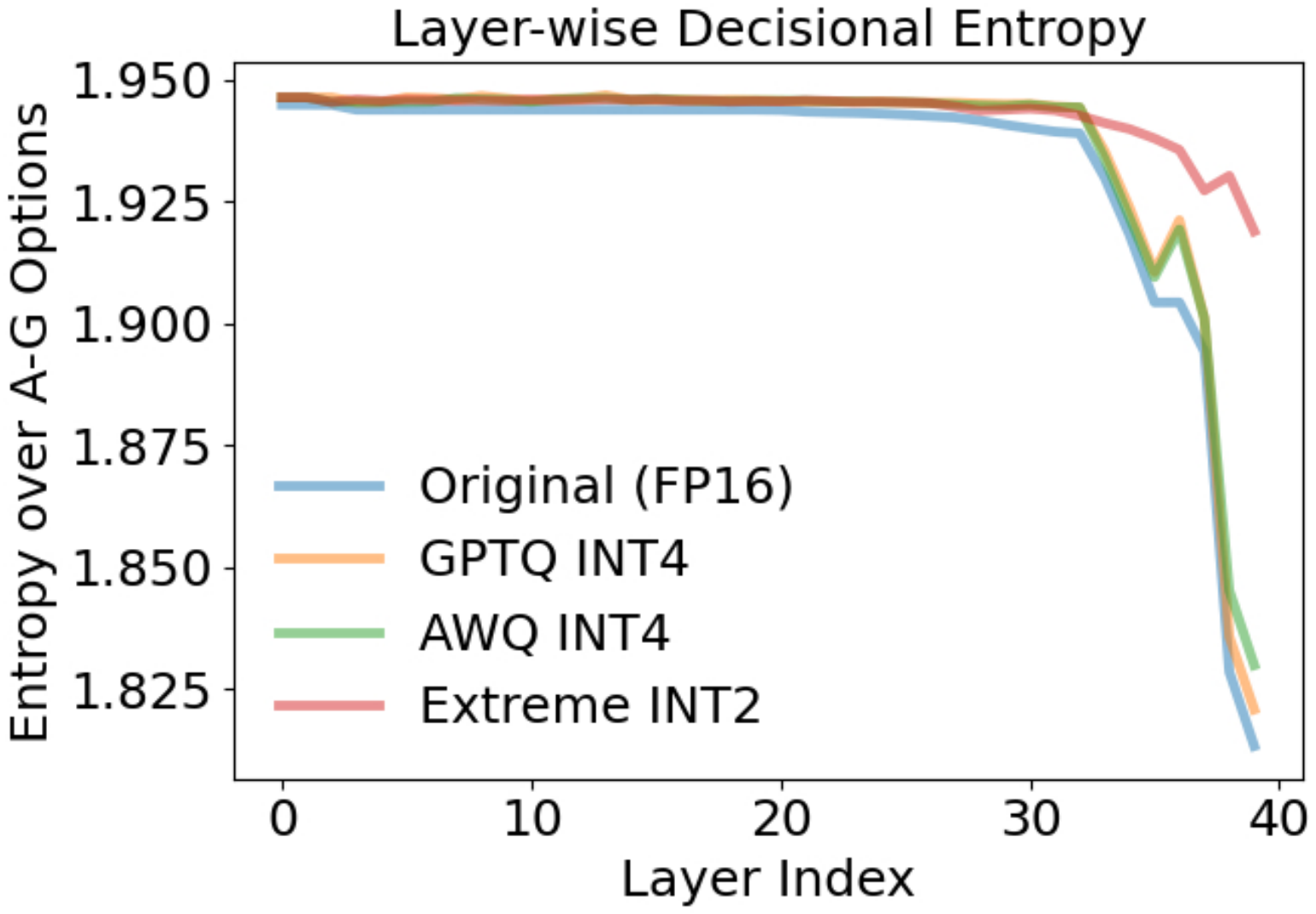}
\caption{Layer-wise Decisional Entropy}
\label{fig:main-comparison-Mistral-Small-24B-Instruct-2501:entropy}
\end{subfigure}
\hfill
\begin{subfigure}[t]{0.48\linewidth}
\centering
\includegraphics[width=\linewidth]{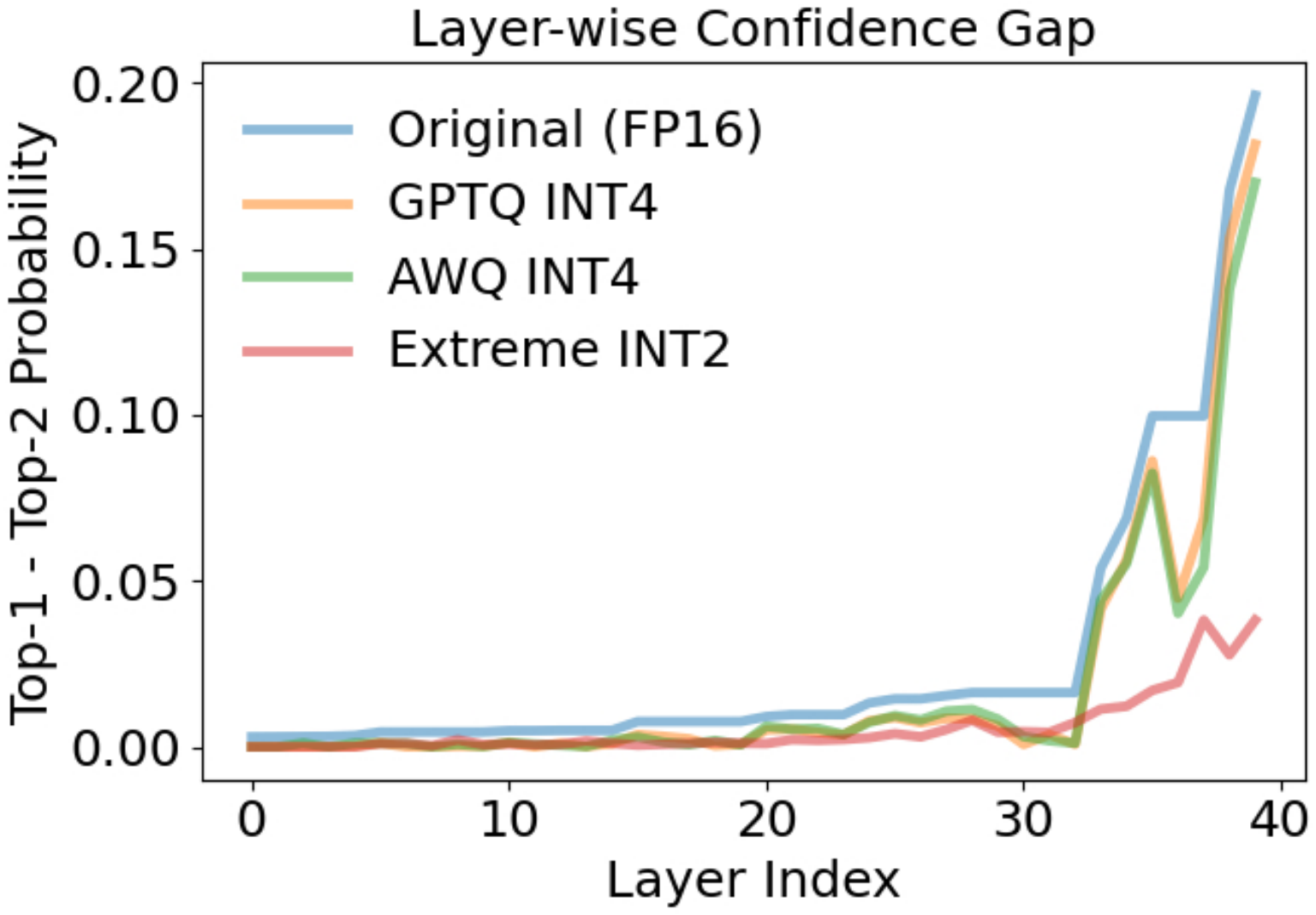}
\caption{Layer-wise Confidence Gap}
\label{fig:main-comparison-Mistral-Small-24B-Instruct-2501:gap}
\end{subfigure}
\caption{Main comparison (Mistral-Small-24B-Instruct-2501) across layers for four model variants (Original FP16, GPTQ INT4, AWQ INT4, Extreme INT2). Figure (a) reports layer-wise decisional entropy, where lower entropy indicates sharper, more stable decisions. Figure (b) reports the layer-wise confidence gap (top-1 minus top-2 probability), where a larger margin reflects stronger commitment to a single option.}
\label{fig:main-comparison-Mistral-Small-24B-Instruct-2501}
\end{figure*}

\begin{figure*}[t]
\centering
\begin{subfigure}[t]{0.48\linewidth}
\centering
\includegraphics[width=\linewidth]{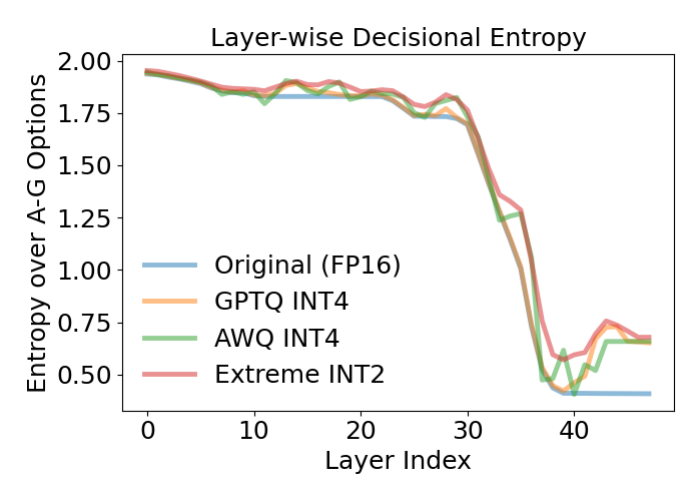}
\caption{Layer-wise Decisional Entropy}
\label{fig:main-comparison-Qwen2.5-14B-Instruct:entropy}
\end{subfigure}
\hfill
\begin{subfigure}[t]{0.48\linewidth}
\centering
\includegraphics[width=\linewidth]{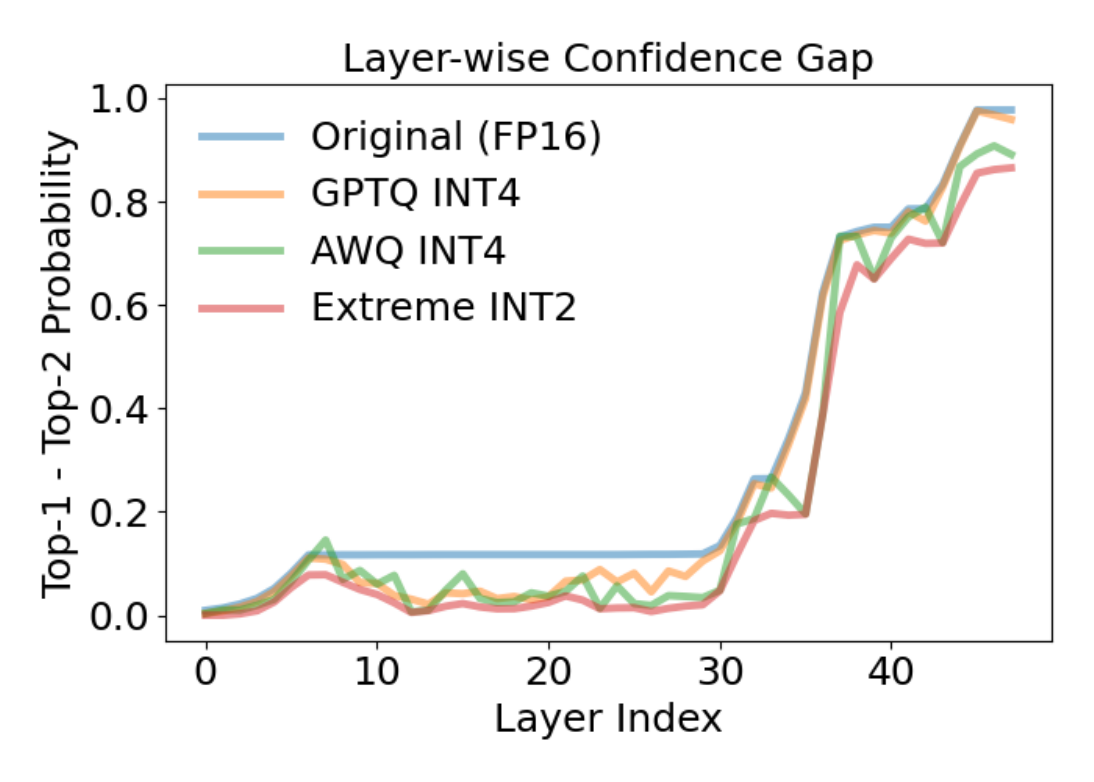}
\caption{Layer-wise Confidence Gap}
\label{fig:main-comparison-Qwen2.5-14B-Instruct:gap}
\end{subfigure}
\caption{Main comparison (Qwen2.5-14B-Instruct) across layers for four model variants (Original FP16, GPTQ INT4, AWQ INT4, Extreme INT2). Figure (a) reports layer-wise decisional entropy, where lower entropy indicates sharper, more stable decisions. Figure (b) reports the layer-wise confidence gap (top-1 minus top-2 probability), where a larger margin reflects stronger commitment to a single option.}
\label{fig:main-comparison-Qwen2.5-14B-Instruct}
\end{figure*}

\begin{figure*}[t]
\centering
\begin{subfigure}[t]{0.48\linewidth}
\centering
\includegraphics[width=\linewidth]{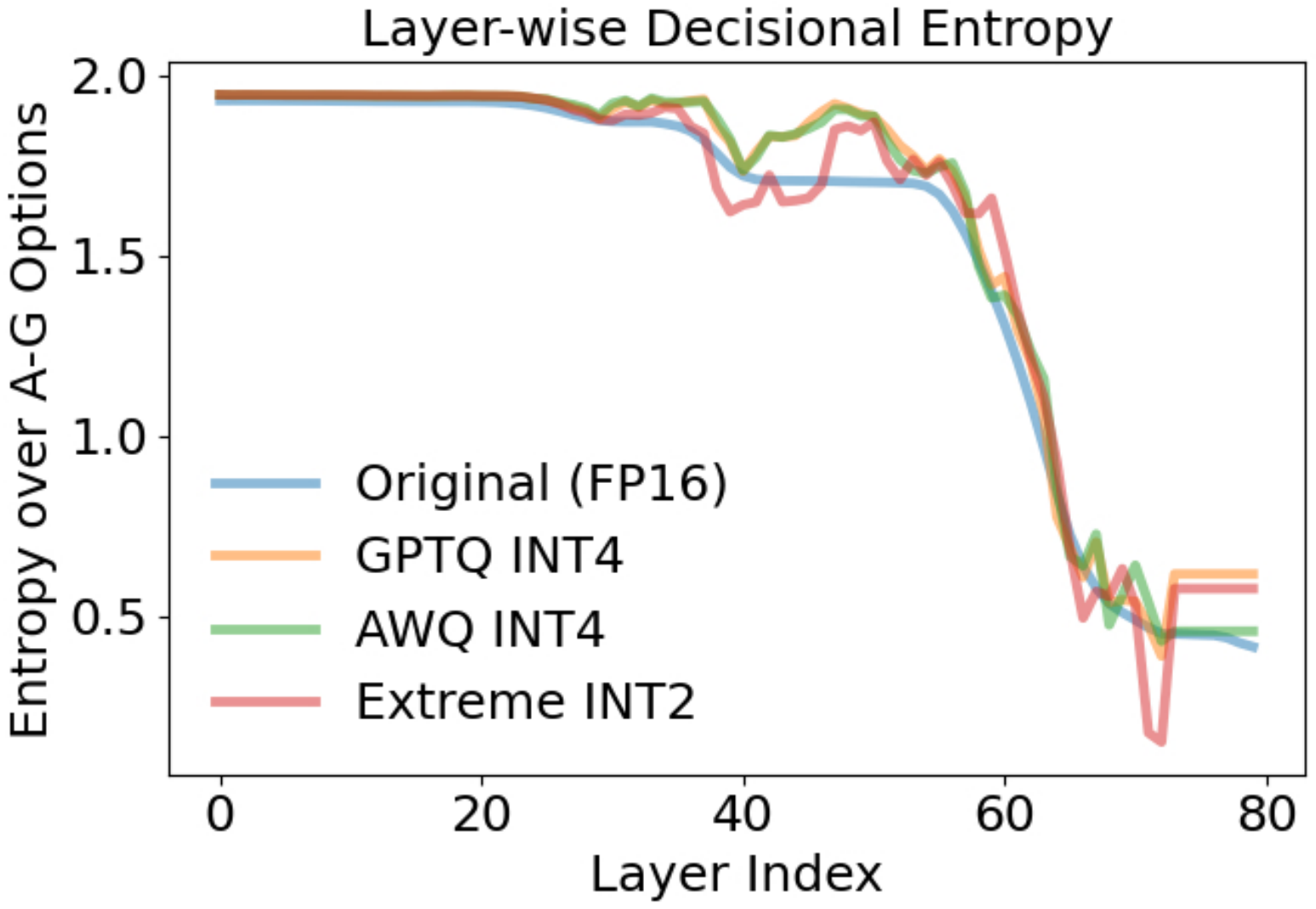}
\caption{Layer-wise Decisional Entropy}
\label{fig:main-comparison-Qwen2.5-72B-Instruct:entropy}
\end{subfigure}
\hfill
\begin{subfigure}[t]{0.48\linewidth}
\centering
\includegraphics[width=\linewidth]{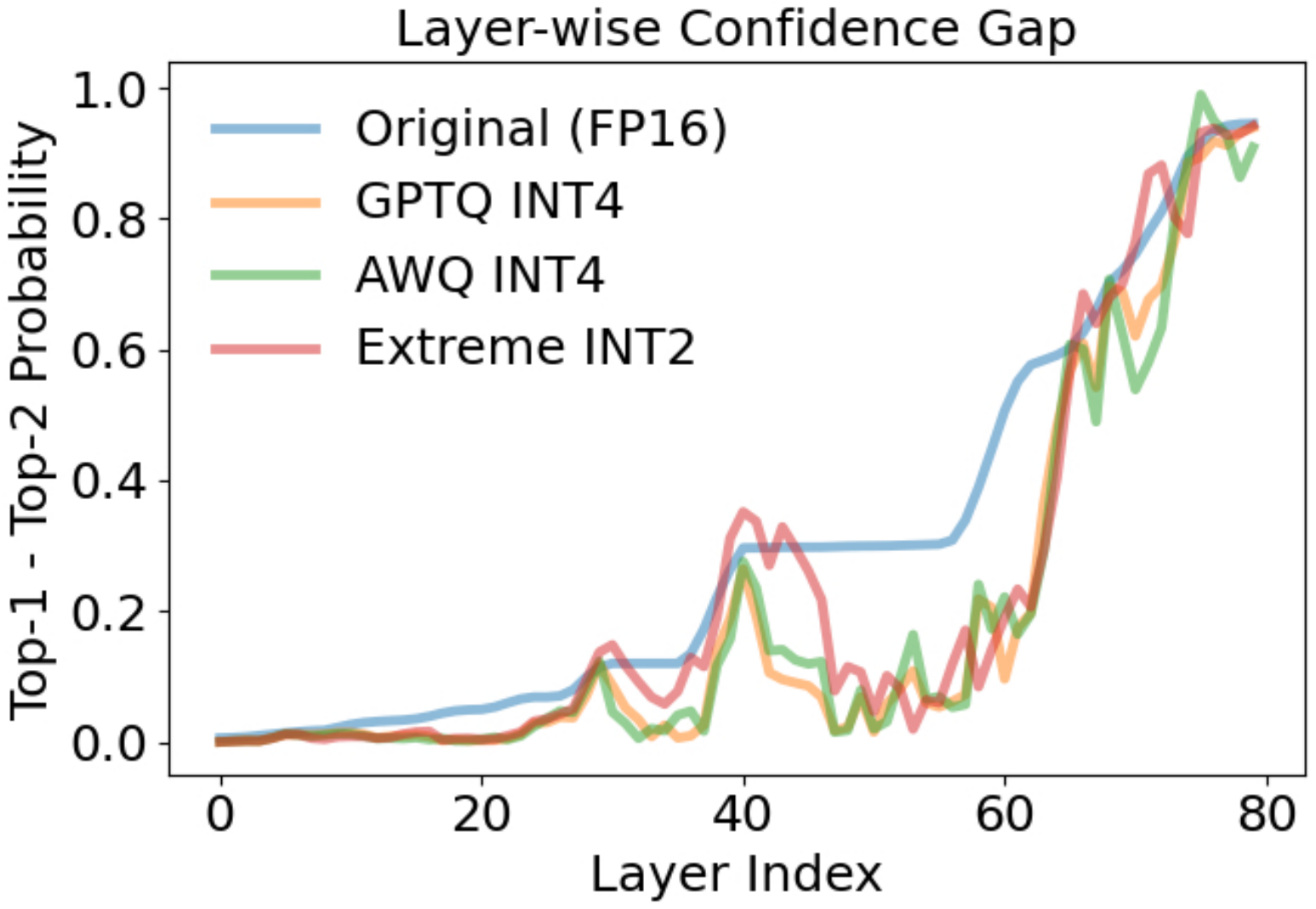}
\caption{Layer-wise Confidence Gap}
\label{fig:main-comparison-Qwen2.5-72B-Instruct:gap}
\end{subfigure}
\caption{Main comparison (Qwen2.5-72B-Instruct) across layers for four model variants (Original FP16, GPTQ INT4, AWQ INT4, Extreme INT2). Figure (a) reports layer-wise decisional entropy, where lower entropy indicates sharper, more stable decisions. Figure (b) reports the layer-wise confidence gap (top-1 minus top-2 probability), where a larger margin reflects stronger commitment to a single option.}
\label{fig:main-comparison-Qwen2.5-72B-Instruct}
\end{figure*}

\begin{table*}[t]
\centering
\small
\setlength{\tabcolsep}{6pt}
\caption{LLaMA3.1-70B-Instruct: Conditional personality consistency under prompt conditioning across quantization levels. Each entry reports the inferred MBTI type when the model is conditioned on a personality-specific prompt (e.g., ENFJ $\rightarrow$ ENFJ).
For each cell, the left marker indicates consistency with the conditioning prompt, while the right marker indicates agreement with the FP16 baseline prediction.}
\label{tab:LLaMA3.1-70B-Instruct_personality_consistency}
\begin{tabular}{cccc}
\toprule
 \textbf{Original FP16} & \textbf{GPTQ INT4} & \textbf{AWQ INT4} & \textbf{Extreme INT2} \\
\midrule
ENFJ $\rightarrow$ ENFJ $\checkmark\checkmark$ & ENFJ $\rightarrow$ ENFJ $\checkmark\checkmark$ & ENFJ $\rightarrow$ ENFJ $\checkmark\checkmark$ & ENFJ $\rightarrow$ ENFJ $\checkmark\checkmark$ \\
ENFP $\rightarrow$ ENFJ $\times\checkmark$ & ENFP $\rightarrow$ ENFP $\checkmark\times$ & ENFP $\rightarrow$ ENFP $\checkmark\times$ & ENFP $\rightarrow$ ENFP $\checkmark\times$ \\
ENTJ $\rightarrow$ ENFJ $\times\checkmark$ & ENTJ $\rightarrow$ ESTJ $\times\times$ & ENTJ $\rightarrow$ ESTJ $\times\times$ & ENTJ $\rightarrow$ ENTJ $\checkmark\times$ \\
ENTP $\rightarrow$ ENFJ $\times\checkmark$ & ENTP $\rightarrow$ ENTP $\checkmark\times$ & ENTP $\rightarrow$ ENTP $\checkmark\times$ & ENTP $\rightarrow$ ENTP $\checkmark\times$ \\

ESFJ $\rightarrow$ ENFJ $\times\checkmark$ & ESFJ $\rightarrow$ ESFJ $\checkmark\times$ & ESFJ $\rightarrow$ ESFJ $\checkmark\times$ & ESFJ $\rightarrow$ ESFJ $\checkmark\times$ \\
ESFP $\rightarrow$ ENFJ $\times\checkmark$ & ESFP $\rightarrow$ ENFP $\times\times$ & ESFP $\rightarrow$ ENFP $\times\times$ & ESFP $\rightarrow$ ENFP $\times\times$ \\

ESTJ $\rightarrow$ ENFJ $\times\checkmark$ & ESTJ $\rightarrow$ ESTJ $\checkmark\times$ & ESTJ $\rightarrow$ ESTJ $\checkmark\times$ & ESTJ $\rightarrow$ ESTJ $\checkmark\times$ \\
ESTP $\rightarrow$ ENFJ $\times\checkmark$ & ESTP $\rightarrow$ ESTP $\checkmark\times$ & ESTP $\rightarrow$ ESTP $\checkmark\times$ & ESTP $\rightarrow$ ESTP $\checkmark\times$ \\

INFJ $\rightarrow$ ENFJ $\times\checkmark$ & INFJ $\rightarrow$ INFJ $\checkmark\times$ & INFJ $\rightarrow$ INFJ $\checkmark\times$ & INFJ $\rightarrow$ INFJ $\checkmark\times$ \\
INFP $\rightarrow$ ENFJ $\times\checkmark$ & INFP $\rightarrow$ INFP $\checkmark\times$ & INFP $\rightarrow$ INFP $\checkmark\times$ & INFP $\rightarrow$ INFP $\checkmark\times$ \\

INTJ $\rightarrow$ ENFJ $\times\checkmark$ & INTJ $\rightarrow$ INTJ $\checkmark\times$ & INTJ $\rightarrow$ INTJ $\checkmark\times$ & INTJ $\rightarrow$ INTJ $\checkmark\times$ \\
INTP $\rightarrow$ ENFJ $\times\checkmark$ & INTP $\rightarrow$ INTP $\checkmark\times$ & INTP $\rightarrow$ INTP $\checkmark\times$ & INTP $\rightarrow$ INTJ $\times\times$ \\

ISFJ $\rightarrow$ ENFJ $\times\checkmark$ & ISFJ $\rightarrow$ ISFJ $\checkmark\times$ & ISFJ $\rightarrow$ ISFJ $\checkmark\times$ & ISFJ $\rightarrow$ ISFJ $\checkmark\times$ \\
ISFP $\rightarrow$ ENFJ $\times\checkmark$ & ISFP $\rightarrow$ INFP $\times\times$ & ISFP $\rightarrow$ INFP $\times\times$ & ISFP $\rightarrow$ INFP $\times\times$ \\

ISTJ $\rightarrow$ ENFJ $\times\checkmark$ & ISTJ $\rightarrow$ ISTJ $\checkmark\times$ & ISTJ $\rightarrow$ ISTJ $\checkmark\times$ & ISTJ $\rightarrow$ ISTJ $\checkmark\times$ \\
ISTP $\rightarrow$ ENFJ $\times\checkmark$ & ISTP $\rightarrow$ ISTJ $\times\times$ & ISTP $\rightarrow$ ISTJ $\times\times$ & ISTP $\rightarrow$ ISTJ $\times\times$ 
\\
\bottomrule
\end{tabular}
\end{table*}

\begin{table*}[t]
\centering
\small
\setlength{\tabcolsep}{6pt}
\caption{Mistral-7B-Instruct-v0.3: Conditional personality consistency under prompt conditioning across quantization levels. Each entry reports the inferred MBTI type when the model is conditioned on a personality-specific prompt (e.g., ENFJ $\rightarrow$ ENFJ).
For each cell, the left marker indicates consistency with the conditioning prompt, while the right marker indicates agreement with the FP16 baseline prediction.}
\label{tab:Mistral-7B-Instruct-v0.3_personality_consistency}
\begin{tabular}{cccc}
\toprule
 \textbf{Original FP16} & \textbf{GPTQ INT4} & \textbf{AWQ INT4} & \textbf{Extreme INT2} \\
\midrule

ENFJ $\rightarrow$ ENFJ $\checkmark\checkmark$ & ENFJ $\rightarrow$ ENFJ $\checkmark\checkmark$ & ENFJ $\rightarrow$ ENFJ $\checkmark\checkmark$ & ENFJ $\rightarrow$ ENFJ $\checkmark\checkmark$ \\
ENFP $\rightarrow$ ENFP $\checkmark\checkmark$ & ENFP $\rightarrow$ ENFP $\checkmark\checkmark$ & ENFP $\rightarrow$ ENFP $\checkmark\checkmark$ & ENFP $\rightarrow$ ENFJ $\times\times$ \\
ENTJ $\rightarrow$ ENTJ $\checkmark\checkmark$ & ENTJ $\rightarrow$ ENTJ $\checkmark\checkmark$ & ENTJ $\rightarrow$ ENTJ $\checkmark\checkmark$ & ENTJ $\rightarrow$ ENFJ $\times\times$ \\
ENTP $\rightarrow$ ENFP $\times\checkmark$ & ENTP $\rightarrow$ ENFP $\times\checkmark$ & ENTP $\rightarrow$ ENFP $\times\checkmark$ & ENTP $\rightarrow$ ENFJ $\times\times$ \\

ESFJ $\rightarrow$ ENFJ $\times\checkmark$ & ESFJ $\rightarrow$ ENFJ $\times\checkmark$ & ESFJ $\rightarrow$ ENFJ $\times\checkmark$ & ESFJ $\rightarrow$ ENFJ $\times\checkmark$ \\
ESFP $\rightarrow$ ENFP $\times\checkmark$ & ESFP $\rightarrow$ ENFP $\times\checkmark$ & ESFP $\rightarrow$ ENFP $\times\checkmark$ & ESFP $\rightarrow$ ENFJ $\times\times$ \\

ESTJ $\rightarrow$ ENTJ $\times\checkmark$ & ESTJ $\rightarrow$ ENTJ $\times\checkmark$ & ESTJ $\rightarrow$ ESTJ $\checkmark\times$ & ESTJ $\rightarrow$ ENFJ $\times\times$ \\
ESTP $\rightarrow$ ESTP $\checkmark\checkmark$ & ESTP $\rightarrow$ ESTP $\checkmark\checkmark$ & ESTP $\rightarrow$ ESTP $\checkmark\checkmark$ & ESTP $\rightarrow$ ENFJ $\times\times$ \\

INFJ $\rightarrow$ ENFJ $\times\checkmark$ & INFJ $\rightarrow$ ENFJ $\times\checkmark$ & INFJ $\rightarrow$ ENFJ $\times\checkmark$ & INFJ $\rightarrow$ ENFJ $\times\checkmark$ \\
INFP $\rightarrow$ ENFP $\times\checkmark$ & INFP $\rightarrow$ ENFP $\times\checkmark$ & INFP $\rightarrow$ ENFP $\times\checkmark$ & INFP $\rightarrow$ ENFJ $\times\times$ \\

INTJ $\rightarrow$ INTJ $\checkmark\checkmark$ & INTJ $\rightarrow$ INTJ $\checkmark\checkmark$ & INTJ $\rightarrow$ INTJ $\checkmark\checkmark$ & INTJ $\rightarrow$ ENFJ $\times\times$ \\
INTP $\rightarrow$ INTJ $\times\checkmark$ & INTP $\rightarrow$ INTJ $\times\checkmark$ & INTP $\rightarrow$ INTJ $\times\checkmark$ & INTP $\rightarrow$ ENFJ $\times\times$ \\

ISFJ $\rightarrow$ INFJ $\times\checkmark$ & ISFJ $\rightarrow$ ENFJ $\times\times$ & ISFJ $\rightarrow$ ENFJ $\times\times$ & ISFJ $\rightarrow$ ENFJ $\times\times$ \\
ISFP $\rightarrow$ INFJ $\times\checkmark$ & ISFP $\rightarrow$ INFJ $\times\checkmark$ & ISFP $\rightarrow$ INFP $\times\times$ & ISFP $\rightarrow$ ENFJ $\times\times$ \\

ISTJ $\rightarrow$ ISTJ $\checkmark\checkmark$ & ISTJ $\rightarrow$ INTJ $\times\times$ & ISTJ $\rightarrow$ ISTJ $\checkmark\checkmark$ & ISTJ $\rightarrow$ ENFJ $\times\times$ \\
ISTP $\rightarrow$ ISTJ $\times\checkmark$ & ISTP $\rightarrow$ INTJ $\times\times$ & ISTP $\rightarrow$ INTJ $\times\times$ & ISTP $\rightarrow$ ENFJ $\times\times$ \\
\bottomrule
\end{tabular}
\end{table*}

\begin{table*}[t]
\centering
\small
\setlength{\tabcolsep}{6pt}
\caption{Mistral-Small-24B-Instruct-2501: Conditional personality consistency under prompt conditioning across quantization levels. Each entry reports the inferred MBTI type when the model is conditioned on a personality-specific prompt (e.g., ENFJ $\rightarrow$ ENFJ).
For each cell, the left marker indicates consistency with the conditioning prompt, while the right marker indicates agreement with the FP16 baseline prediction.}
\label{tab:Mistral-Small-24B-Instruct-2501_personality_consistency}
\begin{tabular}{cccc}
\toprule
 \textbf{Original FP16} & \textbf{GPTQ INT4} & \textbf{AWQ INT4} & \textbf{Extreme INT2} \\
\midrule
ENFJ $\rightarrow$ ENFJ $\checkmark\checkmark$ & ENFJ $\rightarrow$ ENFJ $\checkmark\checkmark$ & ENFJ $\rightarrow$ ENFJ $\checkmark\checkmark$ & ENFJ $\rightarrow$ ENFJ $\checkmark\checkmark$ \\
ENFP $\rightarrow$ ENFP $\checkmark\checkmark$ & ENFP $\rightarrow$ ENFP $\checkmark\checkmark$ & ENFP $\rightarrow$ ENFP $\checkmark\checkmark$ & ENFP $\rightarrow$ ENFJ $\times\times$ \\
ENTJ $\rightarrow$ ESTJ $\times\checkmark$ & ENTJ $\rightarrow$ ESTJ $\times\checkmark$ & ENTJ $\rightarrow$ ESTJ $\times\checkmark$ & ENTJ $\rightarrow$ ESFJ $\times\times$ \\
ENTP $\rightarrow$ ENTP $\checkmark\checkmark$ & ENTP $\rightarrow$ ENTP $\checkmark\checkmark$ & ENTP $\rightarrow$ ENTP $\checkmark\checkmark$ & ENTP $\rightarrow$ ESTJ $\times\times$ \\

ESFJ $\rightarrow$ ESFJ $\checkmark\checkmark$ & ESFJ $\rightarrow$ ESFJ $\checkmark\checkmark$ & ESFJ $\rightarrow$ ESFJ $\checkmark\checkmark$ & ESFJ $\rightarrow$ ENFJ $\times\times$ \\
ESFP $\rightarrow$ ESFP $\checkmark\checkmark$ & ESFP $\rightarrow$ ESFP $\checkmark\checkmark$ & ESFP $\rightarrow$ ESFP $\checkmark\checkmark$ & ESFP $\rightarrow$ ENFJ $\times\times$ \\

ESTJ $\rightarrow$ ESTJ $\checkmark\checkmark$ & ESTJ $\rightarrow$ ESTJ $\checkmark\checkmark$ & ESTJ $\rightarrow$ ESTJ $\checkmark\checkmark$ & ESTJ $\rightarrow$ ENFJ $\times\times$ \\
ESTP $\rightarrow$ ESTP $\checkmark\checkmark$ & ESTP $\rightarrow$ ESTP $\checkmark\checkmark$ & ESTP $\rightarrow$ ESTP $\checkmark\checkmark$ & ESTP $\rightarrow$ ESFJ $\times\times$ \\

INFJ $\rightarrow$ INFJ $\checkmark\checkmark$ & INFJ $\rightarrow$ INFJ $\checkmark\checkmark$ & INFJ $\rightarrow$ INFJ $\checkmark\checkmark$ & INFJ $\rightarrow$ ENFJ $\times\times$ \\
INFP $\rightarrow$ INFP $\checkmark\checkmark$ & INFP $\rightarrow$ INFP $\checkmark\checkmark$ & INFP $\rightarrow$ INFP $\checkmark\checkmark$ & INFP $\rightarrow$ ENFJ $\times\times$ \\

INTJ $\rightarrow$ INTJ $\checkmark\checkmark$ & INTJ $\rightarrow$ INTJ $\checkmark\checkmark$ & INTJ $\rightarrow$ INTJ $\checkmark\checkmark$ & INTJ $\rightarrow$ ENFJ $\times\times$ \\
INTP $\rightarrow$ INTJ $\times\checkmark$ & INTP $\rightarrow$ INTJ $\times\checkmark$ & INTP $\rightarrow$ INTJ $\times\checkmark$ & INTP $\rightarrow$ ESFJ $\times\times$ \\

ISFJ $\rightarrow$ ISFJ $\checkmark\checkmark$ & ISFJ $\rightarrow$ ISFJ $\checkmark\checkmark$ & ISFJ $\rightarrow$ ISFJ $\checkmark\checkmark$ & ISFJ $\rightarrow$ ENFJ $\times\times$ \\
ISFP $\rightarrow$ ISFP $\checkmark\checkmark$ & ISFP $\rightarrow$ ISFP $\checkmark\checkmark$ & ISFP $\rightarrow$ INFP $\times\times$ & ISFP $\rightarrow$ ENFJ $\times\times$ \\

ISTJ $\rightarrow$ ISTJ $\checkmark\checkmark$ & ISTJ $\rightarrow$ ISTJ $\checkmark\checkmark$ & ISTJ $\rightarrow$ ISTJ $\checkmark\checkmark$ & ISTJ $\rightarrow$ ENFJ $\times\times$ \\
ISTP $\rightarrow$ ISTJ $\times\checkmark$ & ISTP $\rightarrow$ ISTJ $\times\checkmark$ & ISTP $\rightarrow$ ISTJ $\times\checkmark$ & ISTP $\rightarrow$ ENFJ $\times\times$ \\
\bottomrule
\end{tabular}
\end{table*}

\begin{table*}[t]
\centering
\small
\setlength{\tabcolsep}{6pt}
\caption{Qwen2.5-14B-Instruct: Conditional personality consistency under prompt conditioning across quantization levels. Each entry reports the inferred MBTI type when the model is conditioned on a personality-specific prompt (e.g., ENFJ $\rightarrow$ ENFJ).
For each cell, the left marker indicates consistency with the conditioning prompt, while the right marker indicates agreement with the FP16 baseline prediction.}
\label{tab:Qwen2.5-14B-Instruct_personality_consistency}
\begin{tabular}{cccc}
\toprule
 \textbf{Original FP16} & \textbf{GPTQ INT4} & \textbf{AWQ INT4} & \textbf{Extreme INT2} \\
\midrule
ENFJ $\rightarrow$ ENFJ $\checkmark\checkmark$ & ENFJ $\rightarrow$ ENFJ $\checkmark\checkmark$ & ENFJ $\rightarrow$ ENFJ $\checkmark\checkmark$ & ENFJ $\rightarrow$ ENFJ $\checkmark\checkmark$ \\
ENFP $\rightarrow$ ENFP $\checkmark\checkmark$ & ENFP $\rightarrow$ ENFP $\checkmark\checkmark$ & ENFP $\rightarrow$ ENFP $\checkmark\checkmark$ & ENFP $\rightarrow$ ENFJ $\times\times$ \\
ENTJ $\rightarrow$ ENTJ $\checkmark\checkmark$ & ENTJ $\rightarrow$ ENTJ $\checkmark\checkmark$ & ENTJ $\rightarrow$ ENTJ $\checkmark\checkmark$ & ENTJ $\rightarrow$ ENFJ $\times\times$ \\
ENTP $\rightarrow$ ENTP $\checkmark\checkmark$ & ENTP $\rightarrow$ ENTP $\checkmark\checkmark$ & ENTP $\rightarrow$ ENTP $\checkmark\checkmark$ & ENTP $\rightarrow$ ENFJ $\times\times$ \\

ESFJ $\rightarrow$ ESFJ $\checkmark\checkmark$ & ESFJ $\rightarrow$ ESFJ $\checkmark\checkmark$ & ESFJ $\rightarrow$ ESFJ $\checkmark\checkmark$ & ESFJ $\rightarrow$ ENFJ $\times\times$ \\
ESFP $\rightarrow$ ENFP $\times\checkmark$ & ESFP $\rightarrow$ ENFP $\times\checkmark$ & ESFP $\rightarrow$ ENFP $\times\checkmark$ & ESFP $\rightarrow$ ENFJ $\times\times$ \\

ESTJ $\rightarrow$ ESTJ $\checkmark\checkmark$ & ESTJ $\rightarrow$ ESTJ $\checkmark\checkmark$ & ESTJ $\rightarrow$ ESTJ $\checkmark\checkmark$ & ESTJ $\rightarrow$ ENFJ $\times\times$ \\
ESTP $\rightarrow$ ESTP $\checkmark\checkmark$ & ESTP $\rightarrow$ ESTP $\checkmark\checkmark$ & ESTP $\rightarrow$ ESTP $\checkmark\checkmark$ & ESTP $\rightarrow$ ENFJ $\times\times$ \\

INFJ $\rightarrow$ INFJ $\checkmark\checkmark$ & INFJ $\rightarrow$ INFJ $\checkmark\checkmark$ & INFJ $\rightarrow$ ENFJ $\times\times$ & INFJ $\rightarrow$ ENFJ $\times\times$ \\
INFP $\rightarrow$ INFP $\checkmark\checkmark$ & INFP $\rightarrow$ INFP $\checkmark\checkmark$ & INFP $\rightarrow$ ENFP $\times\times$ & INFP $\rightarrow$ ENFJ $\times\times$ \\

INTJ $\rightarrow$ INTJ $\checkmark\checkmark$ & INTJ $\rightarrow$ INTJ $\checkmark\checkmark$ & INTJ $\rightarrow$ INTJ $\checkmark\checkmark$ & INTJ $\rightarrow$ ENFJ $\times\times$ \\
INTP $\rightarrow$ INTJ $\times\checkmark$ & INTP $\rightarrow$ INTJ $\times\checkmark$ & INTP $\rightarrow$ INTJ $\times\checkmark$ & INTP $\rightarrow$ ENFJ $\times\times$ \\

ISFJ $\rightarrow$ ISFJ $\checkmark\checkmark$ & ISFJ $\rightarrow$ ISFJ $\checkmark\checkmark$ & ISFJ $\rightarrow$ ISFJ $\checkmark\checkmark$ & ISFJ $\rightarrow$ ENFJ $\times\times$ \\
ISFP $\rightarrow$ ISFP $\checkmark\checkmark$ & ISFP $\rightarrow$ ISFJ $\times\times$ & ISFP $\rightarrow$ INFP $\times\times$ & ISFP $\rightarrow$ ENFJ $\times\times$ \\

ISTJ $\rightarrow$ ISTJ $\checkmark\checkmark$ & ISTJ $\rightarrow$ ISTJ $\checkmark\checkmark$ & ISTJ $\rightarrow$ ISTJ $\checkmark\checkmark$ & ISTJ $\rightarrow$ ENFJ $\times\times$ \\
ISTP $\rightarrow$ ISTJ $\times\checkmark$ & ISTP $\rightarrow$ ISTJ $\times\checkmark$ & ISTP $\rightarrow$ ISTJ $\times\checkmark$ & ISTP $\rightarrow$ ENFJ $\times\times$ \\
\bottomrule
\end{tabular}
\end{table*}

\begin{table*}[t]
\centering
\small
\setlength{\tabcolsep}{6pt}
\caption{Qwen2.5-72B-Instruct: Conditional personality consistency under prompt conditioning across quantization levels. Each entry reports the inferred MBTI type when the model is conditioned on a personality-specific prompt (e.g., ENFJ $\rightarrow$ ENFJ).
For each cell, the left marker indicates consistency with the conditioning prompt, while the right marker indicates agreement with the FP16 baseline prediction.}
\label{tab:Qwen2.5-72B-Instruct_personality_consistency}
\begin{tabular}{cccc}
\toprule
 \textbf{Original FP16} & \textbf{GPTQ INT4} & \textbf{AWQ INT4} & \textbf{Extreme INT2} \\
\midrule
ENFJ $\rightarrow$ ENFJ $\checkmark\checkmark$ & ENFJ $\rightarrow$ ENFJ $\checkmark\checkmark$ & ENFJ $\rightarrow$ ENFJ $\checkmark\checkmark$ & ENFJ $\rightarrow$ ENFJ $\checkmark\checkmark$ \\
ENFP $\rightarrow$ ENFJ $\times\checkmark$ & ENFP $\rightarrow$ ENFP $\checkmark\times$ & ENFP $\rightarrow$ ENFP $\checkmark\times$ & ENFP $\rightarrow$ ENFJ $\times\checkmark$ \\
ENTJ $\rightarrow$ ENFJ $\times\checkmark$ & ENTJ $\rightarrow$ ENTJ $\checkmark\times$ & ENTJ $\rightarrow$ ENTJ $\checkmark\times$ & ENTJ $\rightarrow$ ENTJ $\checkmark\times$ \\
ENTP $\rightarrow$ ENFJ $\times\checkmark$ & ENTP $\rightarrow$ ENFP $\times\times$ & ENTP $\rightarrow$ ENTP $\checkmark\times$ & ENTP $\rightarrow$ ENFJ $\times\checkmark$ \\

ESFJ $\rightarrow$ ENFJ $\times\checkmark$ & ESFJ $\rightarrow$ ENFJ $\times\checkmark$ & ESFJ $\rightarrow$ ENFJ $\times\checkmark$ & ESFJ $\rightarrow$ ENFJ $\times\checkmark$ \\
ESFP $\rightarrow$ ENFJ $\times\checkmark$ & ESFP $\rightarrow$ ENFP $\times\times$ & ESFP $\rightarrow$ ENFP $\times\times$ & ESFP $\rightarrow$ ENFJ $\times\checkmark$ \\

ESTJ $\rightarrow$ ENFJ $\times\checkmark$ & ESTJ $\rightarrow$ ESTJ $\checkmark\times$ & ESTJ $\rightarrow$ ESTJ $\checkmark\times$ & ESTJ $\rightarrow$ ENTJ $\times\times$ \\
ESTP $\rightarrow$ ENFJ $\times\checkmark$ & ESTP $\rightarrow$ ESTP $\checkmark\times$ & ESTP $\rightarrow$ ESTP $\checkmark\times$ & ESTP $\rightarrow$ ENFJ $\times\checkmark$ \\

INFJ $\rightarrow$ ENFJ $\times\checkmark$ & INFJ $\rightarrow$ INFJ $\checkmark\times$ & INFJ $\rightarrow$ INFJ $\checkmark\times$ & INFJ $\rightarrow$ ENFJ $\times\checkmark$ \\
INFP $\rightarrow$ ENFJ $\times\checkmark$ & INFP $\rightarrow$ INFP $\checkmark\times$ & INFP $\rightarrow$ INFP $\checkmark\times$ & INFP $\rightarrow$ ENFJ $\times\checkmark$ \\

INTJ $\rightarrow$ ENFJ $\times\checkmark$ & INTJ $\rightarrow$ INTJ $\checkmark\times$ & INTJ $\rightarrow$ INTJ $\checkmark\times$ & INTJ $\rightarrow$ ENTJ $\times\times$ \\
INTP $\rightarrow$ ENFJ $\times\checkmark$ & INTP $\rightarrow$ INTJ $\times\times$ & INTP $\rightarrow$ INTJ $\times\times$ & INTP $\rightarrow$ ENTJ $\times\times$ \\

ISFJ $\rightarrow$ ENFJ $\times\checkmark$ & ISFJ $\rightarrow$ INFJ $\times\times$ & ISFJ $\rightarrow$ INFJ $\times\times$ & ISFJ $\rightarrow$ ENFJ $\times\checkmark$ \\
ISFP $\rightarrow$ ENFJ $\times\checkmark$ & ISFP $\rightarrow$ INFP $\times\times$ & ISFP $\rightarrow$ INFP $\times\times$ & ISFP $\rightarrow$ ENFJ $\times\checkmark$ \\

ISTJ $\rightarrow$ ENFJ $\times\checkmark$ & ISTJ $\rightarrow$ ISTJ $\checkmark\times$ & ISTJ $\rightarrow$ ISTJ $\checkmark\times$ & ISTJ $\rightarrow$ ENFJ $\times\checkmark$ \\
ISTP $\rightarrow$ ENFJ $\times\checkmark$ & ISTP $\rightarrow$ ISTJ $\times\times$ & ISTP $\rightarrow$ ISTJ $\times\times$ & ISTP $\rightarrow$ ENTJ $\times\times$ \\
\bottomrule
\end{tabular}
\end{table*}

\begin{table}[t]
\centering
\caption{
Switch statistics for LLaMA3.1-8B-Instruct under personality-conditioned prompting across quantization levels.
Prompt-switch counts measure how often the inferred MBTI type differs from the conditioning prompt.
FP16-switch counts measure how often each quantized model differs from the Original FP16 prediction under the same conditioning prompt.
Means and standard errors are computed over the 16 MBTI-conditioned prompts.
}
\label{tab:llama31_8b_conditioned_switch_stats}
\begin{tabular}{lcccc}
\toprule
\textbf{Precision} 
& \textbf{Prompt-switch} 
& \textbf{Prompt-switch} 
& \textbf{FP16-switch} 
& \textbf{FP16-switch} \\
& \textbf{count / 16} 
& \textbf{mean $\pm$ SE} 
& \textbf{count / 16} 
& \textbf{mean $\pm$ SE} \\
\midrule
Original FP16 & 9 / 16  & $0.563 \pm 0.124$ & 0 / 16 & $0.000 \pm 0.000$ \\
GPTQ INT4     & 7 / 16  & $0.438 \pm 0.124$ & 3 / 16 & $0.188 \pm 0.098$ \\
AWQ INT4      & 8 / 16  & $0.500 \pm 0.125$ & 3 / 16 & $0.188 \pm 0.098$ \\
Extreme INT2  & 11 / 16 & $0.688 \pm 0.116$ & 5 / 16 & $0.313 \pm 0.116$ \\
\bottomrule
\end{tabular}
\end{table}

\begin{table}[t]
\centering
\caption{
Switch statistics for LLaMA3.1-70B-Instruct under personality-conditioned prompting across quantization levels.
Prompt-switch counts measure how often the inferred MBTI type differs from the conditioning prompt.
FP16-switch counts measure how often each quantized model differs from the Original FP16 prediction under the same conditioning prompt.
Means and standard errors are computed over the 16 MBTI-conditioned prompts.
}
\label{tab:llama31_70b_conditioned_switch_stats}
\begin{tabular}{lcccc}
\toprule
\textbf{Precision} 
& \textbf{Prompt-switch} 
& \textbf{Prompt-switch} 
& \textbf{FP16-switch} 
& \textbf{FP16-switch} \\
& \textbf{count / 16} 
& \textbf{mean $\pm$ SE} 
& \textbf{count / 16} 
& \textbf{mean $\pm$ SE} \\
\midrule
Original FP16 & 15 / 16 & $0.938 \pm 0.061$ & 0 / 16  & $0.000 \pm 0.000$ \\
GPTQ INT4     & 4 / 16  & $0.250 \pm 0.108$ & 15 / 16 & $0.938 \pm 0.061$ \\
AWQ INT4      & 4 / 16  & $0.250 \pm 0.108$ & 15 / 16 & $0.938 \pm 0.061$ \\
Extreme INT2  & 4 / 16  & $0.250 \pm 0.108$ & 15 / 16 & $0.938 \pm 0.061$ \\
\bottomrule
\end{tabular}
\end{table}

\begin{table}[t]
\centering
\caption{
Switch statistics for Mistral-7B-Instruct-v0.3 under personality-conditioned prompting across quantization levels.
Prompt-switch counts measure how often the inferred MBTI type differs from the conditioning prompt.
FP16-switch counts measure how often each quantized model differs from the Original FP16 prediction under the same conditioning prompt.
Means and standard errors are computed over the 16 MBTI-conditioned prompts.
}
\label{tab:mistral_7b_conditioned_switch_stats}
\begin{tabular}{lcccc}
\toprule
\textbf{Precision} 
& \textbf{Prompt-switch} 
& \textbf{Prompt-switch} 
& \textbf{FP16-switch} 
& \textbf{FP16-switch} \\
& \textbf{count / 16} 
& \textbf{mean $\pm$ SE} 
& \textbf{count / 16} 
& \textbf{mean $\pm$ SE} \\
\midrule
Original FP16 & 10 / 16 & $0.625 \pm 0.121$ & 0 / 16  & $0.000 \pm 0.000$ \\
GPTQ INT4     & 11 / 16 & $0.688 \pm 0.116$ & 3 / 16  & $0.188 \pm 0.098$ \\
AWQ INT4      & 9 / 16  & $0.563 \pm 0.124$ & 4 / 16  & $0.250 \pm 0.108$ \\
Extreme INT2  & 15 / 16 & $0.938 \pm 0.061$ & 13 / 16 & $0.813 \pm 0.098$ \\
\bottomrule
\end{tabular}
\end{table}

\begin{table}[t]
\centering
\caption{
Switch statistics for Mistral-Small-24B-Instruct-2501 under personality-conditioned prompting across quantization levels.
Prompt-switch counts measure how often the inferred MBTI type differs from the conditioning prompt.
FP16-switch counts measure how often each quantized model differs from the Original FP16 prediction under the same conditioning prompt.
Means and standard errors are computed over the 16 MBTI-conditioned prompts.
}
\label{tab:mistral_small_24b_conditioned_switch_stats}
\begin{tabular}{lcccc}
\toprule
\textbf{Precision} 
& \textbf{Prompt-switch} 
& \textbf{Prompt-switch} 
& \textbf{FP16-switch} 
& \textbf{FP16-switch} \\
& \textbf{count / 16} 
& \textbf{mean $\pm$ SE} 
& \textbf{count / 16} 
& \textbf{mean $\pm$ SE} \\
\midrule
Original FP16 & 3 / 16  & $0.188 \pm 0.098$ & 0 / 16  & $0.000 \pm 0.000$ \\
GPTQ INT4     & 3 / 16  & $0.188 \pm 0.098$ & 0 / 16  & $0.000 \pm 0.000$ \\
AWQ INT4      & 4 / 16  & $0.250 \pm 0.108$ & 1 / 16  & $0.063 \pm 0.061$ \\
Extreme INT2  & 15 / 16 & $0.938 \pm 0.061$ & 15 / 16 & $0.938 \pm 0.061$ \\
\bottomrule
\end{tabular}
\end{table}

\begin{table}[t]
\centering
\caption{
Switch statistics for Qwen2.5-14B-Instruct under personality-conditioned prompting across quantization levels.
Prompt-switch counts measure how often the inferred MBTI type differs from the conditioning prompt.
FP16-switch counts measure how often each quantized model differs from the Original FP16 prediction under the same conditioning prompt.
Means and standard errors are computed over the 16 MBTI-conditioned prompts.
}
\label{tab:qwen25_14b_conditioned_switch_stats}
\begin{tabular}{lcccc}
\toprule
\textbf{Precision}
& \textbf{Prompt-switch}
& \textbf{Prompt-switch}
& \textbf{FP16-switch}
& \textbf{FP16-switch} \\
& \textbf{count / 16}
& \textbf{mean $\pm$ SE}
& \textbf{count / 16}
& \textbf{mean $\pm$ SE} \\
\midrule
Original FP16 & 3 / 16  & $0.188 \pm 0.098$ & 0 / 16  & $0.000 \pm 0.000$ \\
GPTQ INT4     & 4 / 16  & $0.250 \pm 0.108$ & 1 / 16  & $0.063 \pm 0.061$ \\
AWQ INT4      & 6 / 16  & $0.375 \pm 0.121$ & 3 / 16  & $0.188 \pm 0.098$ \\
Extreme INT2  & 15 / 16 & $0.938 \pm 0.061$ & 15 / 16 & $0.938 \pm 0.061$ \\
\bottomrule
\end{tabular}
\end{table}

\begin{table}[t]
\centering
\caption{
Switch statistics for Qwen2.5-72B-Instruct under personality-conditioned prompting across quantization levels.
Prompt-switch counts measure how often the inferred MBTI type differs from the conditioning prompt.
FP16-switch counts measure how often each quantized model differs from the Original FP16 prediction under the same conditioning prompt.
Means and standard errors are computed over the 16 MBTI-conditioned prompts.
}
\label{tab:qwen25_72b_conditioned_switch_stats}
\begin{tabular}{lcccc}
\toprule
\textbf{Precision}
& \textbf{Prompt-switch}
& \textbf{Prompt-switch}
& \textbf{FP16-switch}
& \textbf{FP16-switch} \\
& \textbf{count / 16}
& \textbf{mean $\pm$ SE}
& \textbf{count / 16}
& \textbf{mean $\pm$ SE} \\
\midrule
Original FP16 & 15 / 16 & $0.938 \pm 0.061$ & 0 / 16  & $0.000 \pm 0.000$ \\
GPTQ INT4     & 7 / 16  & $0.438 \pm 0.124$ & 14 / 16 & $0.875 \pm 0.083$ \\
AWQ INT4      & 6 / 16  & $0.375 \pm 0.121$ & 14 / 16 & $0.875 \pm 0.083$ \\
Extreme INT2  & 14 / 16 & $0.875 \pm 0.083$ & 5 / 16  & $0.313 \pm 0.116$ \\
\bottomrule
\end{tabular}
\end{table}

\begin{figure*}[t]
\centering

\begin{subfigure}[t]{0.24\linewidth}
\centering
\includegraphics[width=\linewidth]{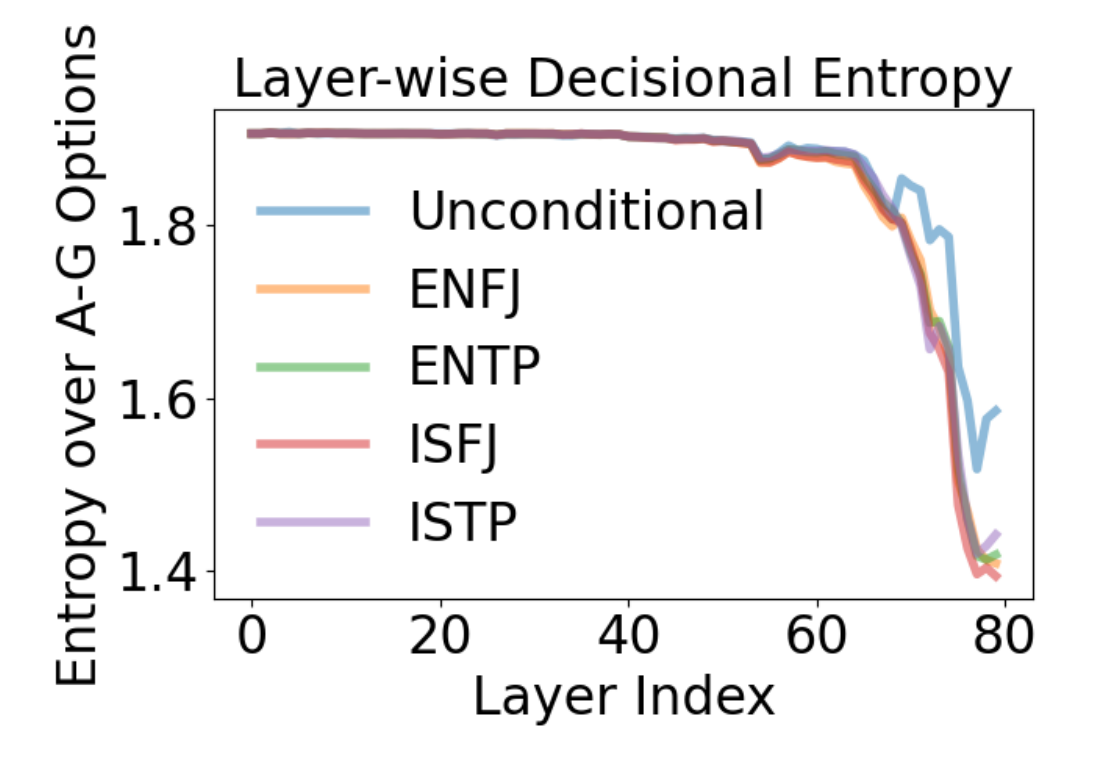}
\caption{FP16}
\label{fig:LLaMA3.1-70B-Instruct-conditional-entropy-breakdown:a}
\end{subfigure}
\hfill
\begin{subfigure}[t]{0.24\linewidth}
\centering
\includegraphics[width=\linewidth]{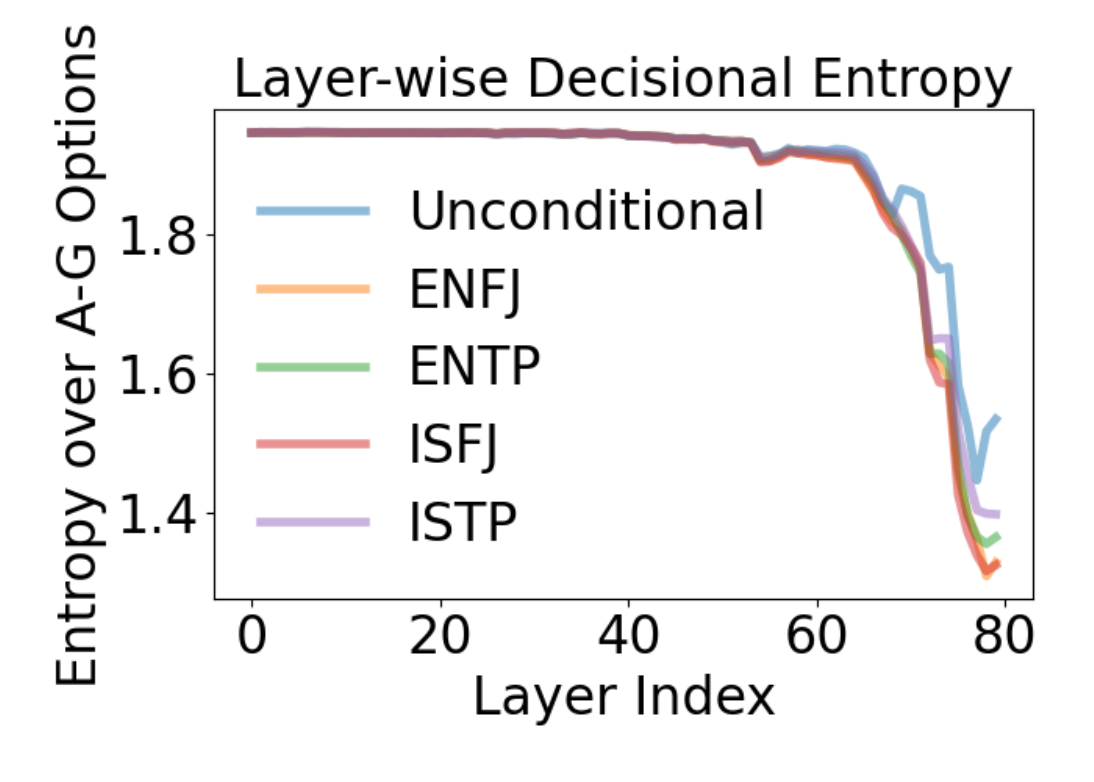}
\caption{GPTQ INT4}
\label{fig:LLaMA3.1-70B-Instruct-conditional-entropy-breakdown:b}
\end{subfigure}
\hfill
\begin{subfigure}[t]{0.24\linewidth}
\centering
\includegraphics[width=\linewidth]{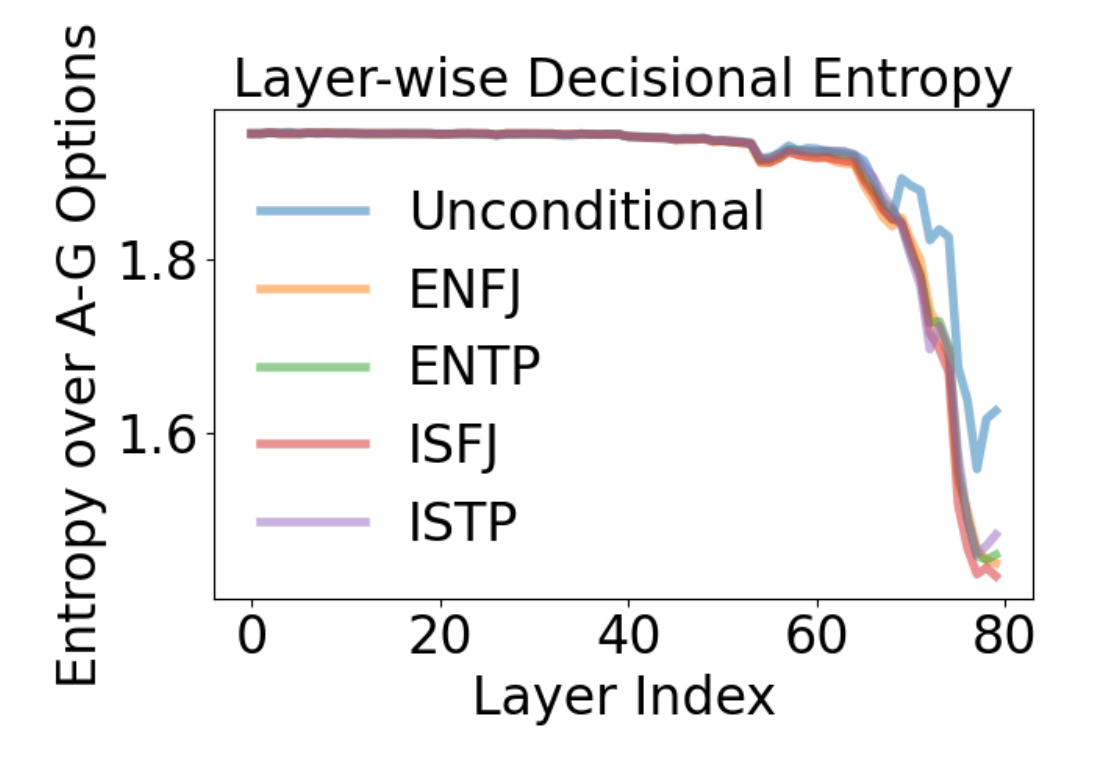}
\caption{AWQ INT4}
\label{fig:LLaMA3.1-70B-Instruct-conditional-entropy-breakdown:c}
\end{subfigure}
\hfill
\begin{subfigure}[t]{0.24\linewidth}
\centering
\includegraphics[width=\linewidth]{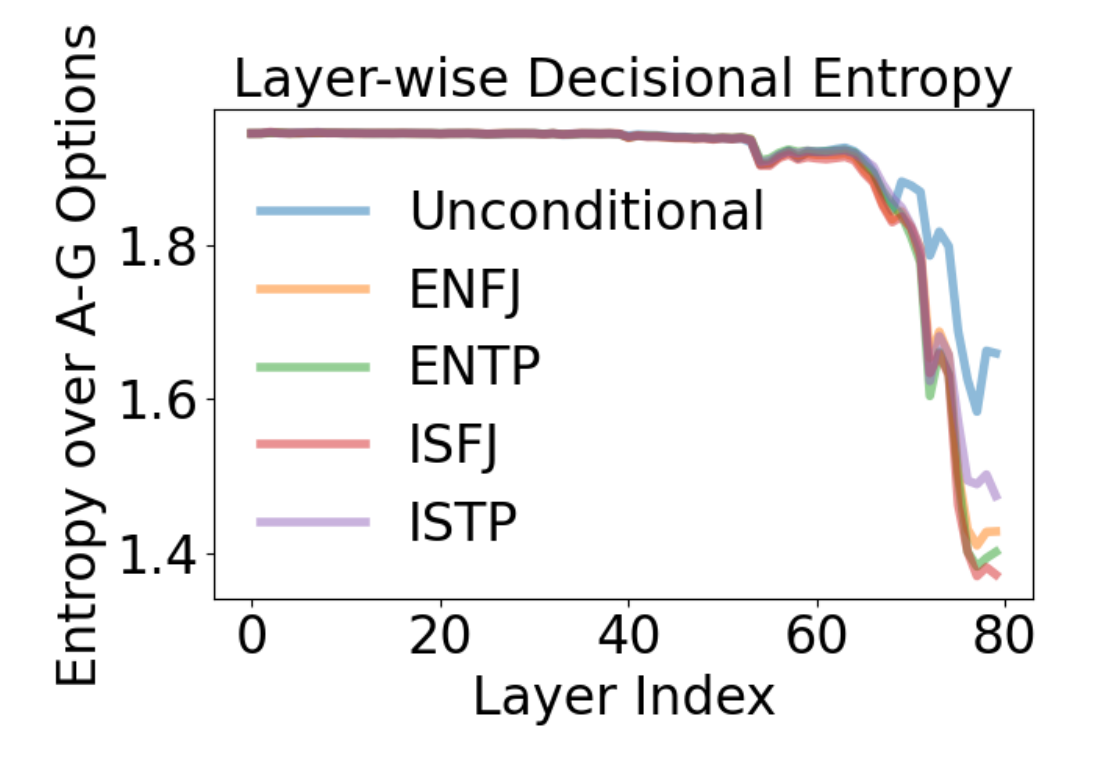}
\caption{AQLM 2-bit}
\label{fig:LLaMA3.1-70B-Instruct-conditional-entropy-breakdown:d}
\end{subfigure}

\vspace{0.5em}

\begin{subfigure}[t]{0.24\linewidth}
\centering
\includegraphics[width=\linewidth]{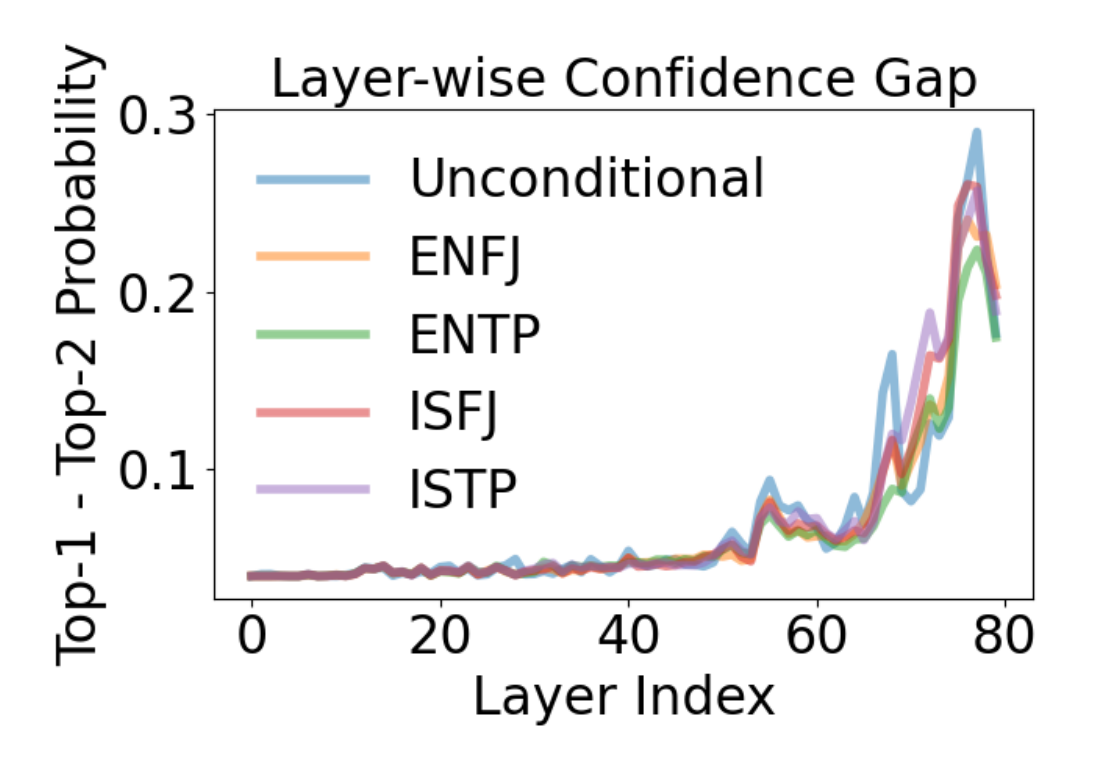}
\caption{FP16}
\label{fig:LLaMA3.1-70B-Instruct-conditional-gap-breakdown:a}
\end{subfigure}
\hfill
\begin{subfigure}[t]{0.24\linewidth}
\centering
\includegraphics[width=\linewidth]{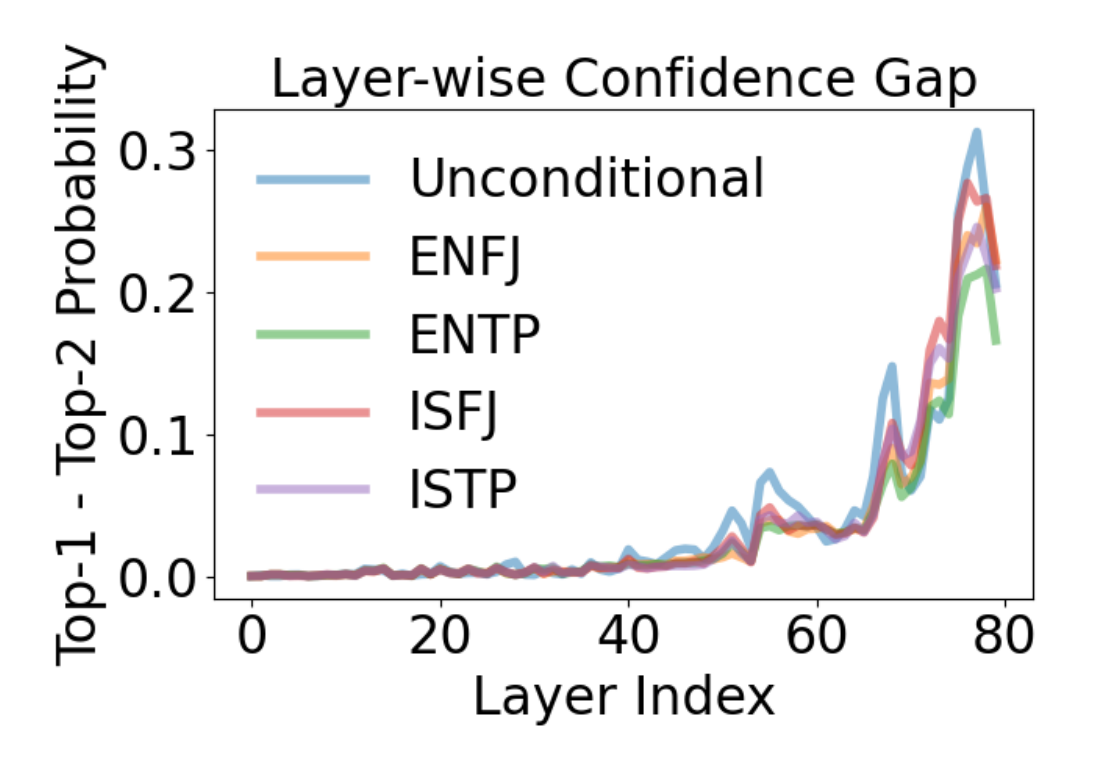}
\caption{GPTQ INT4}
\label{fig:LLaMA3.1-70B-Instruct-conditional-gap-breakdown:b}
\end{subfigure}
\hfill
\begin{subfigure}[t]{0.24\linewidth}
\centering
\includegraphics[width=\linewidth]{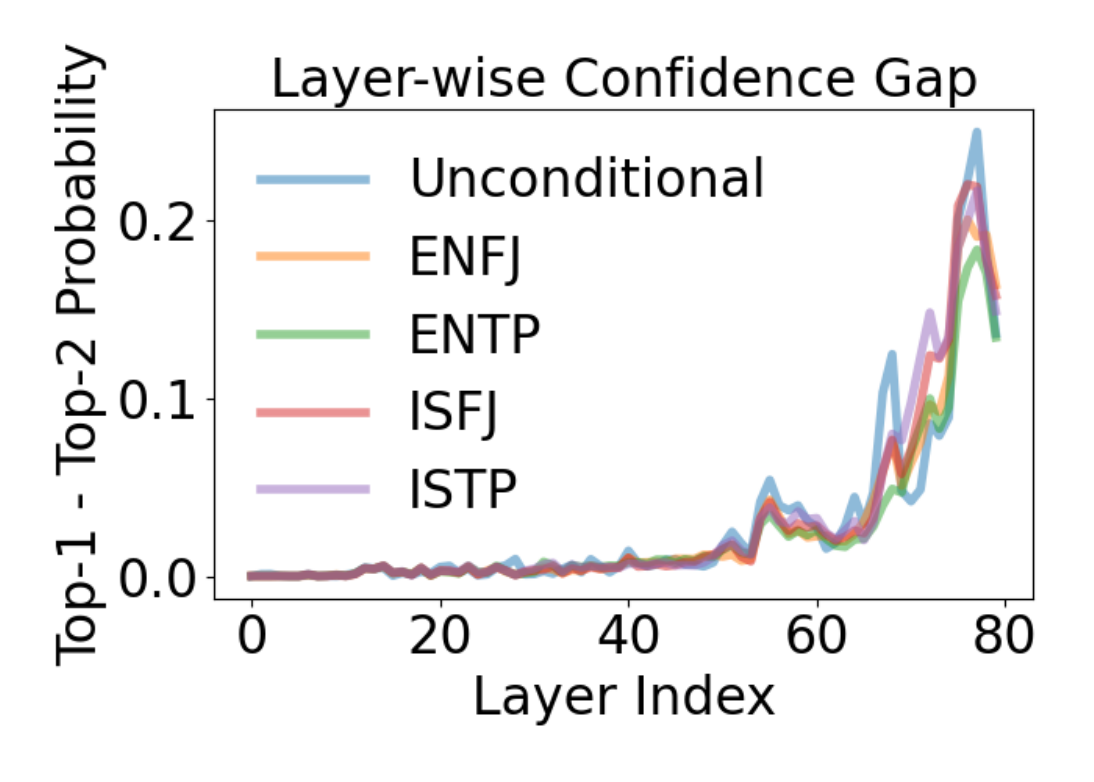}
\caption{AWQ INT4}
\label{fig:LLaMA3.1-70B-Instruct-conditional-gap-breakdown:c}
\end{subfigure}
\hfill
\begin{subfigure}[t]{0.24\linewidth}
\centering
\includegraphics[width=\linewidth]{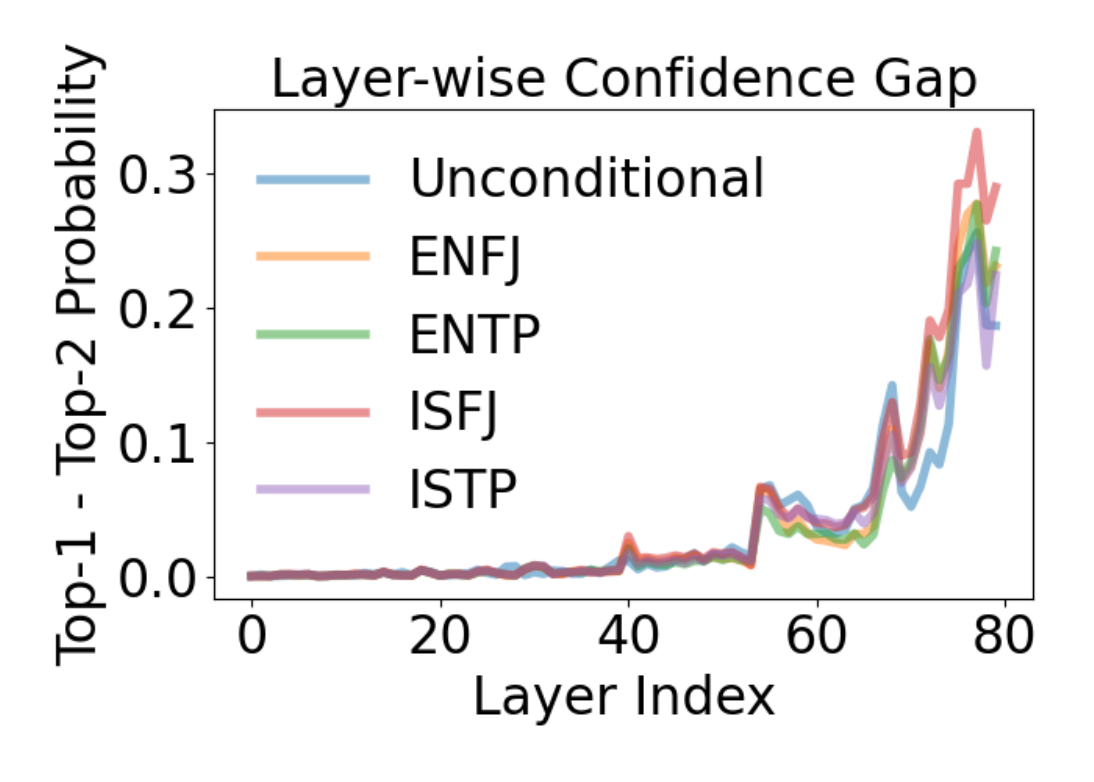}
\caption{AQLM 2-bit}
\label{fig:LLaMA3.1-70B-Instruct-conditional-gap-breakdown:d}
\end{subfigure}

\caption{LLaMA3.1-70B-Instruct: layer-wise decisional entropy (top row) and top-1 vs. top-2 probability gap (bottom row) across quantization variants. Early layers remain highly uncertain, while deeper layers progressively sharpen decisions.}
\label{fig:LLaMA3.1-70B-Instruct-conditional-combined}
\end{figure*}

\begin{figure*}[t]
\centering

\begin{subfigure}[t]{0.24\linewidth}
\centering
\includegraphics[width=\linewidth]{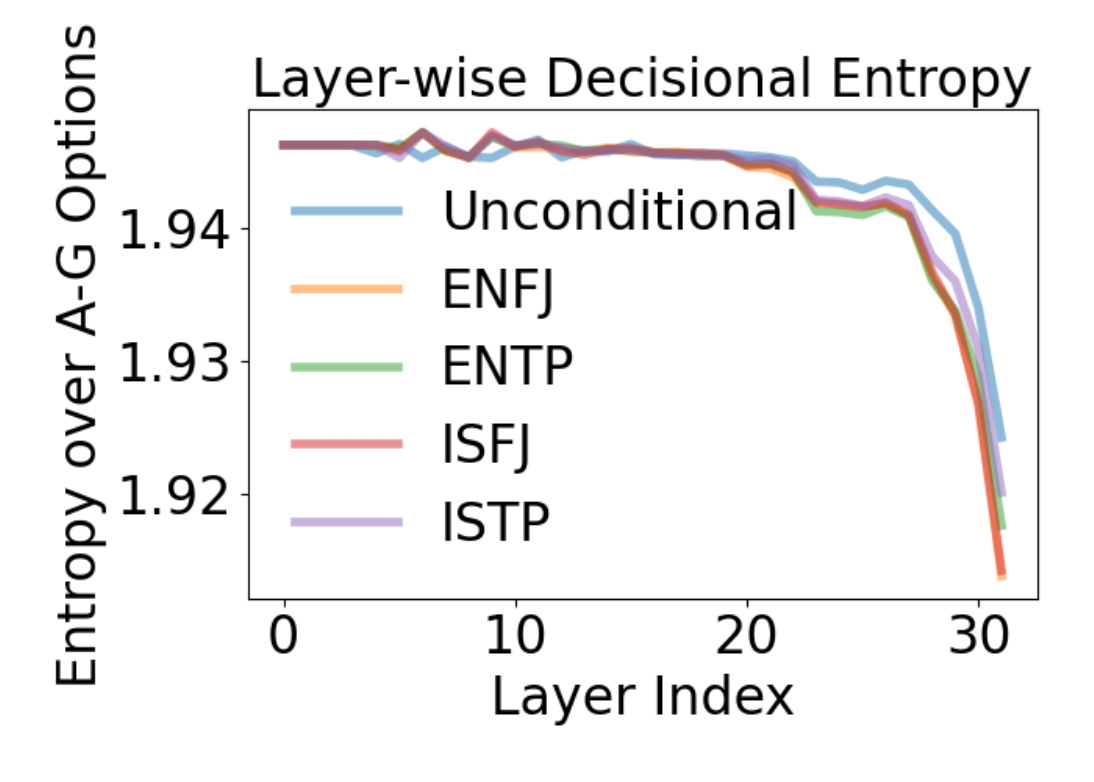}
\caption{FP16}
\label{fig:Mistral-7B-Instruct-v0.3-conditional-entropy-breakdown:a}
\end{subfigure}
\hfill
\begin{subfigure}[t]{0.24\linewidth}
\centering
\includegraphics[width=\linewidth]{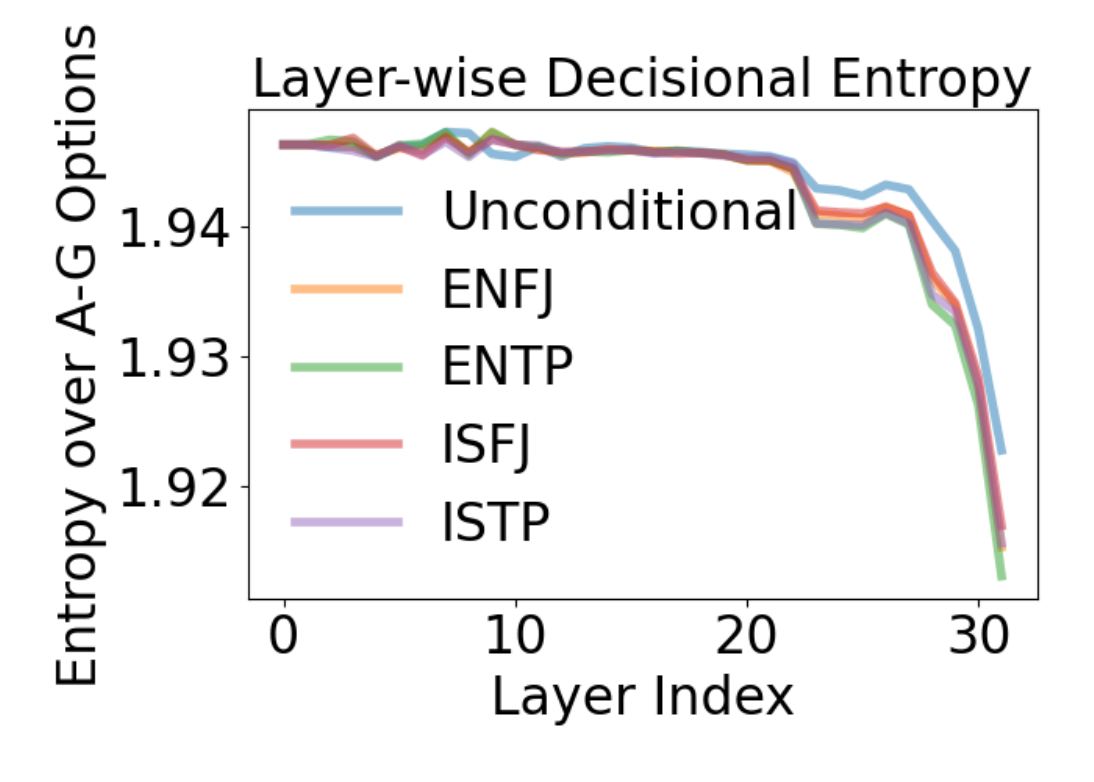}
\caption{GPTQ INT4}
\label{fig:Mistral-7B-Instruct-v0.3-conditional-entropy-breakdown:b}
\end{subfigure}
\hfill
\begin{subfigure}[t]{0.24\linewidth}
\centering
\includegraphics[width=\linewidth]{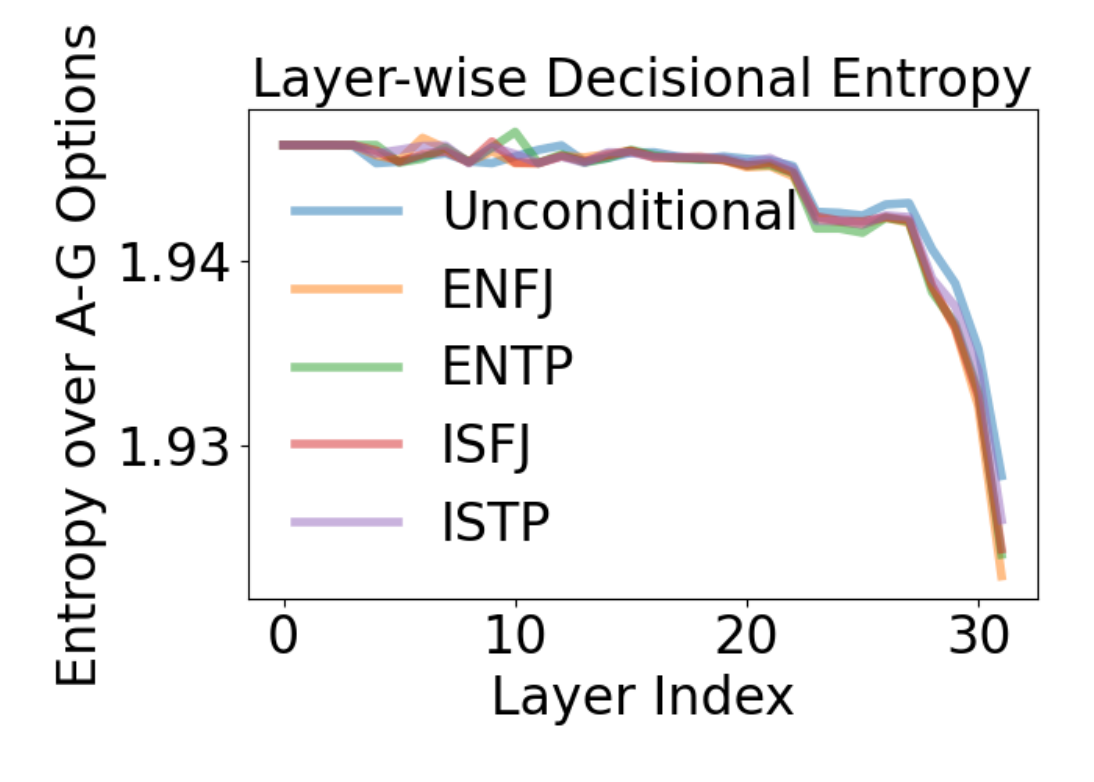}
\caption{AWQ INT4}
\label{fig:Mistral-7B-Instruct-v0.3-conditional-entropy-breakdown:c}
\end{subfigure}
\hfill
\begin{subfigure}[t]{0.24\linewidth}
\centering
\includegraphics[width=\linewidth]{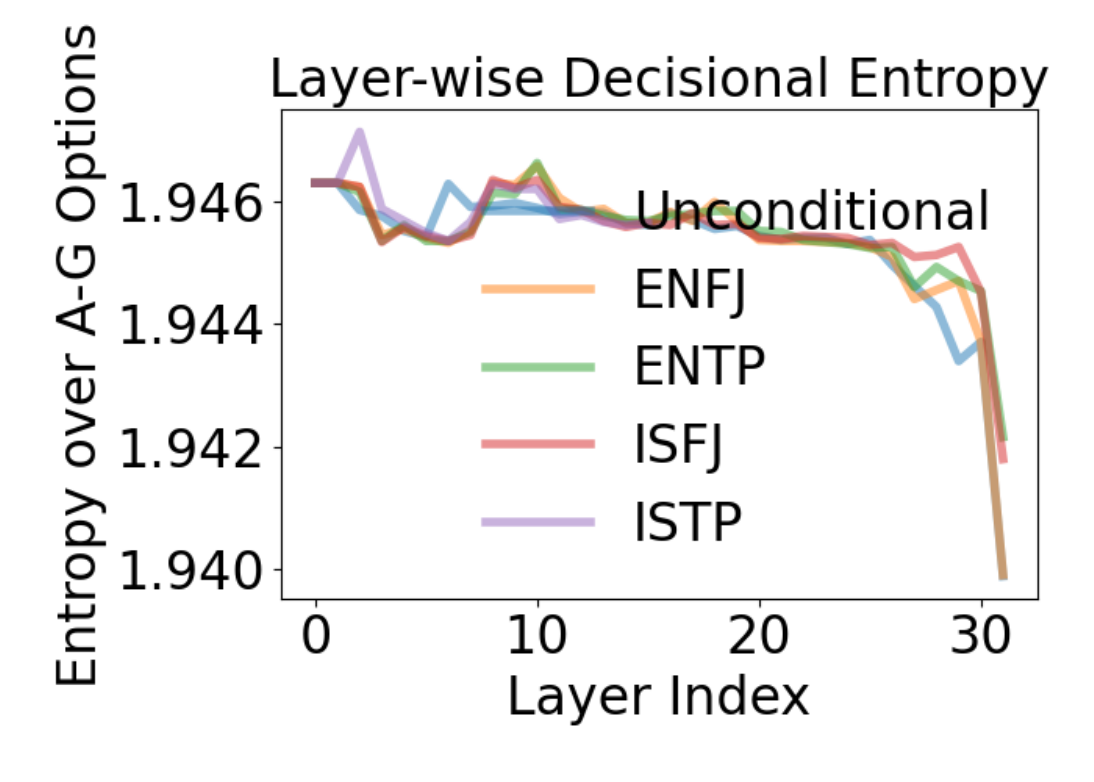}
\caption{AQLM 2-bit}
\label{fig:Mistral-7B-Instruct-v0.3-conditional-entropy-breakdown:d}
\end{subfigure}

\vspace{0.5em}

\begin{subfigure}[t]{0.24\linewidth}
\centering
\includegraphics[width=\linewidth]{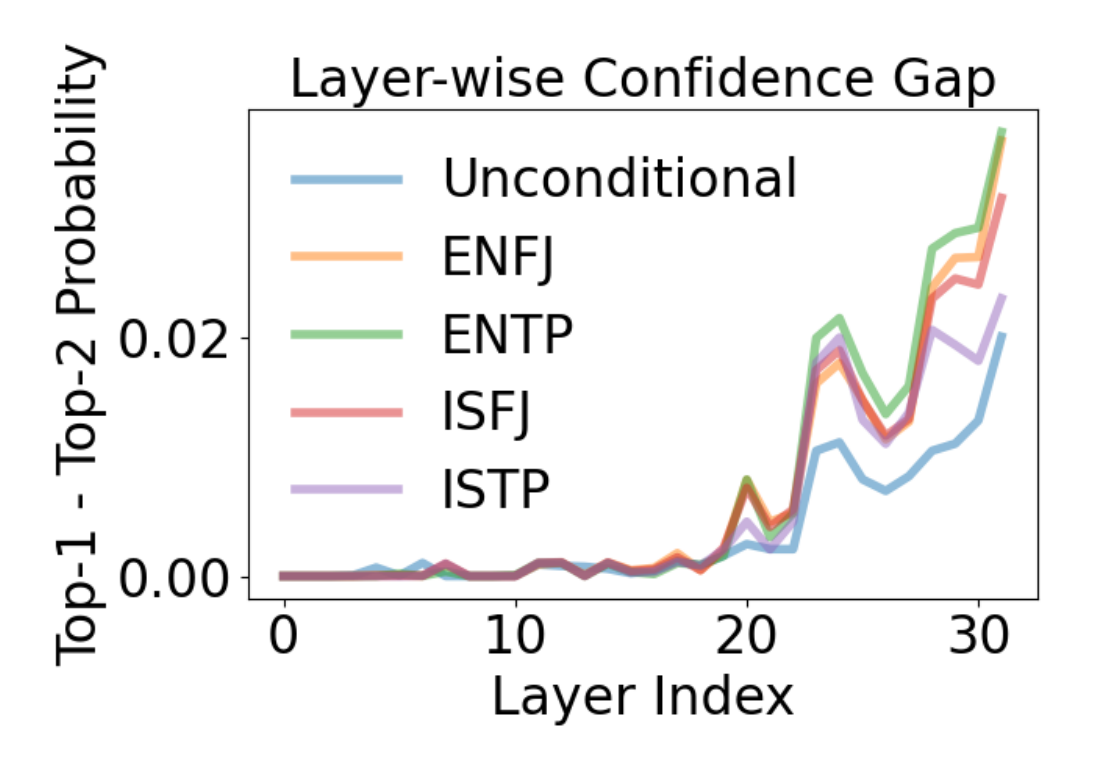}
\caption{FP16}
\label{fig:Mistral-7B-Instruct-v0.3-conditional-gap-breakdown:a}
\end{subfigure}
\hfill
\begin{subfigure}[t]{0.24\linewidth}
\centering
\includegraphics[width=\linewidth]{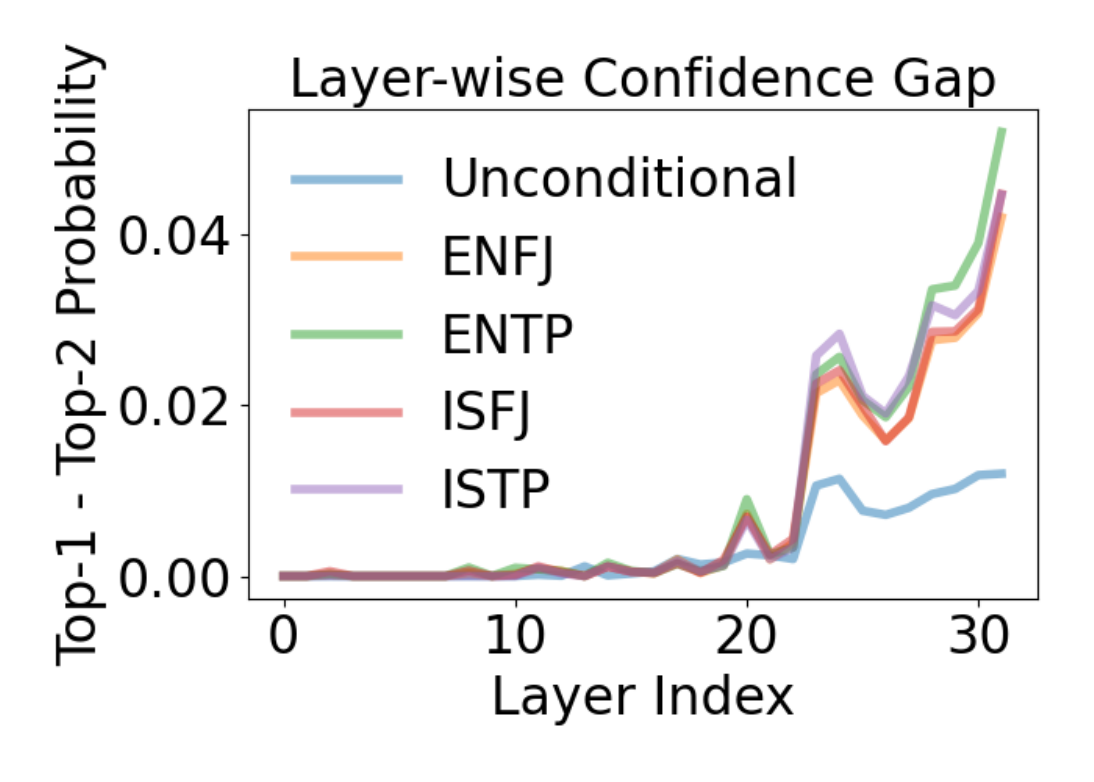}
\caption{GPTQ INT4}
\label{fig:Mistral-7B-Instruct-v0.3-conditional-gap-breakdown:b}
\end{subfigure}
\hfill
\begin{subfigure}[t]{0.24\linewidth}
\centering
\includegraphics[width=\linewidth]{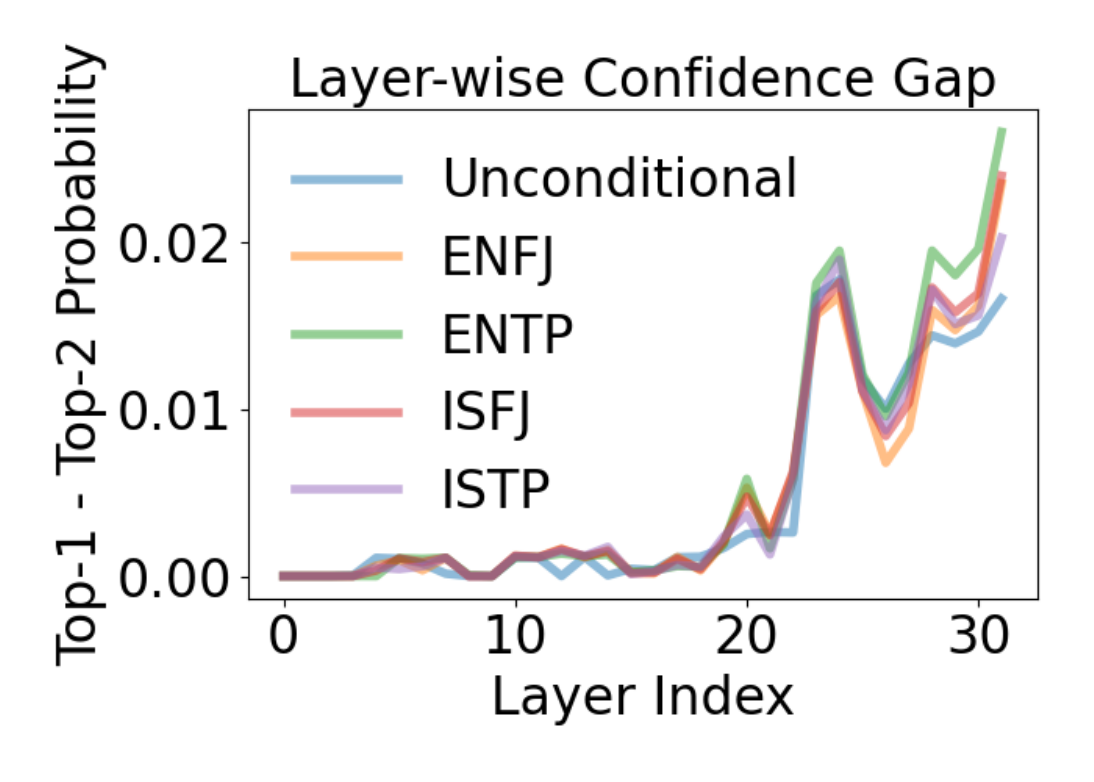}
\caption{AWQ INT4}
\label{fig:Mistral-7B-Instruct-v0.3-conditional-gap-breakdown:c}
\end{subfigure}
\hfill
\begin{subfigure}[t]{0.24\linewidth}
\centering
\includegraphics[width=\linewidth]{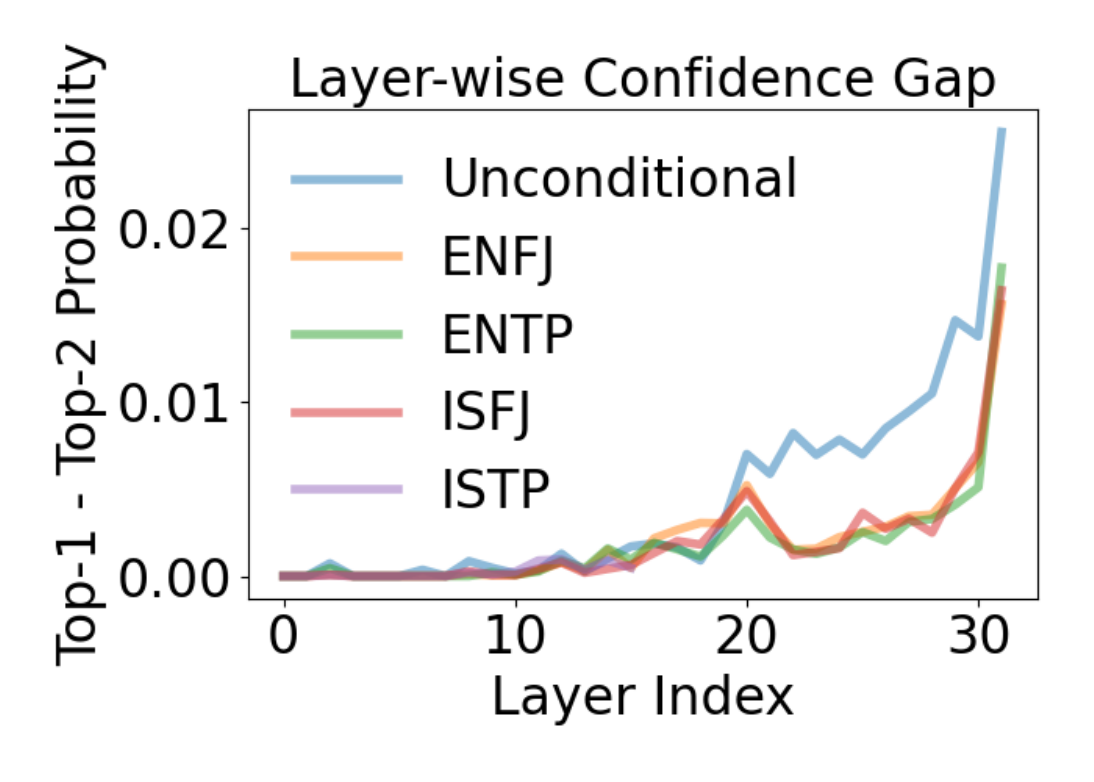}
\caption{AQLM 2-bit}
\label{fig:Mistral-7B-Instruct-v0.3-conditional-gap-breakdown:d}
\end{subfigure}

\caption{Mistral-7B-Instruct-v0.3: layer-wise decisional entropy (top row) and top-1 vs. top-2 probability gap (bottom row) across quantization variants. Early layers show substantial uncertainty, whereas later layers progressively become more decisive.}
\label{fig:Mistral-7B-Instruct-v0.3-conditional-combined}
\end{figure*}

\begin{figure*}[t]
\centering

\begin{subfigure}[t]{0.24\linewidth}
\centering
\includegraphics[width=\linewidth]{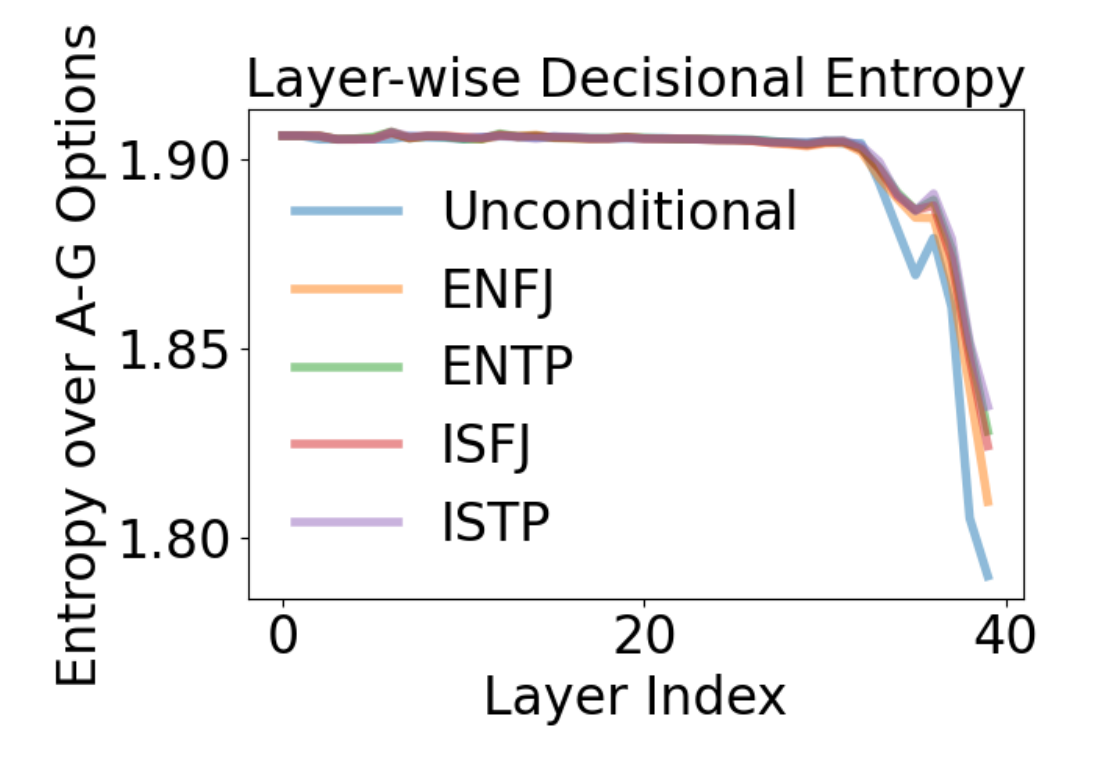}
\caption{FP16}
\label{fig:Mistral-Small-24B-Instruct-2501-conditional-entropy-breakdown:a}
\end{subfigure}
\hfill
\begin{subfigure}[t]{0.24\linewidth}
\centering
\includegraphics[width=\linewidth]{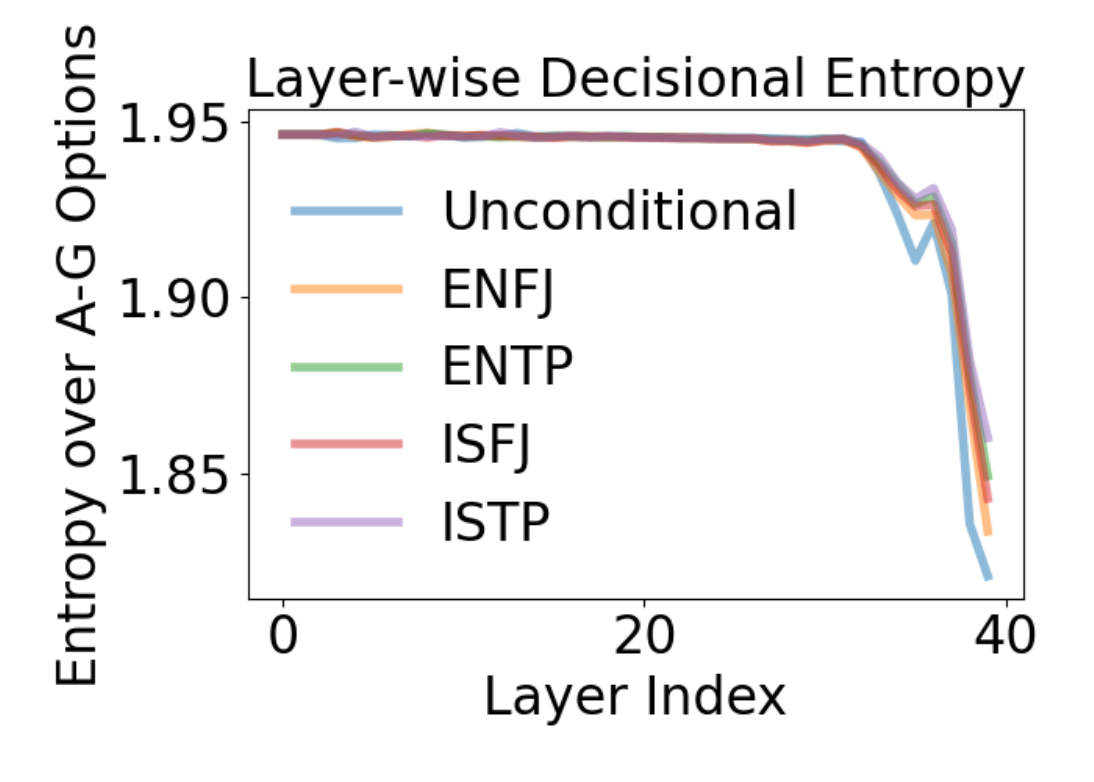}
\caption{GPTQ INT4}
\label{fig:Mistral-Small-24B-Instruct-2501-entropy-breakdown:b}
\end{subfigure}
\hfill
\begin{subfigure}[t]{0.24\linewidth}
\centering
\includegraphics[width=\linewidth]{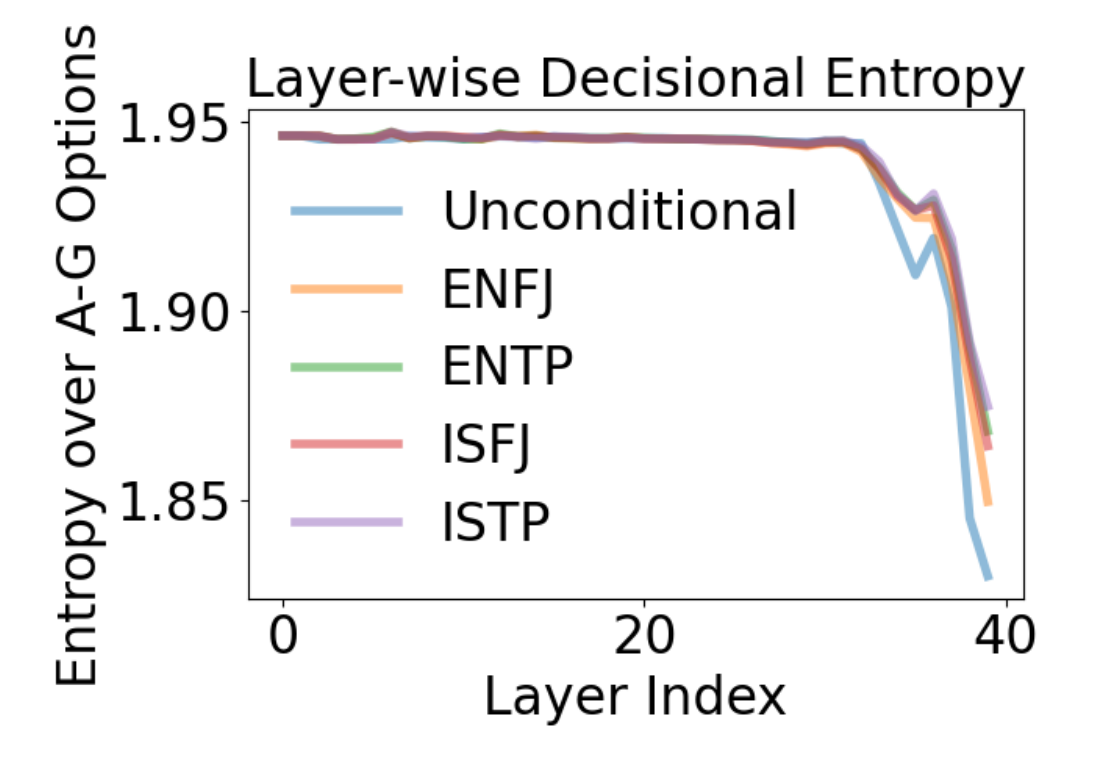}
\caption{AWQ INT4}
\label{fig:Mistral-Small-24B-Instruct-2501-conditional-entropy-breakdown:c}
\end{subfigure}
\hfill
\begin{subfigure}[t]{0.24\linewidth}
\centering
\includegraphics[width=\linewidth]{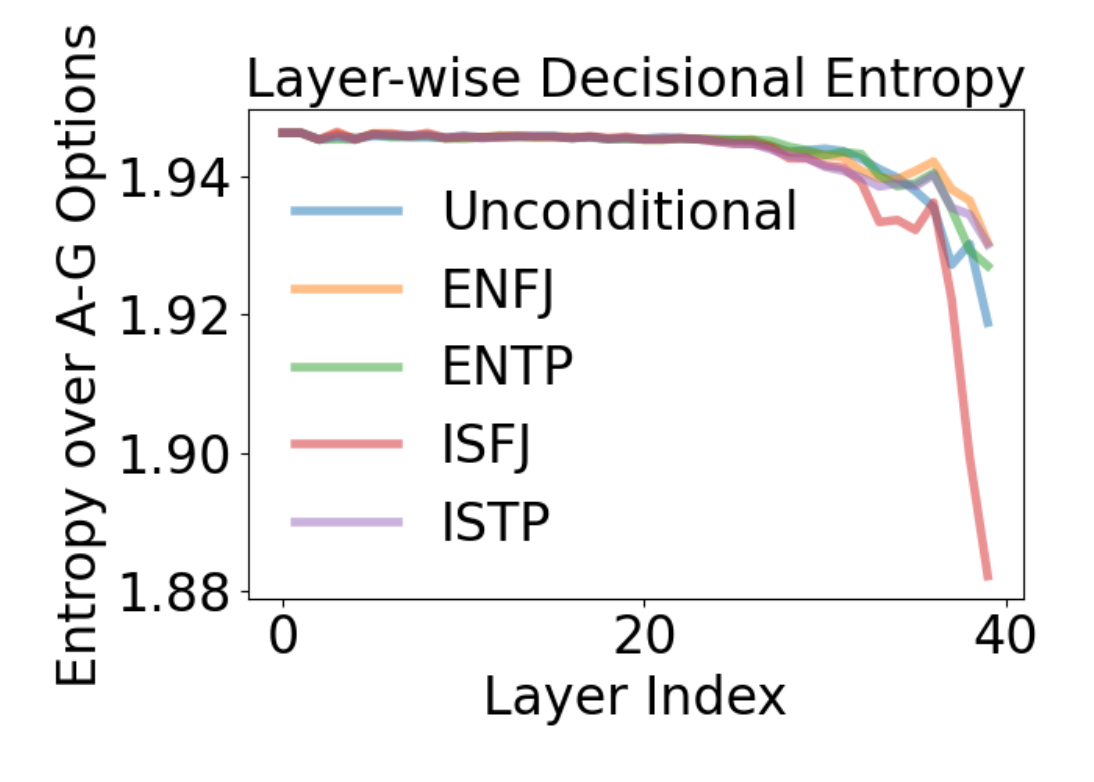}
\caption{AQLM 2-bit}
\label{fig:Mistral-Small-24B-Instruct-2501-conditional-entropy-breakdown:d}
\end{subfigure}

\vspace{0.5em}

\begin{subfigure}[t]{0.24\linewidth}
\centering
\includegraphics[width=\linewidth]{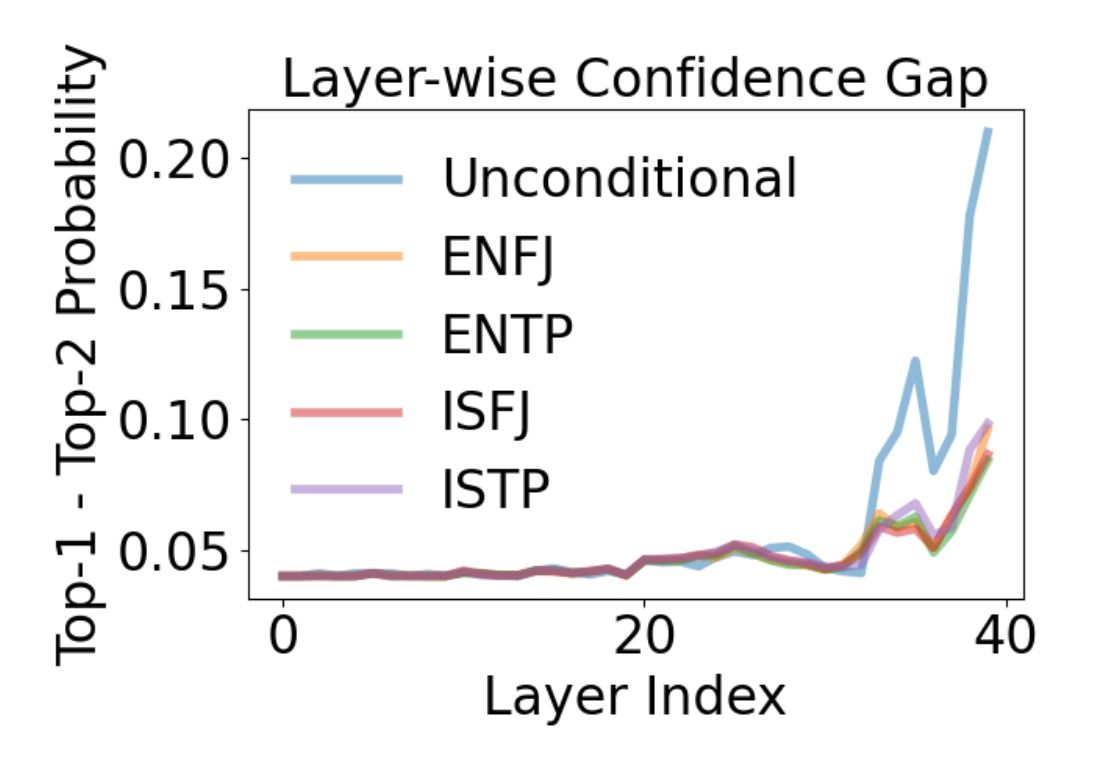}
\caption{FP16}
\label{fig:Mistral-Small-24B-Instruct-2501-conditional-gap-breakdown:a}
\end{subfigure}
\hfill
\begin{subfigure}[t]{0.24\linewidth}
\centering
\includegraphics[width=\linewidth]{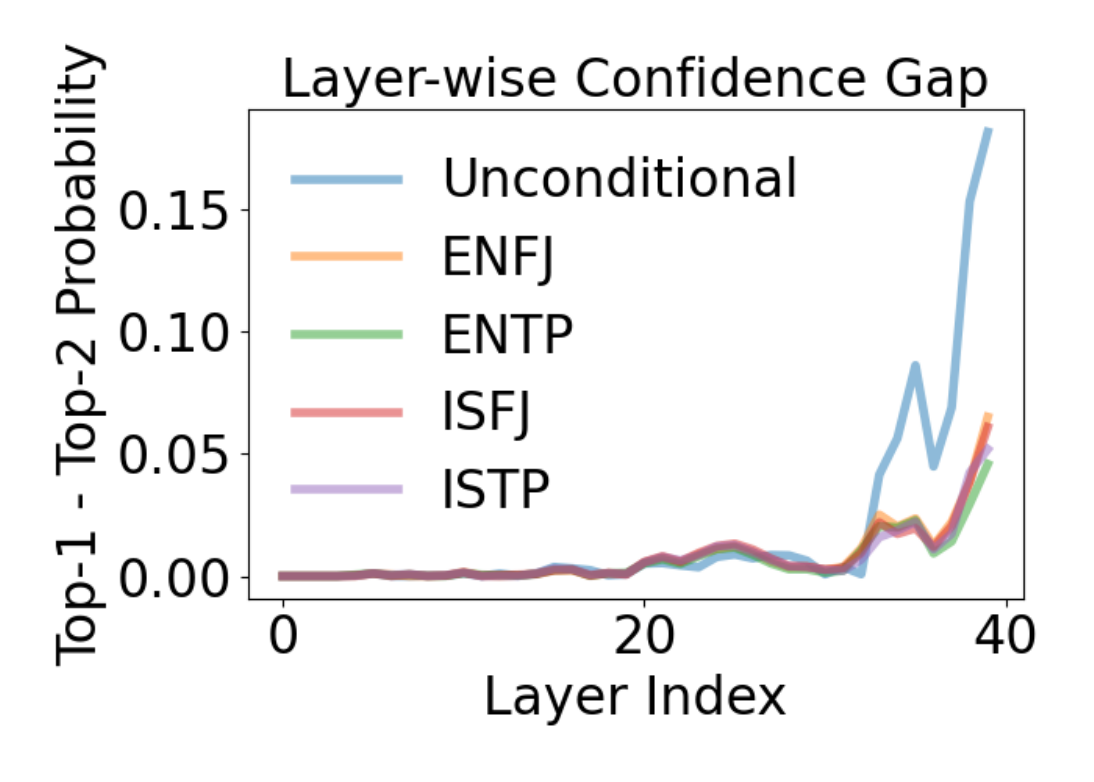}
\caption{GPTQ INT4}
\label{fig:Mistral-Small-24B-Instruct-2501-conditional-gap-breakdown:b}
\end{subfigure}
\hfill
\begin{subfigure}[t]{0.24\linewidth}
\centering
\includegraphics[width=\linewidth]{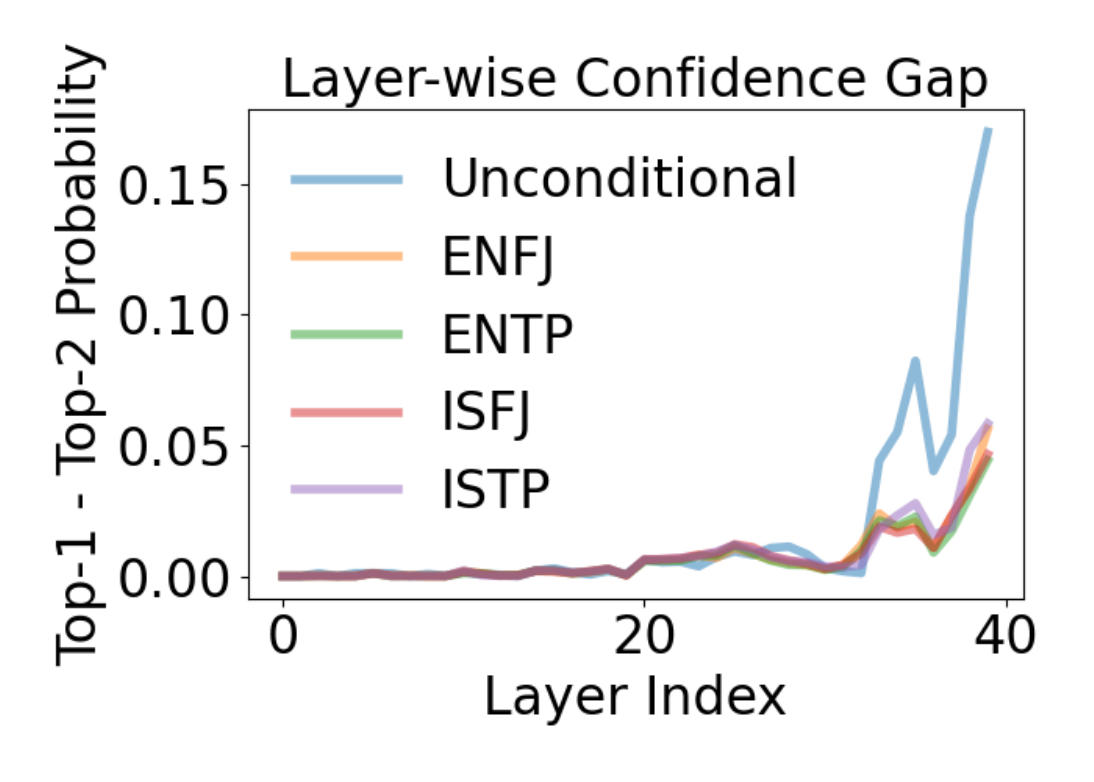}
\caption{AWQ INT4}
\label{fig:Mistral-Small-24B-Instruct-2501-conditional-gap-breakdown:c}
\end{subfigure}
\hfill
\begin{subfigure}[t]{0.24\linewidth}
\centering
\includegraphics[width=\linewidth]{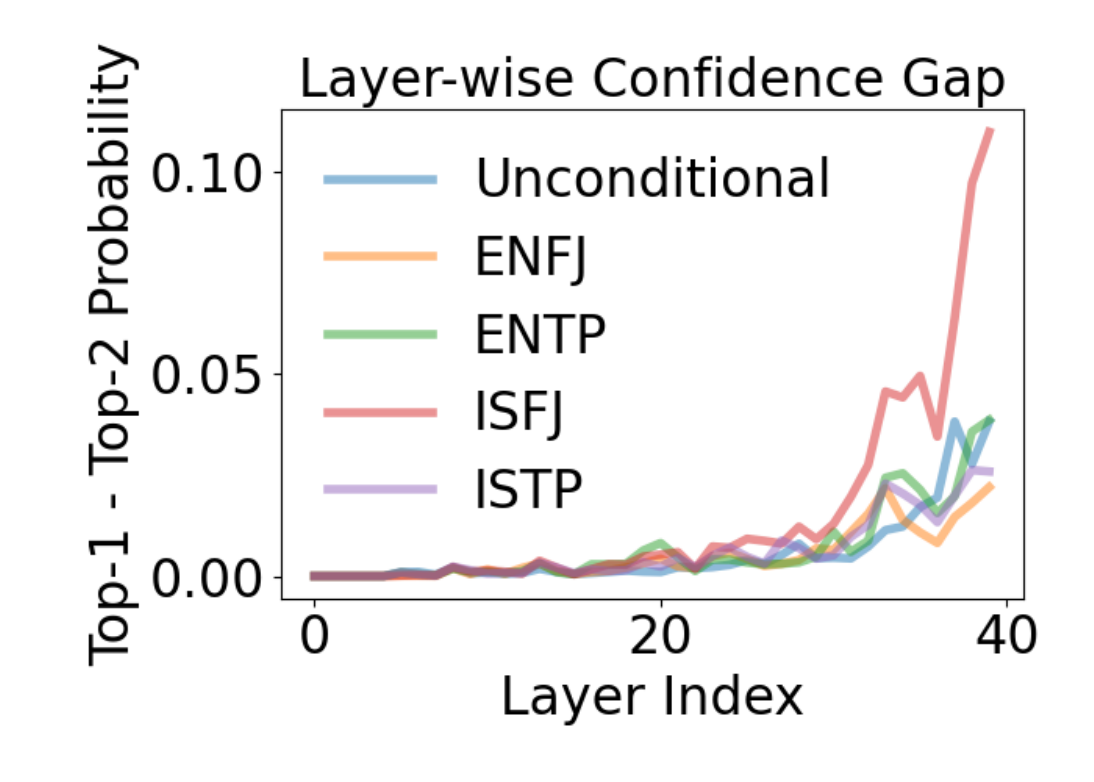}
\caption{AQLM 2-bit}
\label{fig:Mistral-Small-24B-Instruct-2501-conditional-gap-breakdown:d}
\end{subfigure}

\caption{
Mistral-Small-24B-Instruct-2501: layer-wise decisional entropy (top row) and top-1 vs. top-2 probability gap (bottom row) across quantization variants. Deeper layers generally exhibit lower entropy and larger confidence margins, while stronger quantization can disrupt this late-layer consolidation and reduce decisional sharpness.
}
\label{fig:Mistral-Small-24B-Instruct-2501-conditional-combined}
\end{figure*}

\begin{figure*}[t]
\centering

\begin{subfigure}[t]{0.24\linewidth}
\centering
\includegraphics[width=\linewidth]{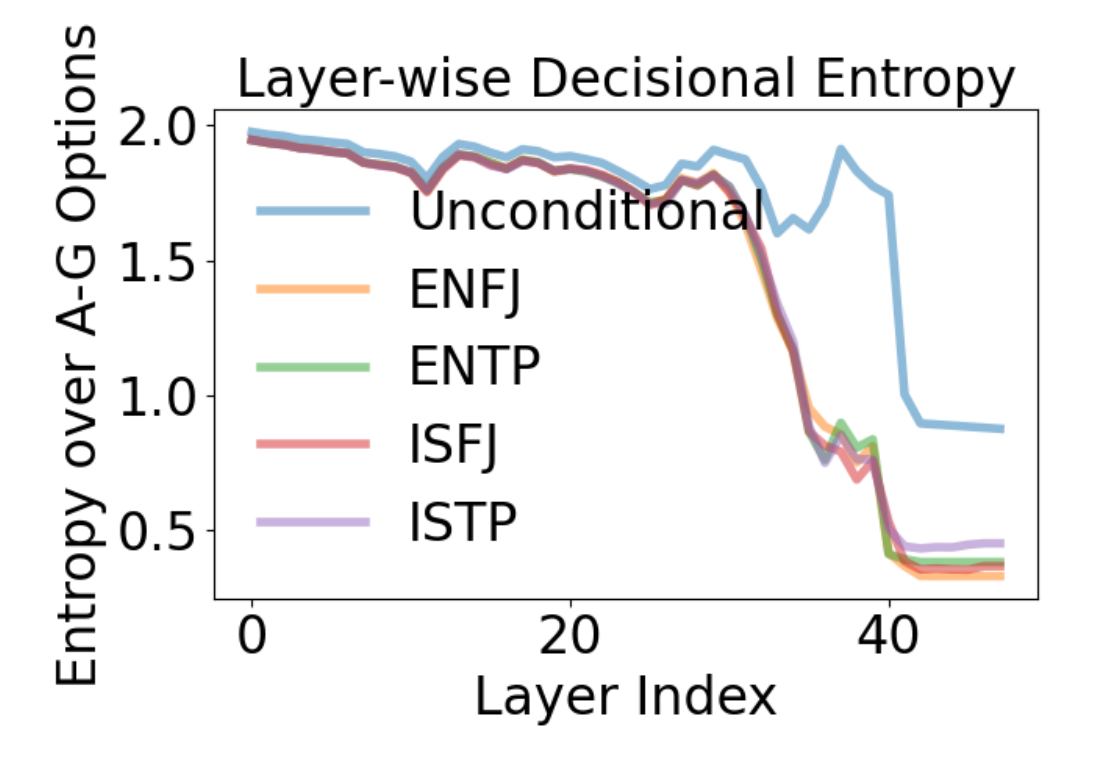}
\caption{FP16}
\label{fig:Qwen2.5-14B-Instruct-conditional-entropy-breakdown:a}
\end{subfigure}
\hfill
\begin{subfigure}[t]{0.24\linewidth}
\centering
\includegraphics[width=\linewidth]{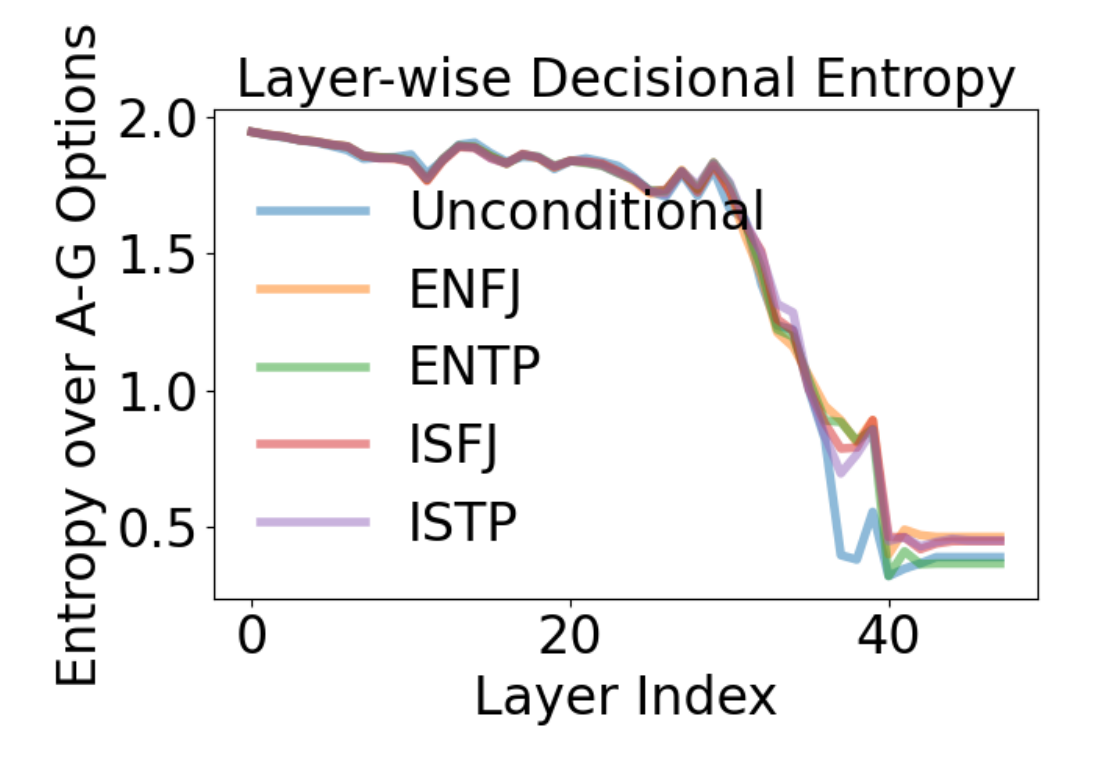}
\caption{GPTQ INT4}
\label{fig:Qwen2.5-14B-Instruct-conditional-entropy-breakdown:b}
\end{subfigure}
\hfill
\begin{subfigure}[t]{0.24\linewidth}
\centering
\includegraphics[width=\linewidth]{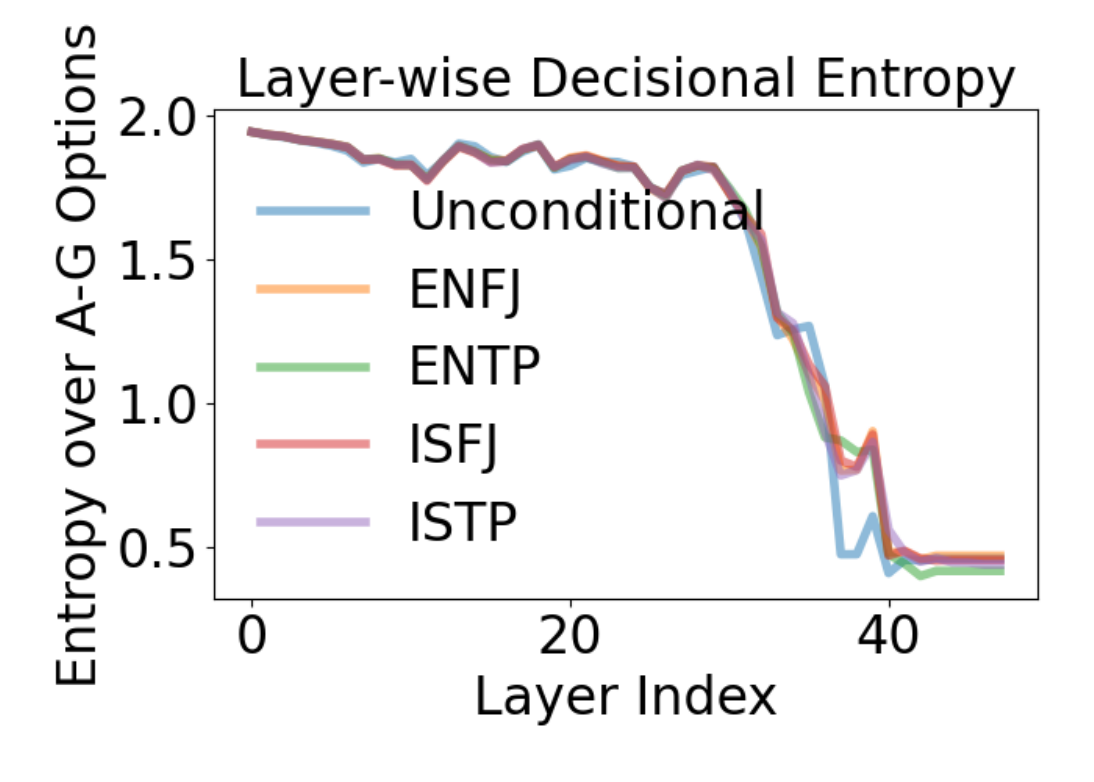}
\caption{AWQ INT4}
\label{fig:Qwen2.5-14B-Instruct-conditional-entropy-breakdown:c}
\end{subfigure}
\hfill
\begin{subfigure}[t]{0.24\linewidth}
\centering
\includegraphics[width=\linewidth]{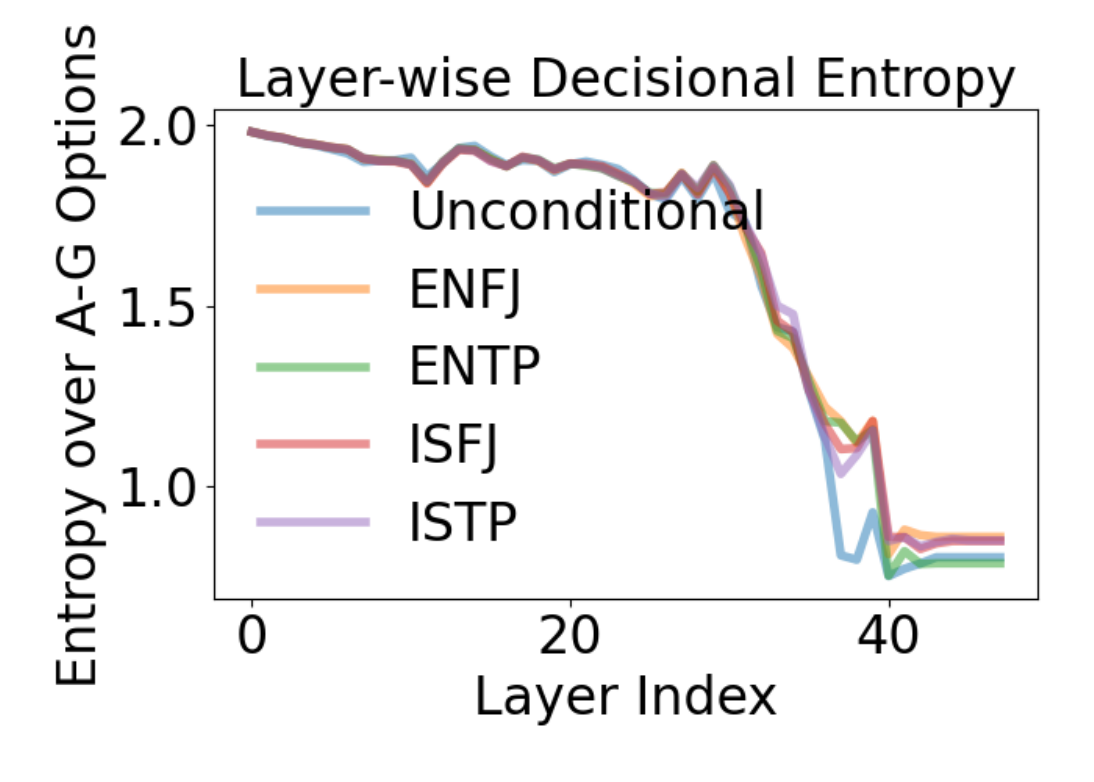}
\caption{AQLM 2-bit}
\label{fig:Qwen2.5-14B-Instruct-conditional-entropy-breakdown:d}
\end{subfigure}

\vspace{0.5em}

\begin{subfigure}[t]{0.24\linewidth}
\centering
\includegraphics[width=\linewidth]{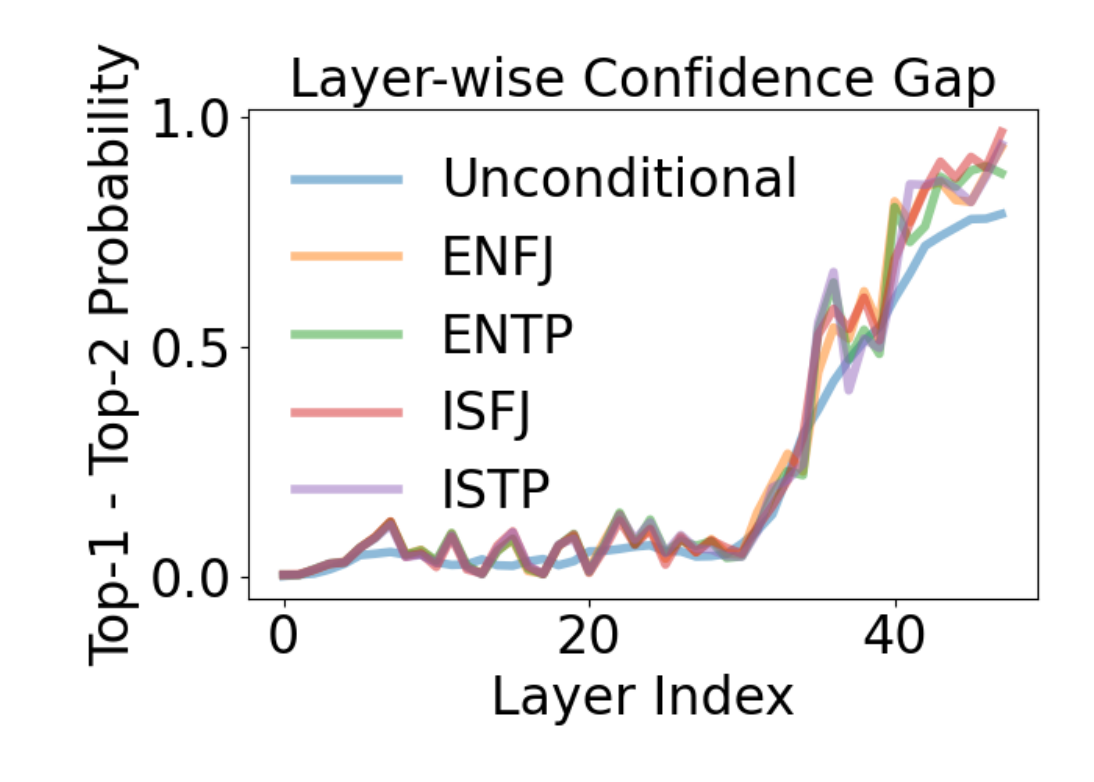}
\caption{FP16}
\label{fig:Qwen2.5-14B-Instruct-conditional-gap-breakdown:a}
\end{subfigure}
\hfill
\begin{subfigure}[t]{0.24\linewidth}
\centering
\includegraphics[width=\linewidth]{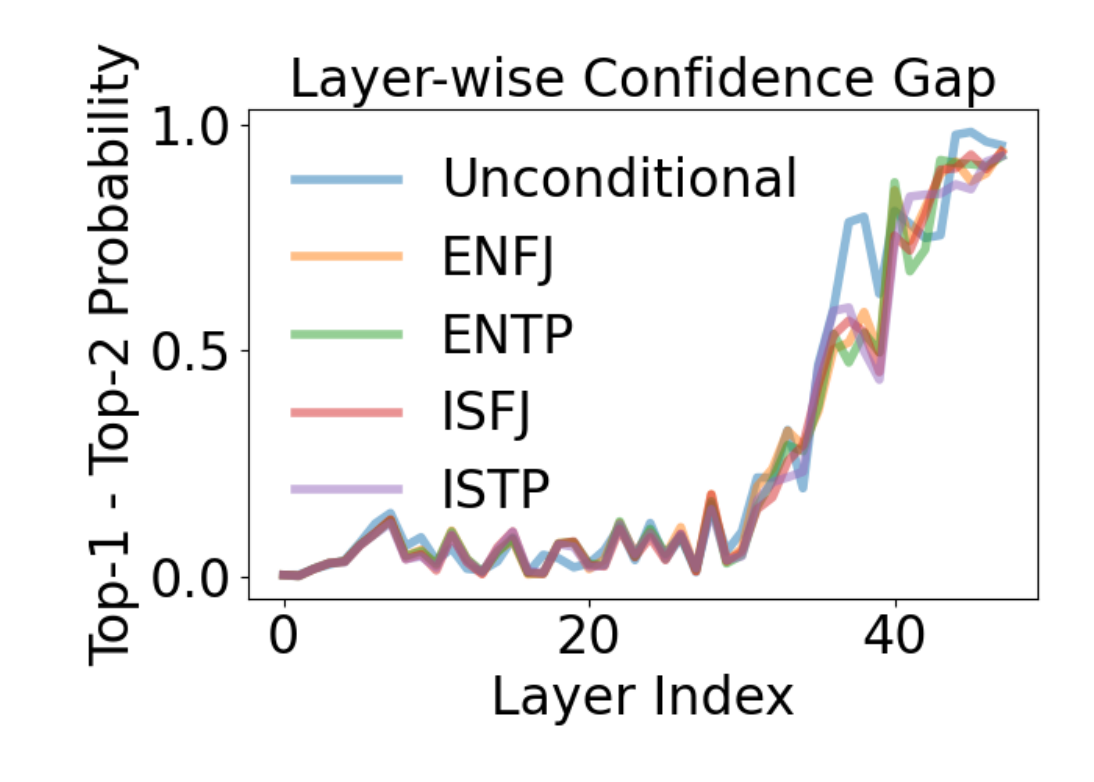}
\caption{GPTQ INT4}
\label{fig:Qwen2.5-14B-Instruct-conditional-gap-breakdown:b}
\end{subfigure}
\hfill
\begin{subfigure}[t]{0.24\linewidth}
\centering
\includegraphics[width=\linewidth]{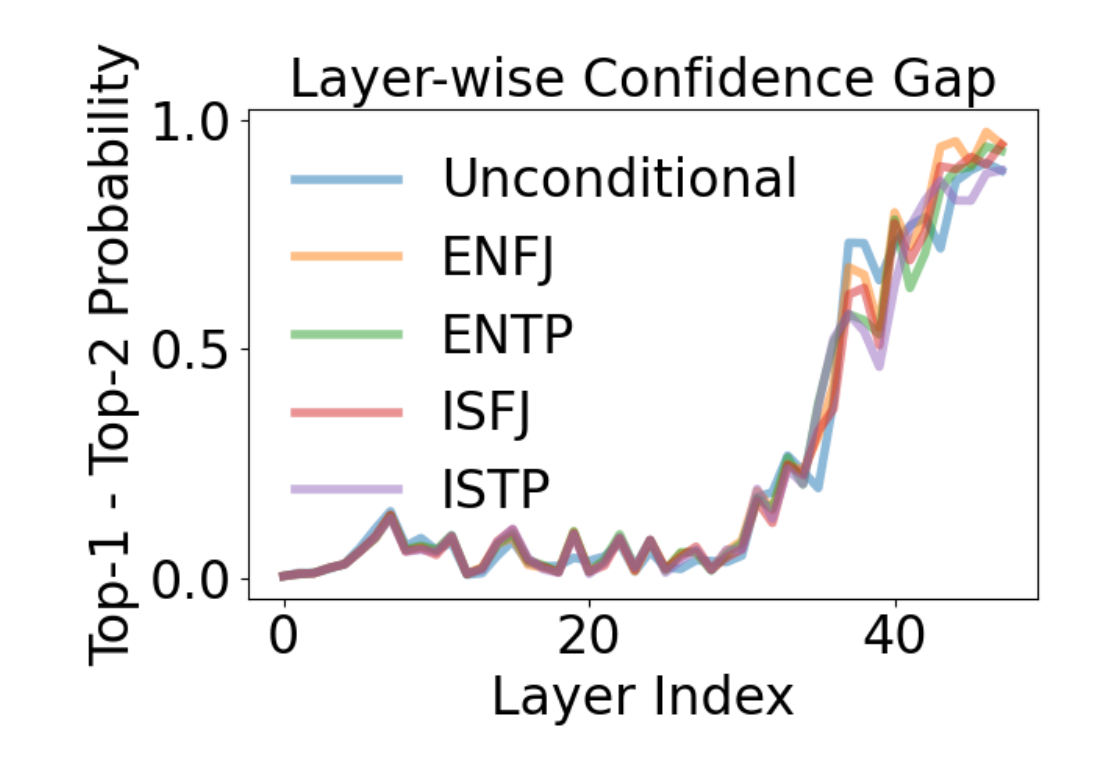}
\caption{AWQ INT4}
\label{fig:Qwen2.5-14B-Instruct-conditional-gap-breakdown:c}
\end{subfigure}
\hfill
\begin{subfigure}[t]{0.24\linewidth}
\centering
\includegraphics[width=\linewidth]{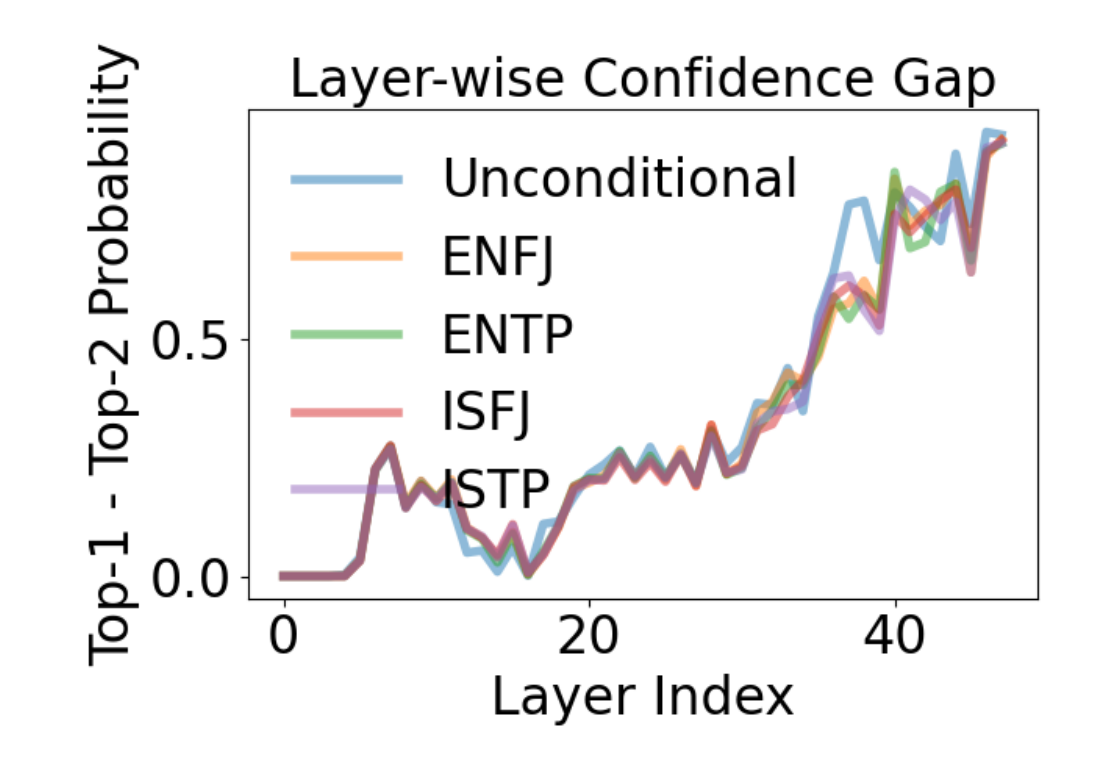}
\caption{AQLM 2-bit}
\label{fig:Qwen2.5-14B-Instruct-conditional-gap-breakdown:d}
\end{subfigure}

\caption{
Qwen2.5-14B-Instruct: layer-wise decisional entropy (top row) and top-1 vs. top-2 probability gap (bottom row) across quantization variants. Early layers remain uncertain, whereas deeper layers increasingly sharpen response preferences. Extreme quantization disrupts this progression, leading to higher entropy and weaker confidence margins near the output.
}
\label{fig:Qwen2.5-14B-Instruct-conditional-combined}
\end{figure*}

\begin{figure*}[t]
\centering

\begin{subfigure}[t]{0.24\linewidth}
\centering
\includegraphics[width=\linewidth]{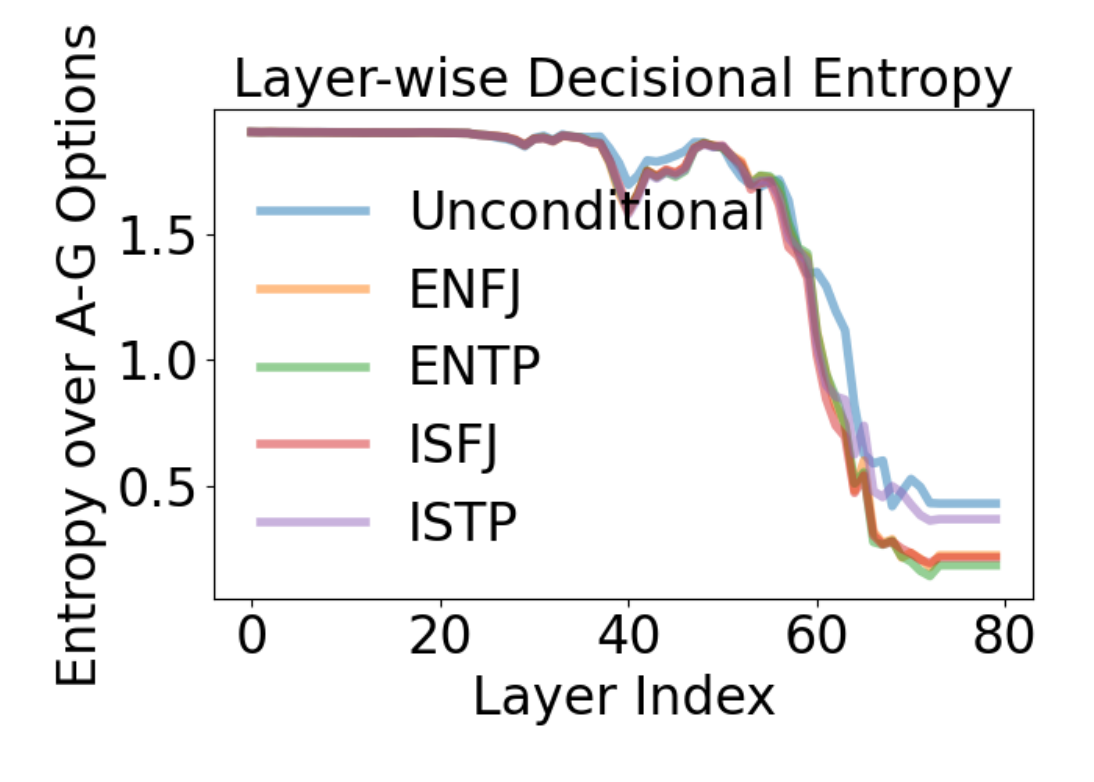}
\caption{FP16}
\label{fig:Qwen2.5-72B-Instruct-conditional-entropy-breakdown:a}
\end{subfigure}
\hfill
\begin{subfigure}[t]{0.24\linewidth}
\centering
\includegraphics[width=\linewidth]{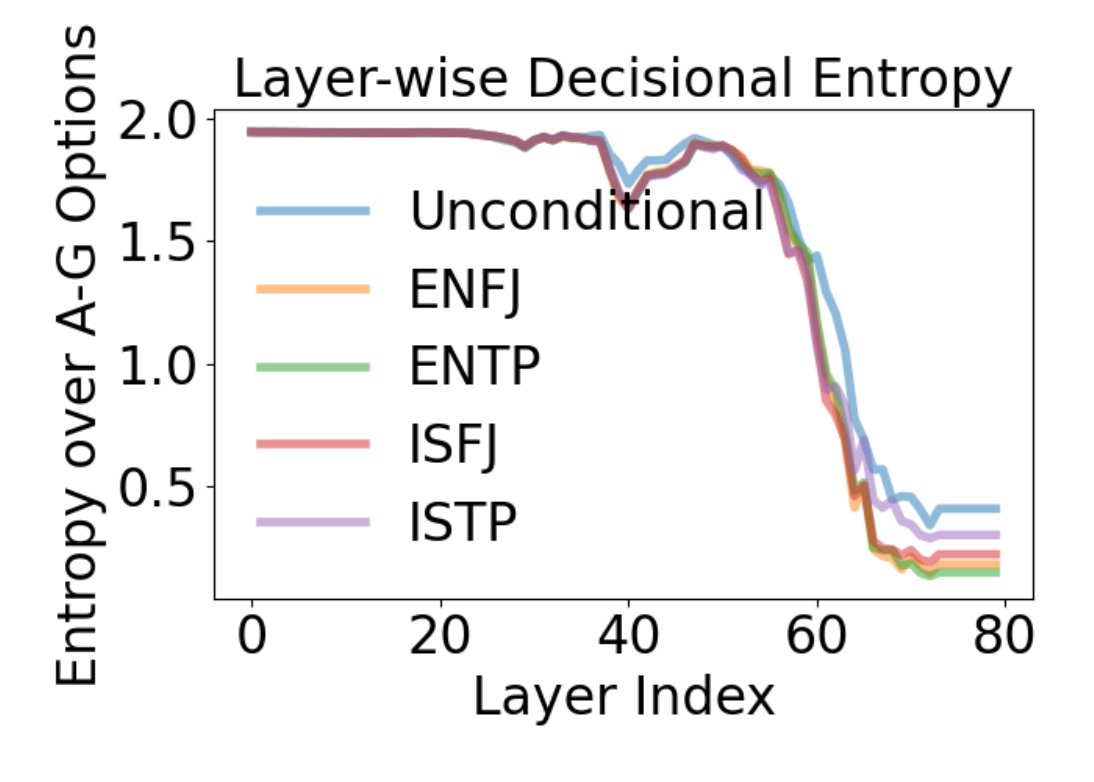}
\caption{GPTQ INT4}
\label{fig:Qwen2.5-72B-Instruct-conditional-entropy-breakdown:b}
\end{subfigure}
\hfill
\begin{subfigure}[t]{0.24\linewidth}
\centering
\includegraphics[width=\linewidth]{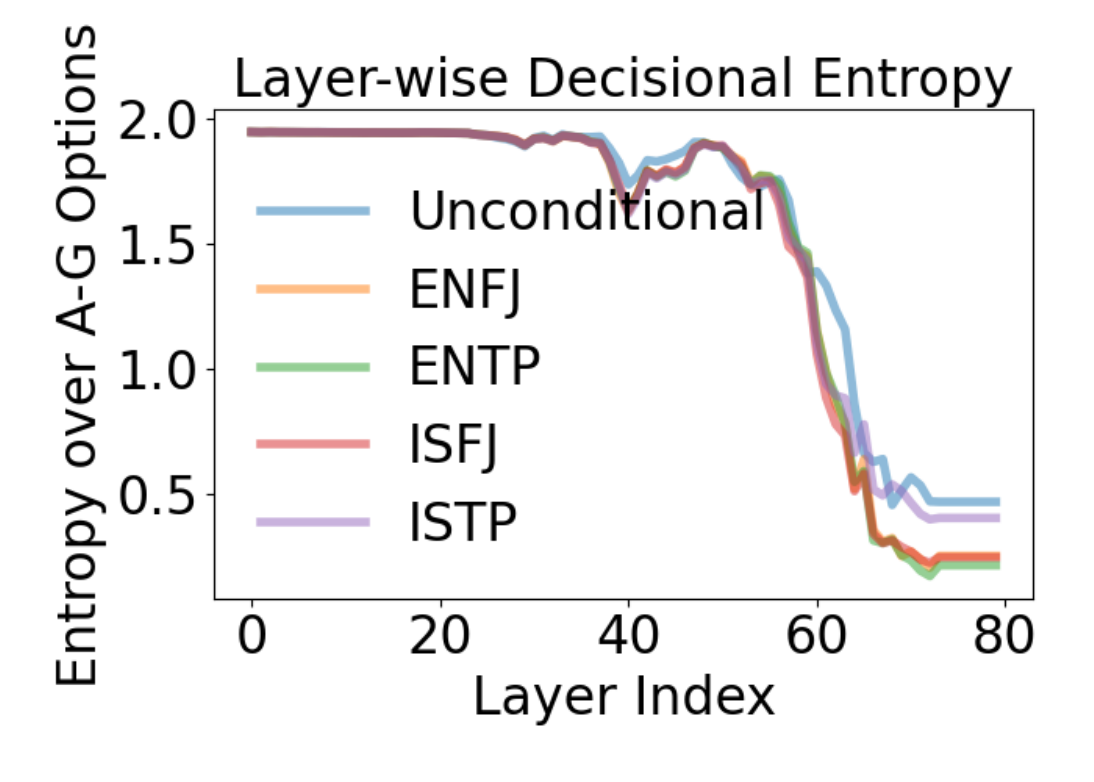}
\caption{AWQ INT4}
\label{fig:Qwen2.5-72B-Instruct-conditional-entropy-breakdown:c}
\end{subfigure}
\hfill
\begin{subfigure}[t]{0.24\linewidth}
\centering
\includegraphics[width=\linewidth]{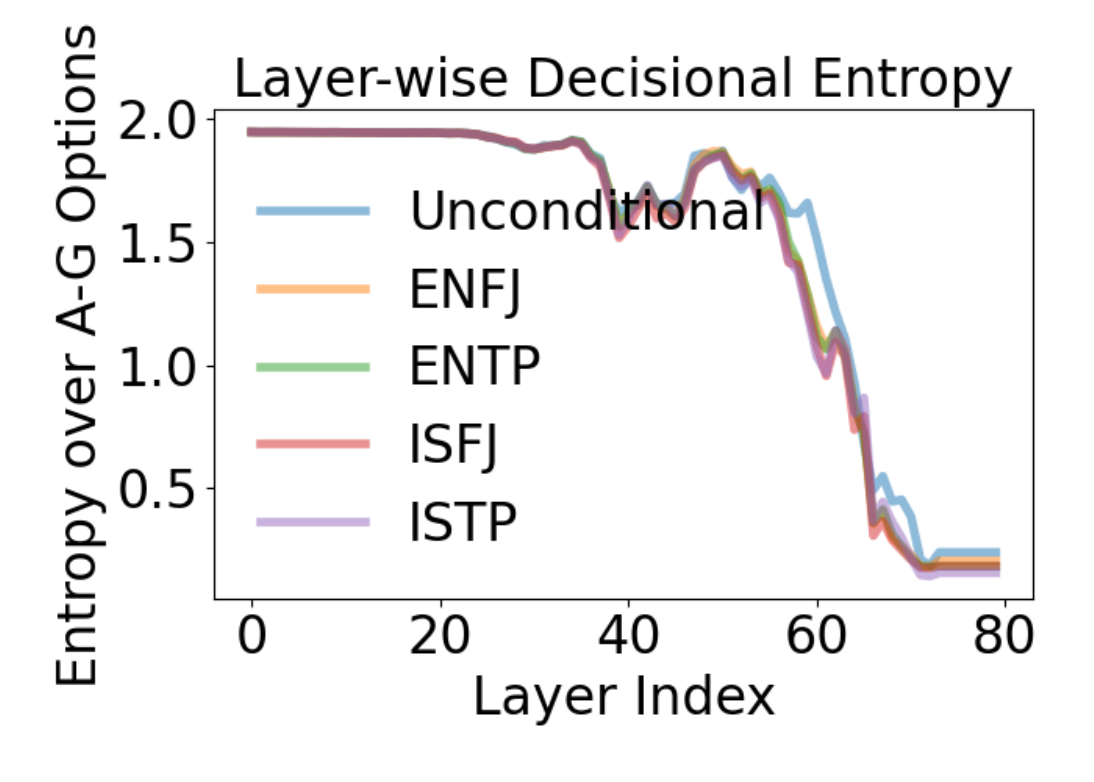}
\caption{AQLM 2-bit}
\label{fig:Qwen2.5-72B-Instruct-conditional-entropy-breakdown:d}
\end{subfigure}

\vspace{0.5em}

\begin{subfigure}[t]{0.24\linewidth}
\centering
\includegraphics[width=\linewidth]{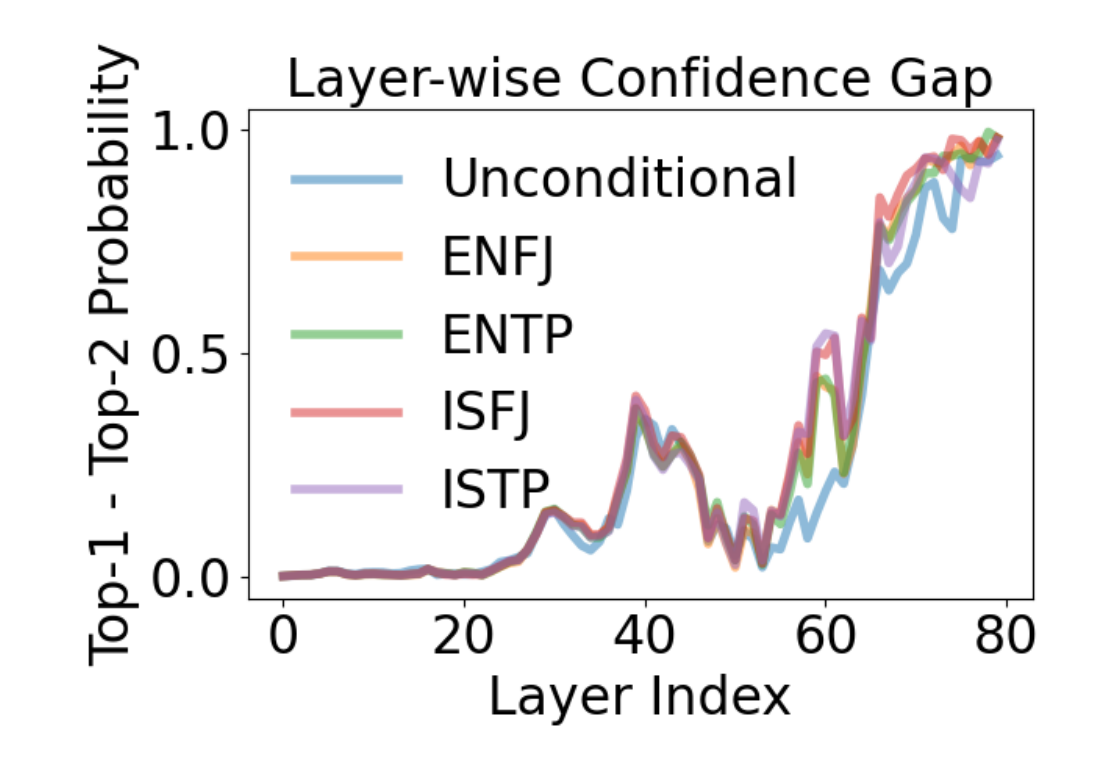}
\caption{FP16}
\label{fig:Qwen2.5-72B-Instruct-conditional-gap-breakdown:a}
\end{subfigure}
\hfill
\begin{subfigure}[t]{0.24\linewidth}
\centering
\includegraphics[width=\linewidth]{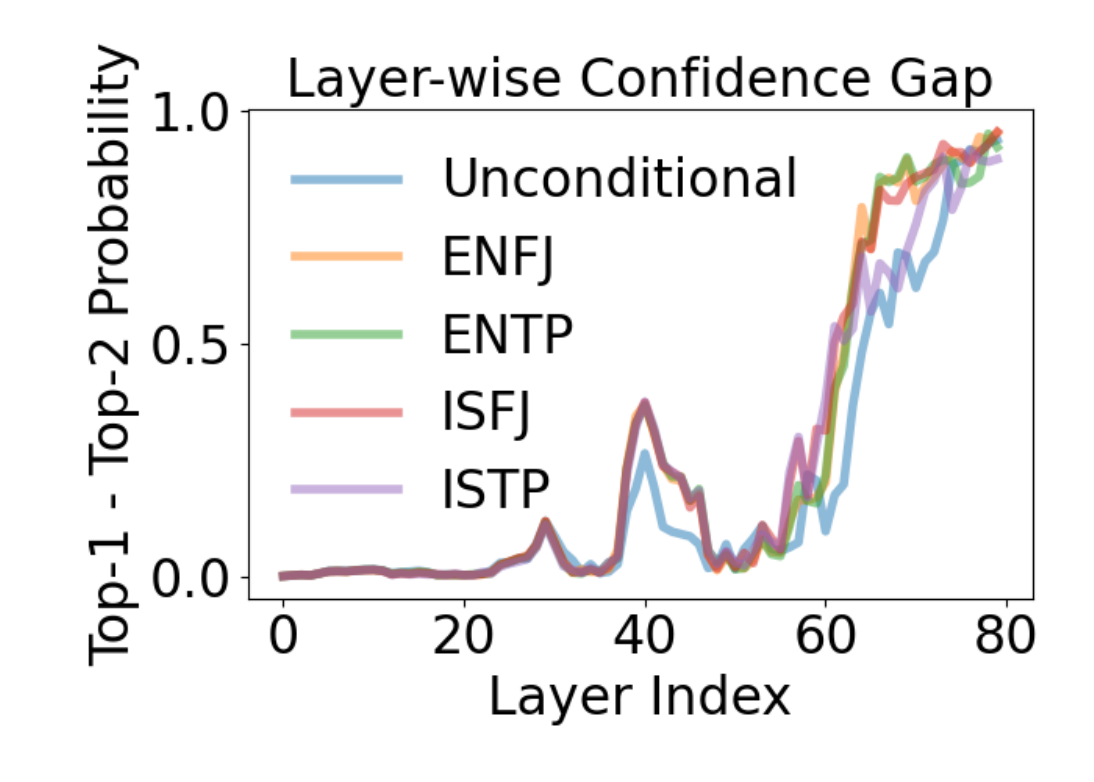}
\caption{GPTQ INT4}
\label{fig:Qwen2.5-72B-Instruct-conditional-gap-breakdown:b}
\end{subfigure}
\hfill
\begin{subfigure}[t]{0.24\linewidth}
\centering
\includegraphics[width=\linewidth]{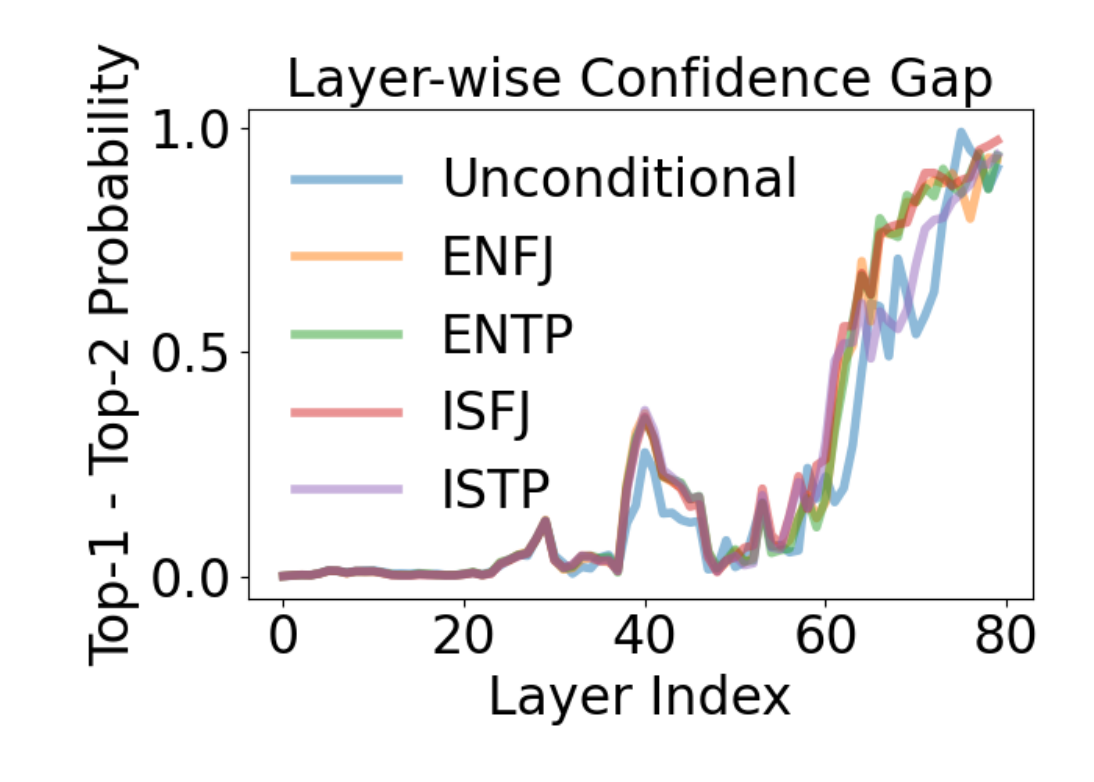}
\caption{AWQ INT4}
\label{fig:Qwen2.5-72B-Instruct-conditional-gap-breakdown:c}
\end{subfigure}
\hfill
\begin{subfigure}[t]{0.24\linewidth}
\centering
\includegraphics[width=\linewidth]{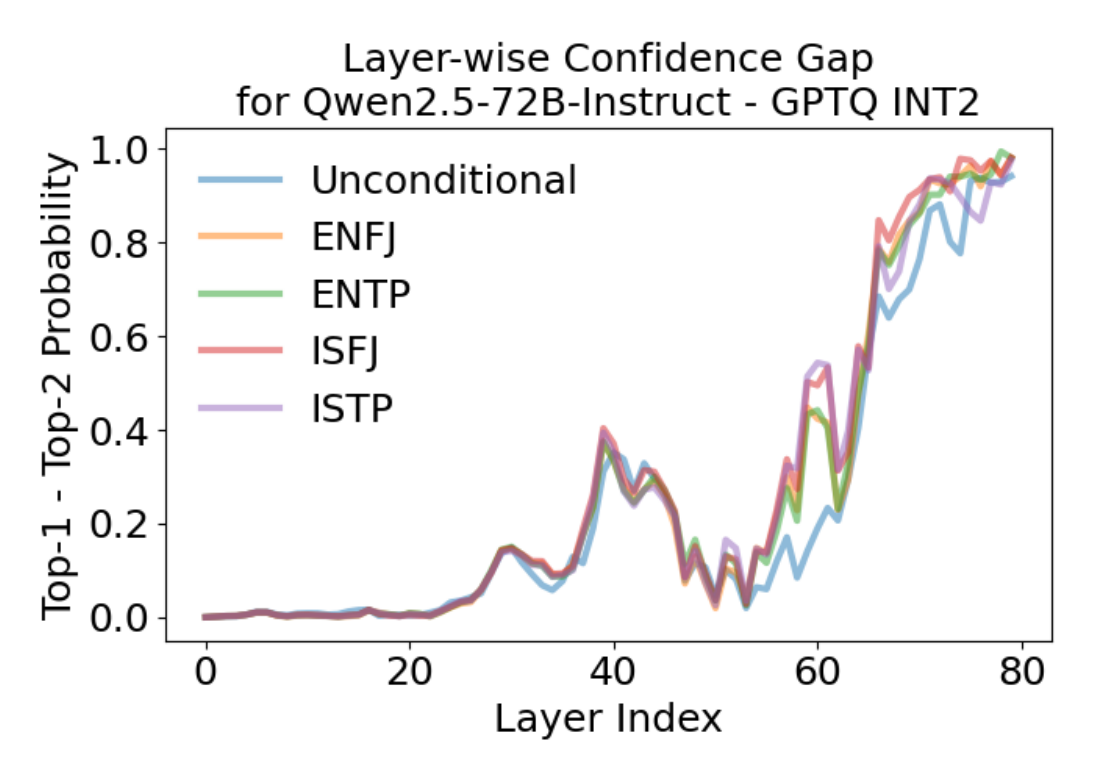}
\caption{AQLM 2-bit}
\label{fig:Qwen2.5-72B-Instruct-conditional-gap-breakdown:d}
\end{subfigure}

\caption{
Qwen2.5-72B-Instruct: layer-wise decisional entropy (top row) and top-1 vs. top-2 probability gap (bottom row) across quantization variants. Deeper layers generally show reduced uncertainty and stronger commitment, while stronger quantization can attenuate decisional sharpening and destabilize confidence margins near the output.
}
\label{fig:Qwen2.5-72B-Instruct-conditional-combined}
\end{figure*}

\begin{figure}[t]
\centering
\includegraphics[width=0.95\linewidth]{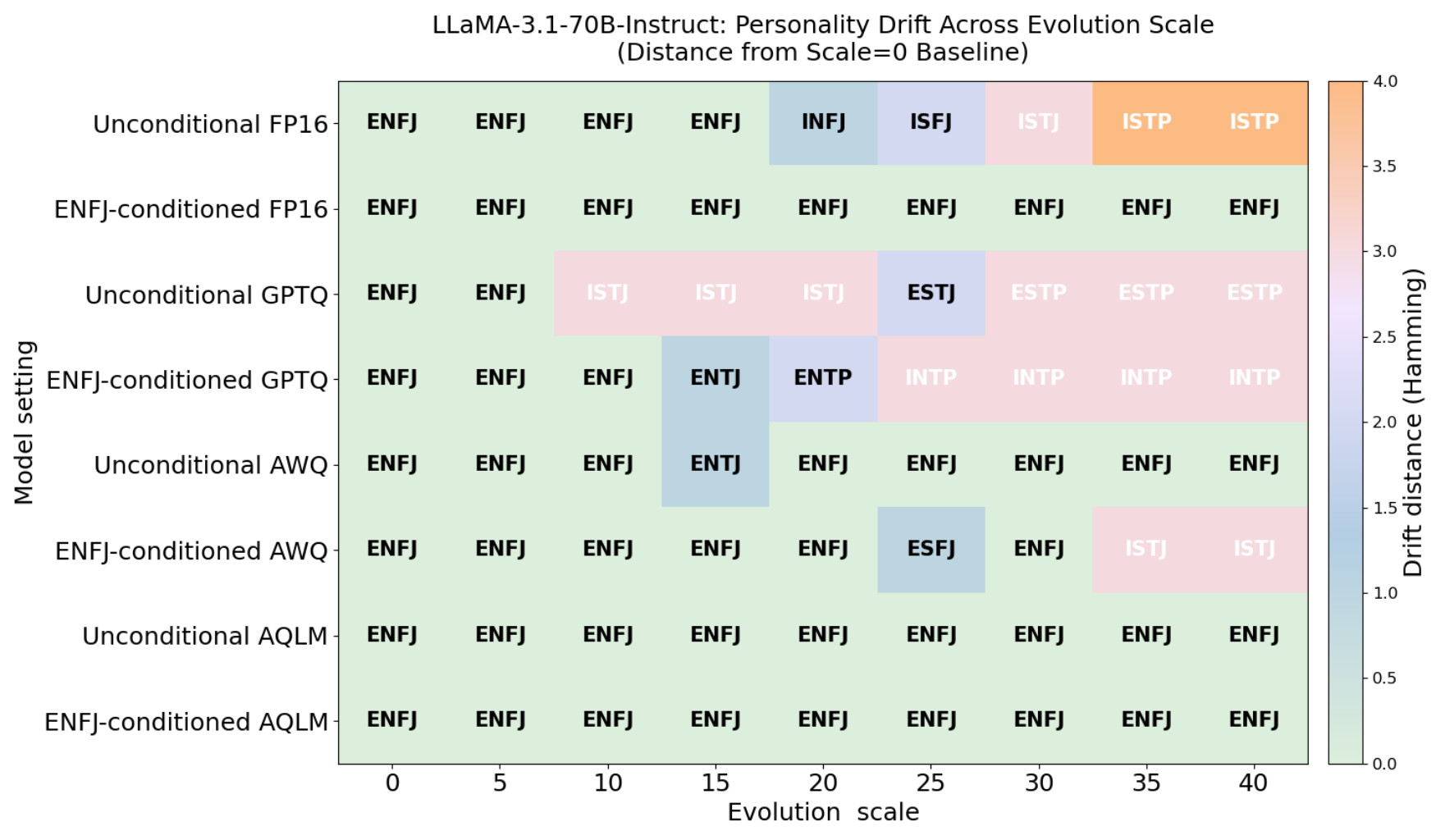}
\caption{Personality drift under increasing evolution scale. Each row corresponds to a model--prompt setting, and each column corresponds to an evolution scale. Cell color indicates Hamming distance between the predicted MBTI type at a given scale and the baseline prediction at scale $0$. 
Annotated labels show the inferred MBTI type.}
\label{fig:llama3.1_70b_instruct_dola_drift}
\end{figure}

\begin{figure}[t]
\centering
\includegraphics[width=0.95\linewidth]{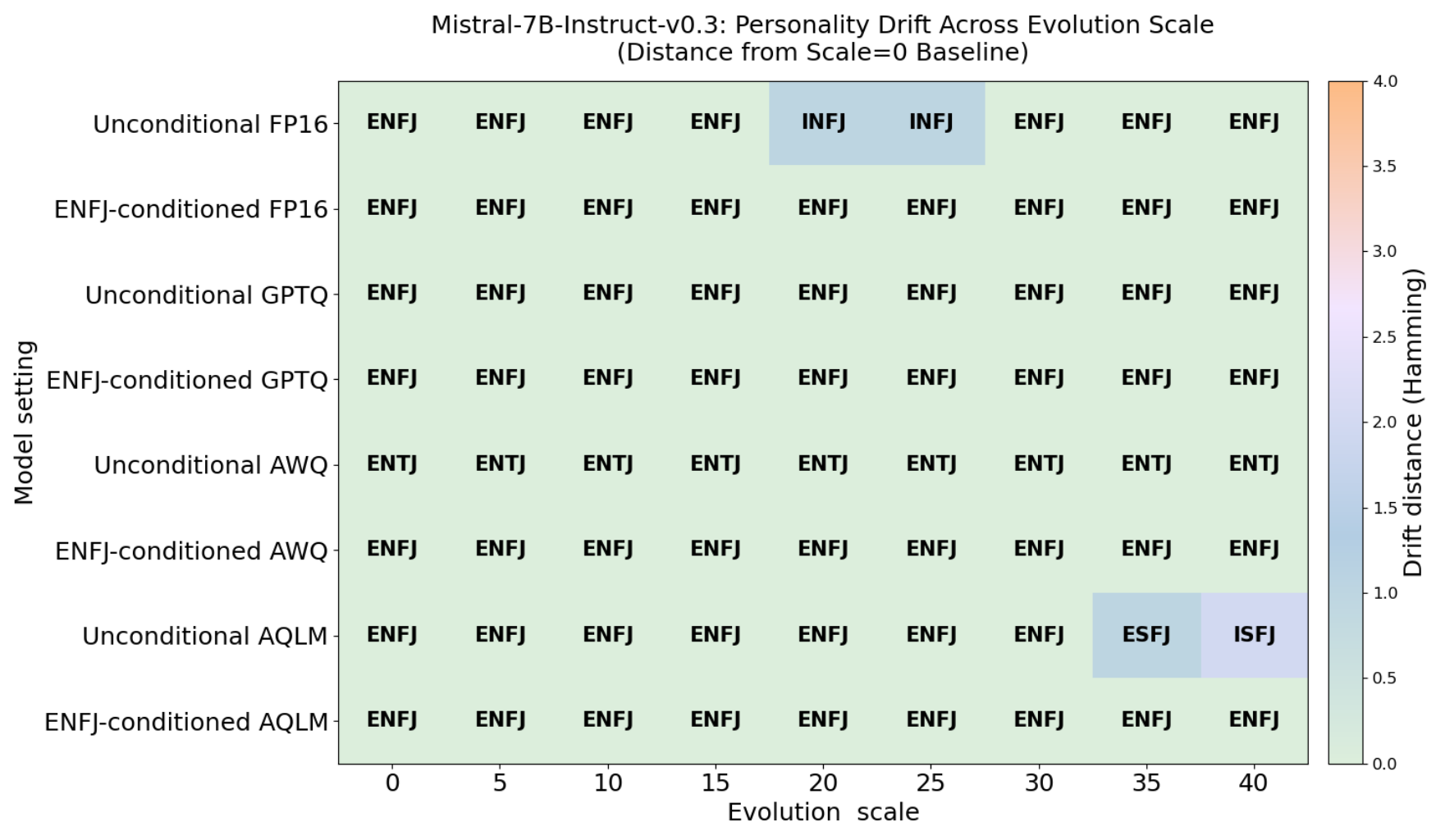}
\caption{Personality drift under increasing evolution scale. Each row corresponds to a model--prompt setting, and each column corresponds to an evolution scale. Cell color indicates Hamming distance between the predicted MBTI type at a given scale and the baseline prediction at scale $0$. Annotated labels show the inferred MBTI type.}
\label{fig:mistral3_7b_instruct_dola_drift}
\end{figure}

\begin{figure}[t]
\centering
\includegraphics[width=0.95\linewidth]{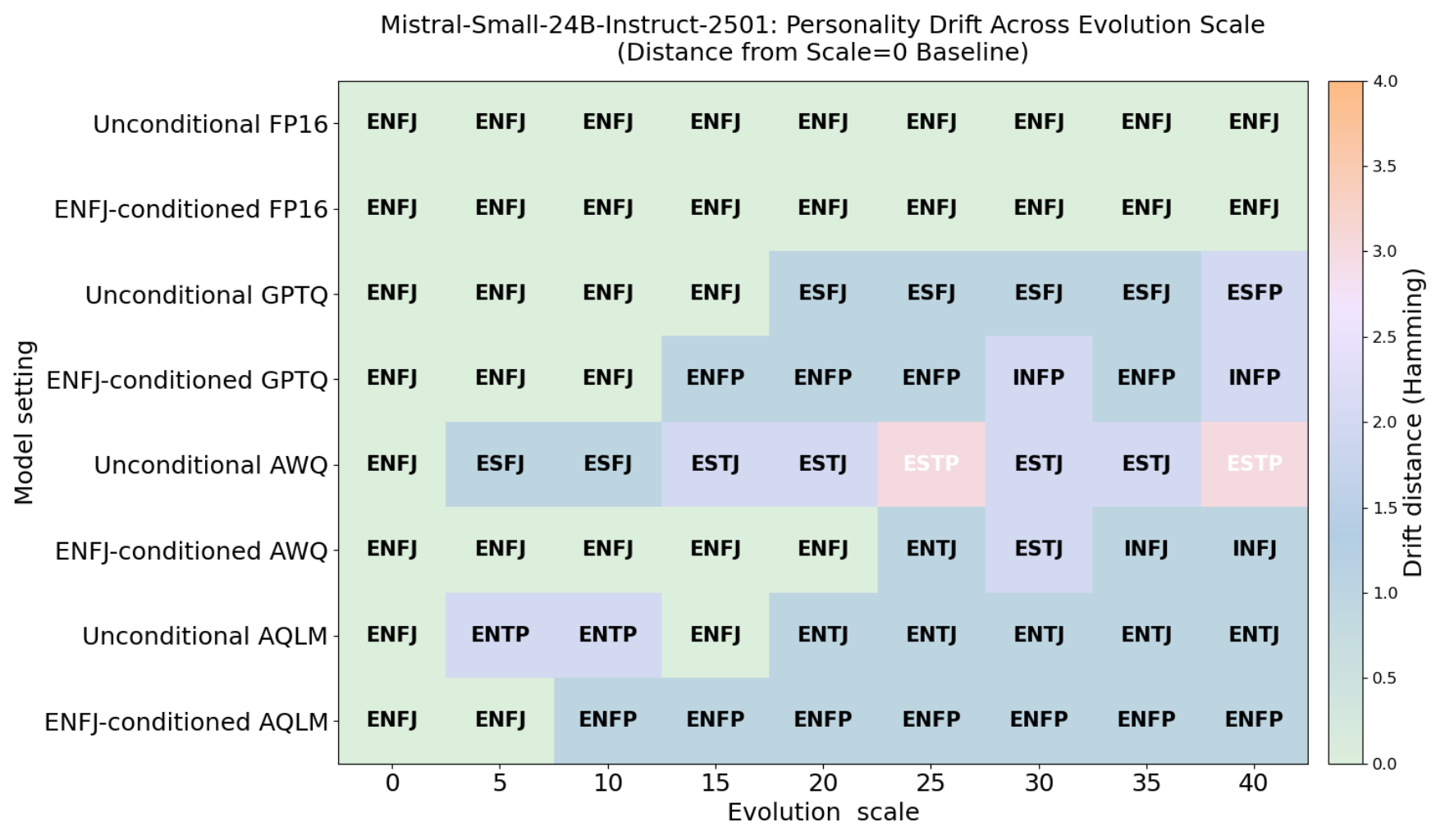}
\caption{Personality drift under increasing evolution scale. Each row corresponds to a model--prompt setting, and each column corresponds to an evolution scale. Cell color indicates Hamming distance between the predicted MBTI type at a given scale and the baseline prediction at scale $0$. Annotated labels show the inferred MBTI type.}
\label{fig:mistral_small_24b_instruct_2501_dola_drift}
\end{figure}

\begin{figure}[t]
\centering
\includegraphics[width=0.95\linewidth]{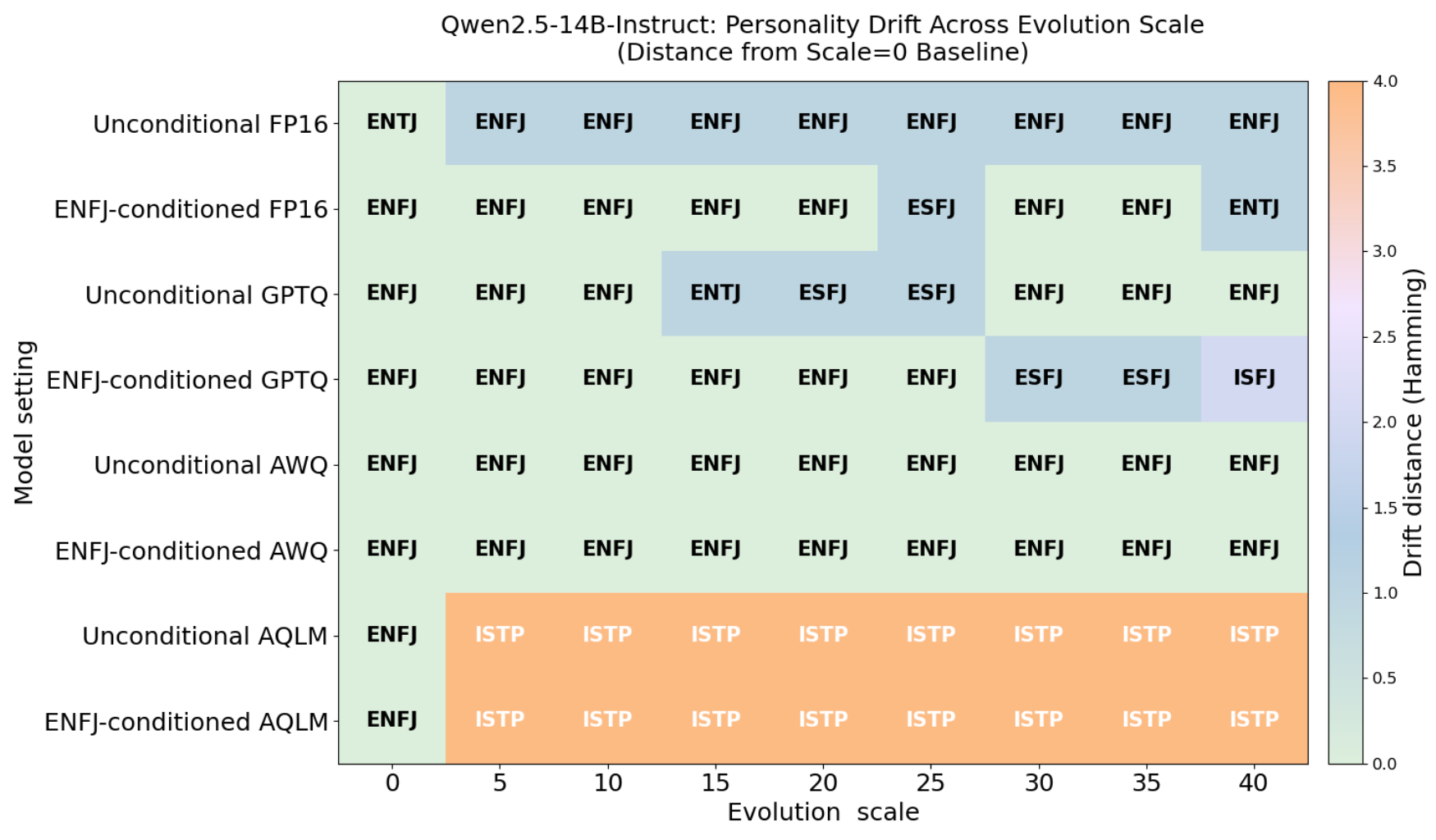}
\caption{Personality drift under increasing evolution scale.Each row corresponds to a model--prompt setting, and each column corresponds to an evolution scale. Cell color indicates Hamming distance between the predicted MBTI type at a given scale and the baseline prediction at scale $0$. Annotated labels show the inferred MBTI type.}
\label{fig:qwen2_5_14b_instruct_dola_drift}
\end{figure}

\begin{figure}[t]
\centering
\includegraphics[width=0.95\linewidth]{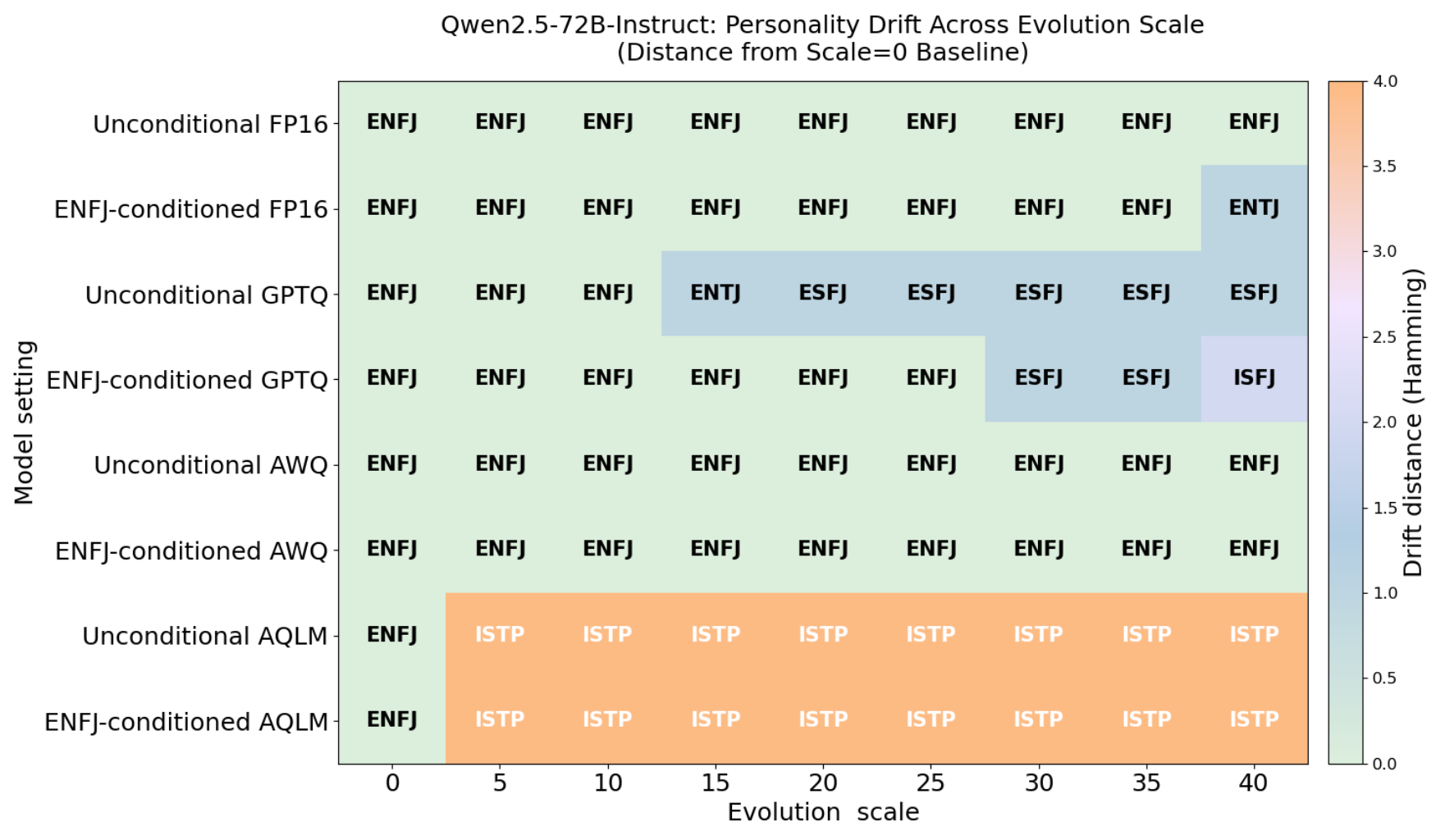}
\caption{Personality drift under increasing evolution scale. Each row corresponds to a model--prompt setting, and each column corresponds to an evolution scale. Cell color indicates Hamming distance between the predicted MBTI type at a given scale and the baseline prediction at scale $0$. Annotated labels show the inferred MBTI type.}
\label{fig:qwen2_5_72b_instruct_dola_drift}
\end{figure}

\end{document}